\pdfoutput=1
\documentclass[11pt]{article}

\usepackage{acl}

\usepackage{times}
\usepackage{latexsym}

\usepackage[T1]{fontenc}
\usepackage[utf8]{inputenc}

\usepackage{microtype}

\usepackage{enumitem}
\usepackage{graphicx}
\usepackage{epstopdf}
\usepackage{subfigure}
\usepackage{caption}

\usepackage{amsmath}
\usepackage{algorithm}
\usepackage{algorithmic}
\usepackage{hyperref}
\usepackage{multirow}
\usepackage{diagbox}
\usepackage{array}

\usepackage{tabularx}
\usepackage{longtable}
\usepackage{boldline}
\usepackage{placeins}
\usepackage{afterpage}
\usepackage{amsfonts}
\usepackage{amsmath}
\usepackage{stfloats}
\usepackage{xcolor}
\usepackage{listings}
\usepackage{color}

\definecolor{dkgreen}{rgb}{0,0.6,0}
\definecolor{gray}{rgb}{0.5,0.5,0.5}
\definecolor{mauve}{rgb}{0.58,0,0.82}

\title{Benchmarking and Improving Compositional Generalization of Multi-aspect Controllable Text Generation}

\author{Tianqi Zhong$^{1}$\thanks{~~The first two authors contributed equally to this work.},$\ $ Zhaoyi Li$^{1,2}$\textsuperscript{$*$}, Quan Wang$^3$, Linqi Song$^{2}$\\  \textbf{Ying Wei}$^{4}$, \textbf{Defu Lian}$^{1}$, \textbf{Zhendong Mao}$^{1}$\thanks{~~Corresponding author: Zhendong Mao.} \\
$^{1}$University of Science and Technology of China, $^{2}$City University of Hong Kong\\
$^{3}$Beijing University of Posts and Telecommunications, 
$^{4}$Nanyang Technological University\\
\texttt{\{ztq602656097, lizhaoyi777\}@mail.ustc.edu.cn}\\
\texttt{wangquan@bupt.edu.cn, linqi.song@cityu.edu.hk}\\ 
\texttt{ying.wei@ntu.edu.sg, \{liandefu, zdmao\}@ustc.edu.cn}
}
\begin{document}
\maketitle
\begin{abstract}
Compositional generalization, representing the model's ability to generate text with new attribute combinations obtained by recombining single attributes from the training data, is a crucial property for multi-aspect controllable text generation (MCTG) methods.
Nonetheless, a comprehensive compositional generalization evaluation benchmark of MCTG is still lacking. 
We propose CompMCTG, a benchmark encompassing diverse multi-aspect labeled datasets and a crafted three-dimensional evaluation protocol, to holistically evaluate the compositional generalization of MCTG approaches. 
We observe that existing MCTG works generally confront a noticeable performance drop in compositional testing. 
To mitigate this issue, we introduce Meta-MCTG, a training framework incorporating meta-learning, where we enable models to learn how to generalize by simulating compositional generalization scenarios in the training phase.
We demonstrate the effectiveness of Meta-MCTG through achieving obvious improvement (by at most 3.64\%) for compositional testing performance in 94.4\% cases\footnote{The code implementation is available at \url{https://github.com/tqzhong/CG4MCTG}.}.
\end{abstract}
\section{Introduction}
Multi-aspect Controllable Text Generation aims to generate fluent text with a combination of attributes from diverse aspects (e.g. sentiment, topic, tense, person, and stuff). In comparison with single-aspect controllable text generation~\cite{zhang-song-2022-discup}, it is more challenging and calls for increasing attention in recent years~\cite{gu-etal-2022-distributional, yang-etal-2023-tailor}.

\begin{figure}[ht]
    \raggedright
    \includegraphics[width=1.0\linewidth, keepaspectratio=true]{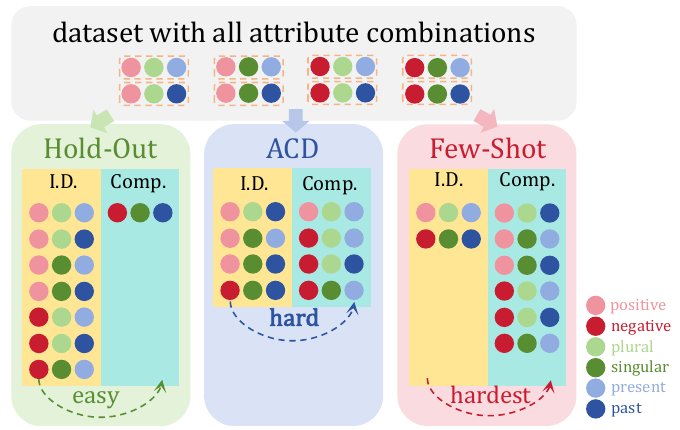}
    \caption{Three evaluation protocols in CompMCTG benchmark, where each set of three colored balls represents texts with these three attribute labels (e.g., positive, plural, and present). "I.D." denotes the \textit{In-Distribution} set and "Comp." denotes the \textit{Compositional} set.}
    \label{fig:figure1}
\end{figure}

Current MCTG methods involve decoding-time-based~\cite{dathathri2019plug, yang-klein-2021-fudge} that modulate output distribution by a well-trained classifier, separate-training-based~\cite{gu-etal-2022-distributional, huang-etal-2023-extensible, gu-etal-2023-controllable, yang-etal-2023-tailor} that train multiple single-aspect modules in turn with single-aspect data and generating multi-aspect text by fusing them, and joint-training-based~\cite{keskar2019ctrl,contrastive-prefix, seen-to-unseen-acl2023}, which train multiple single-aspect modules simultaneously or multi-aspect modules with multi-aspect data. These methods based on pre-trained language models~\cite{radford2019language} have achieved promising results on this task.

However, seldom works focus on the investigation of compositional generalization, a crucial property of MCTG approaches, which refers to the model's ability to generate text with new attribute combinations obtained by recombining single attributes from the training data. For example, we aim for the model to generate text with the attribute combination (\textcolor{red}{negative}, \textcolor{blue}{male}) after training on data with (positive, \textcolor{blue}{male}) and (\textcolor{red}{negative}, female). Due to the difficulties in collecting data with all possible attribute combinations in most real-world scenarios, the capability for compositional generalization is paramount.

To this end, We propose CompMCTG, a comprehensive benchmark to evaluate the compositional generalization of MCTG approaches (Section \ref{sec3:construction}).
% where its comprehensiveness is reflected in both the dimensions of the dataset and the evaluation protocols (Section~\ref{sec3:construction}).
We first collect four popular datasets (from a minimum of two-aspect, eight attribute combinations to a maximum of four-aspect, forty attribute combinations) in the MCTG field to comprise CompMCTG.
The next crucial issue is how to split the dataset to better unveil the compositional generalization risk of MCTG methods. 
Generally, we split the whole dataset $\mathcal C$ into two disjoints sets: in-distribution set $\mathcal{C}_{i.d.}$ and compositional set $\mathcal{C}_{comp}$, where the MCTG model is trained on $\mathcal{C}_{i.d.}$ and tested on both $\mathcal{C}_{i.d.}$ (\textbf{in-distribution testing}) and $\mathcal{C}_{comp}$ (\textbf{compositional testing}).
For an all-sided evaluation, we propose a three-dimensional evaluation protocol containing \textit{Hold-Out}, \textit{ACD}, and \textit{Few-Shot}, which is depicted in Figure \ref{fig:figure1}.
Among them, \textit{Hold-Out} is an easy protocol, which holds a few attribute combinations out from $\mathcal{C}$ as $\mathcal{C}_{comp}$ and uses the remaining combinations as $\mathcal{C}_{i.d.}$. 
\textit{Few-Shot} is the hardest protocol, in which we guarantee every single attribute appears in the $\mathcal C_{i.d.}$ while minimizing $|\mathcal{C}_{i.d.}|$\footnote{We define $|\mathcal C|$ as the number of attribute combinations in $\mathcal C$}.
% We propose a three-dimensional evaluation protocol containing \textit{Hold-Out}, \textit{ACD}, and \textit{Few-Shot} to split the whole dataset $\mathcal{C}$ into two disjoint sets: in-distribution set $\mathcal{C}_{i.d.}$ and compositional set $\mathcal{C}_{comp}$, where the MCTG model is trained on $\mathcal{C}_{i.d.}$ and tested on both $\mathcal{C}_{i.d.}$ (\textbf{in-distribution testing}) and $\mathcal{C}_{comp}$ (\textbf{compositional testing}).
% As depicted in Figure \ref{fig:figure1}, \textit{Hold-Out} and \textit{Few-Shot} are two extreme testing protocols, which are used to test under conditions of the dataset with abundant and scarce attribute combinations respectively.
% \textit{Hold-Out} is the simplest testing, which holds a few attribute combinations out from $\mathcal{C}$ as $\mathcal{C}_{comp}$ and uses the remaining combinations as $\mathcal{C}_{i.d.}$. 
% \textit{Few-Shot} is to test the compositional generalization of MCTG models with limited attribute combinations, where we guarantee each single attribute appears at least one time in the $\mathcal{C}_{i.d.}$ and meanwhile $|\mathcal{C}_{i.d.}|$ is minimized.
To better reflect the capacity of models in cases that $|\mathcal{C}_{comp}|$ is comparable to $|\mathcal{C}_{i.d.}|$, which are closer to real-world scenarios, we design \textit{\textbf{A}ttribute \textbf{C}ompound \textbf{D}ivergence} (\textit{ACD}), where we make $|\mathcal{C}_{i.d.}| = |\mathcal{C}_{comp}|$.
The core idea of \textit{ACD} is to maximize the distributional divergence between $\mathcal C_{i.d.}$ and $\mathcal C_{comp}$.
Compared with random sampling that contributes to similar distributions between $\mathcal C_{i.d.}$ and $\mathcal C_{comp}$ easily~\cite{seen-to-unseen-acl2023}, \textit{ACD} can better amplify the compositional generalization risk while random-based splits often lead to gross under-estimation (Section~\ref{sec3:analysis}).
% Due to the exponential complexity (Appendix~\ref{appendix:complexity}), we are incapable of sweeping over all possible such splits.
% The core idea of \textit{ACD} is to calculate the distributions of attribute compounds\footnote{attribute compound refers to the tuple of two single attributes. e.g., (``positive”,``present”).} in $\mathcal{C}_{i.d.}$ and $\mathcal{C}_{comp}$ and maximize the divergence between them.
% Note that \citet{seen-to-unseen-acl2023} study the compositional generalization of Multi-aspect Controllable Dialogue Generation (MCDG) while they only adopt a two-aspect dataset and employ random sampling for dataset splitting.
% In comparison with randomly sampling half attribute combinations as $\mathcal{C}_{comp}$ (like \citet{seen-to-unseen-acl2023}), we demonstrate that adopting \textit{ACD} can better unveil the hidden compositional risk while random-based splits often lead to gross under-estimation (Section~\ref{sec3:analysis}). 

Through the results on CompMCTG (Section \ref{sec3:evaluation}), we observe that all of the evaluated MCTG baseline approaches are faced with a noticeable performance drop between in-distribution and compositional testing.
% To mitigate this issue, we propose Meta-MCTG, an easy-to-implement and broadly applicable training framework for most MCTG methods, incorporating Model-Agnostic Meta Learning (MAML)~\cite{maml-icml2017} into MCTG's training procedure.
To further enhance the compositional generalization performance of joint-training-based methods which generally perform the best among all baselines, we propose Meta-MCTG (Section~\ref{sec4}), a training framework incorporating meta-learning~\cite{maml-icml2017}, in which we enable models to learn how to generalize by simulating compositional generalization scenarios in the training phase.
Firstly, we train the original model on a training batch $\mathcal B_{train}$, perform one step of gradient descent, and save the updated parameters to a backup model without updating the original model's parameters. Secondly, we sample a ``pseudo compositional” batch $\mathcal B_{pcomp}$ from the training set where the attribute combinations are the re-combination of those in $\mathcal B_{train}$ and train the backup model on $\mathcal B_{pcomp}$. Finally, we combine the losses from both steps and perform one step of gradient descent to update the original model's parameters.
Compared with solely training the model on $\mathcal B_{train}$, introducing $\mathcal B_{pcomp}$ enables the model's parameters to update in a direction that not only focuses on fitting the training data but also takes out-of-distribution data into account, which helps to elevate model's capability of compositional generalization. 
% for each training batch $\mathcal B_{train}$, we sample a ``pseudo compositional” batch $\mathcal B_{pcomp}$ from training set in which the attribute combinations are the re-combination of those in $\mathcal B_{train}$
% where we design an auxiliary training objective for MCTG and leverage meta-learning~\cite{maml-icml2017} to jointly optimizing it and the original one.
% Among them, we observe that joint-training-based baselines generally achieve better overall performance and lower performance drop in compositional testing. Nonetheless, there still exists a non-negligible performance drop for them in compositional testing, which highly calls for our attention.
% We analyze the failure of generating text under the control of novel combinations of familiar attributes can be mainly attributed to the over-fitting of language models to local optima of attribute combinations in the training set. 
% Inspired by previous works~\cite{Li2018MLDG,wang-etal-2021-meta,conklin-etal-2021-meta}, we mimic compositional generalization scenarios in the training phase: for a training batch $\mathcal{B}_{train}$, we sample a ``pseudo compositional” batch $\mathcal{B}_{pcomp}$ in which the control conditions are the re-combination of the control attributes in $\mathcal{B}_{train}$. We design an auxiliary optimization objective (on $\mathcal{B}_{pcomp}$) and append it to the original training objective (on $\mathcal{B}_{train}$), navigating the training process away from the local optima that only fits training control conditions.
We implement Meta-MCTG on three top-performing joint-training-based MCTG baselines and conduct extensive experiments on CompMCTG, demonstrating the effectiveness of Meta-MCTG through achieving obvious improvement (by at most 3.64\%) for compositional testing in 94.4\% cases.

Our main contributions are three-fold: 
% \begin{itemize}[leftmargin=*]
%     \item We propose CompMCTG, the first holistic benchmark targeting on compositional generalization for MCTG, incorporating four popular datasets and a crafted three-dimensional evaluation protocols.
%     \item We conduct extensive experiments on CompMCTG with eight representative MCTG baselines and two additional LLMs, unveiling noticeable compositional generalization risk in them and demonstrating the necessity of designs in CompMCTG.
%     \item We propose Meta-MCTG, incorporating meta-learning into MCTG training process, to mitigate MCTG models’ over-fitting to attribute combinations seen in the training phase and improve their capacity of compositional generalization. 
% \end{itemize}
(1) We propose CompMCTG, the first holistic benchmark targeting compositional generalization for MCTG, incorporating four popular datasets and a crafted three-dimensional evaluation protocol.
(2) We conduct extensive experiments on CompMCTG with eight representative MCTG baselines and two additional LLMs, unveiling noticeable compositional generalization risk in them and demonstrating the necessity of designs in CompMCTG.
(3) We propose Meta-MCTG, incorporating meta-learning into the MCTG training process, to mitigate MCTG models’ over-fitting to attribute combinations seen in the training phase and improve their capacity for compositional generalization. 
To the best of our knowledge, we are the first to comprehensively evaluate MCTG on compositional generalization and introduce meta-learning into MCTG to improve composition generalization.
% 1) We construct CompMCTG in Section~\ref{sec3:construction}, the first holistic benchmark targeting on compositional generalization for open-domain MCTG, 2) We conduct extensive experiments on CompMCTG benchmark with eight popular baseline MCTG approaches to reveal significant compositional generalization risk in them and demonstrate the effectiveness of our design in construction of CompMCTG as well in Section~\ref{sec3:evaluation} and ~\ref{sec3:analysis} and 3) We propose MetaMCTG in Section~\ref{sec4}, incorporating meta-learning~\cite{maml-icml2017} into MCTG training process, to relieve the overfitting of text generators to training combinations of control attributes and encourage them to better generalize to novel combinations of familiar control attributes through mimicing the compositional testing in the training phase. To the best of our knowledge, we are the first to introduce meta-learning into MCTG towards improving composition generalization.

\section{Related Work}
\paragraph{Multi-aspect Controllable Text Generation}
%% Zhong
\label{sec-related works:mctg}
Existing works on MCTG primarily fall into the following three categories: The first is \textbf{decoding-time-based}~\cite{dathathri2019plug,yang-klein-2021-fudge,krause2021gedi}, which uses a well-trained classifier or conditional language model to adjust the output probability distribution of a frozen causal language model. The second is \textbf{separate-training-based}, which trains single-attribute modules~\cite{yang-etal-2023-tailor, huang-etal-2023-extensible}, Energy-based Models~\cite{mireshghallah-etal-2022-mix, NEURIPS2022_3e25d1af} or latent space representations~\cite{gu-etal-2022-distributional,gu-etal-2023-controllable}  using single-attribute label data, and controls the generation by concatenating individual modules, Energy-based Models or seeking the intersection of different attribute representations in the latent space. The third is \textbf{joint-training-based}, which trains multi-attribute modules~\cite{keskar2019ctrl,seen-to-unseen-acl2023,qian-etal-2022-controllable} simultaneously using multi-attribute label data. \citet{qian-etal-2022-controllable} add a prefix~\cite{li-liang-2021-prefix} for each attribute and train these prefixes using a contrastive loss. \citet{seen-to-unseen-acl2023} encode different control codes (word embedding of attribute tokens) into prompts~\cite{prompt-tuning} using a fully connected layer and train this layer using a contrastive loss similar to \citet{qian-etal-2022-controllable}.
% Multi-attribute controllable text generation (MCTG) can be categorized into three main types~\cite{zhang2022survey}. The first type is retraining or refactoring~\cite{keskar2019ctrl, chan2020cocon, zhang-etal-2020-pointer}, which aims to fully retrain a language model from scratch or modify the language model structure. The second type is decoding-time approach~\cite{dathathri2019plug, krause2021gedi, yang-klein-2021-fudge}, whose main idea is to adjust the probability distribution of the next token during the decoding stage of the language model. The third type is parameter efficient tuning. These methods generally build upon the prefix-tuning~\cite{li-liang-2021-prefix} or prompt-tuning~\cite{lester-etal-2021-power} and have the advantages of low training cost and fast inference speed~\cite{qian-etal-2022-controllable, zhang-song-2022-discup, gu-etal-2022-distributional, huang-etal-2023-extensible, yang-etal-2023-tailor}. Although these methods have achieved remarkable results in MCTG tasks, they typically perform poorly in generating texts with unknown attribute combinations, indicating that their ability of compositional generalization is lacking~\cite{seen-to-unseen-acl2023}.
\paragraph{Compositional Generalization}
%% Li: Seq2Seq:{Semantic Parsing, Machine Translation}; Text Classification; CZSL & VQA(Multi-Modality & Vision)
%% Li: Controllable Text Generation: Dialogue(acl'23) -> deficiency(specific to dialogue; assessment is not comprehensive enough).
Existing works on compositional generalization involve various NLP topics: Semantic Parsing~\cite{spanparse-acl2021, transformer_comp-acl2022, drozdov2023compositional, l2s2-acl2023}, Machine Translation~\cite{cognition-acl2021, dangle-acl2022, composition-emnlp2023}, Text Classification~\cite{cls-acl2021,chai2023compositional}, Complex Reasoning~\cite{l2m-iclr2023,press-etal-2023-measuring,li2024understanding,lin2024ask} and stuff.
Nonetheless, in the field of open-domain controllable text generation, compositional generalization, which we target and reveal as the necessity for the robustness of neural language generators in this paper, remains under-explored. 
~\cite{seen-to-unseen-acl2023} investigates compositional generalization focusing on a neighboring topic, controllable dialogue generation. We regard their work as a starting point of our research and further depict the deficiency of its naive evaluation protocol, for the underestimation of the compositionality gap in more realistic scenarios~\cite{mcd-iclr2020}.
% Describe Tree-MAML 
% Consequently, our work provides a comprehensive and crafted benchmark to systematically measure the compositionality gap of existing approaches and a general meta-learning~\cite{maml-icml2017} based procedure to further narrow the gap on the basis of them.

% \input{sections/benchmarking}
\section{Benchmark: CompMCTG}

% a breif sentence to introduce the main content of this section.
We propose CompMCTG, a novel benchmark to comprehensively evaluate the compositional generalization capacity of MCTG approaches.
The superiority and novelty of CompMCTG are out of its scale of dataset and its three-dimensional evaluation protocol (Section~\ref{sec3:construction}).
We select eight representative baseline approaches (Section~\ref{sec3:baseline}), evaluate their performance on our CompMCTG benchmark, and unveil their struggling on compositional testing (Section~\ref{sec3:evaluation}).
Moreover, systematic analysis towards exploring the behaviors of baseline approaches under different evaluation protocols of CompMCTG is provided in Section~\ref{sec3:analysis}, which highlights: 1) its capacity to dig out the potential generalization risk of evaluated approaches and 2) the undervalued compositionality gap in the previous work~\cite{seen-to-unseen-acl2023} as well.

\subsection{On the Construction of CompMCTG}
\label{sec3:construction}
\paragraph{Data Source}
We collect commonly used and open-sourced datasets for our usage. Consequently, we select a shopping review dataset: \textit{Amazon Review}~\cite{amazon_review}, a mixture of movie(IMDB~\cite{imdb}), tablet, automobile(Sentube~\cite{Sentube}) and hotel(OpenNER~\cite{OpenNER}) review dataset: \textit{Mixture}~\cite{Mixture}, and two restaurant review datasets: \textit{YELP}~\cite{YELP0,YELP} and \textit{Fyelp}~\cite{Fyelp}. 
Details of these datasets are concluded in Appendix~\ref{appendix-datasets}.

\paragraph{Three-Dimensional evaluation Protocol}
We design a three-dimensional(\textit{Hold-Out}, \textit{ACD} and \textit{Few-Shot}) evaluation protocol, aiming to sufficiently explore the compositional generalization capacity of existing approaches.
Supposing that dataset $\mathcal{D}$\footnote{Each datum in $\mathcal{D}$ consists of two components: $(c, x)$, where $c$ denotes the \textit{condition part}, a combination of several attributes of different aspects (e.g., sentiment:``positive”, tense:``past”, and topic:``basketball”) and $x$ denotes the \textit{text part}, a span of text corresponding to these conditions. For brevity, we omit the text part and use the \textit{condition part} to represent the data in this section.} contains $m$ distinct aspect sets: $\mathcal{A}_1$, $\mathcal{A}_2$, ..., $\mathcal{A}_m$ and a specific aspect $\mathcal{A}_i$ ($1\leq i \leq m$) has $a_i$ kinds of different attribute values in its set: $\mathcal{A}_i = \{A_i^{1}, A_i^{2}, ..., A_i^{a_i}\}$, we denote the whole attribute combination set as the continued Cartesian product $\mathcal{C} = \mathcal{A}_1 \times \mathcal{A}_2 \times ... \times \mathcal{A}_m = \{(A_i^{t_i})_{1 \leq i \leq m} | 1\leq t_i \leq a_i\}$.  
The core of constructing CompMCTG is to \textbf{split} the attribute combination set $\mathcal{C}$ into \textit{in-distribution} set $\mathcal{C}_{i.d.}$ and \textit{compositional} set $\mathcal{C}_{comp}$.  % introduce split
Basically, $\mathcal{C}_{comp}$ has no intersection with $\mathcal{C}_{i.d.}$ and any attribute combination in $\mathcal{C}_{comp}$ can be derived through recombining single attributes in $\mathcal{C}_{i.d.}$. 
% The motivation behind the "split" is that we can train generative language models on text with control conditions in $\mathcal{C}_{i.d.}$ and test their compositional generalization with control conditions in $\mathcal{C}_{comp}$. 
% We train generative language models on text with control conditions in $\mathcal{C}_{i.d.}$ and test their in-distribution (in-distribution testing) and compositional (compositional testing) generalization with control conditions in $\mathcal{C}_{i.d.}$ and $\mathcal{C}_{comp}$ respectively. 
Hence we have the formal definition of \textbf{an eligible split} $s(\mathcal{C}) = \mathcal{C}_{i.d.}, \mathcal{C}_{comp}$ as following:

\small
\begin{equation}
\begin{aligned}
    & \mathcal{C}= \mathcal{C}_{i.d.}\cup \mathcal{C}_{comp},\ \mathcal{C}_{i.d.}\cap \mathcal{C}_{comp} = \emptyset \\
    &  \{attribute | \exists c \in \mathcal{C}_{comp}, attribute \in c \} \subseteq \\
    &  \{attribute | \exists c \in \mathcal{C}_{i.d.}, attribute \in c\}
\end{aligned}
\label{eq:def_split}
\end{equation}

\normalsize
\textit{Hold-Out} is an easy evaluation protocol, which holds a few attribute combinations out from $\mathcal{C}$ as $\mathcal{C}_{comp}$ and uses the remaining attribute combinations as $\mathcal C_{i.d.}$. Supposing $|\mathcal{C}_{comp}|$ equals to $k$ ($k$ is relatively small so that the split is eligible), there are $\binom{|\mathcal{C}|}{k}$ different kinds of splits. In our benchmark, we set $k=1$, and the final result is the average across $\binom{|\mathcal{C}|}{k}$ scenarios to eliminate bias.
% Formally, the set of \textit{Hold-Out} split can be denoted as:
% $S_{hold-out} = \{(\mathcal{C}_{i.d.}, \mathcal{C}_{comp}) | \mathcal{C}_{comp} \subset \mathcal{C},|\mathcal{C}_{comp}|=k, \mathcal{C}_{i.d.} = \mathcal{C}  \backslash \mathcal{C}_{comp})\}$.
% Such split method is also adopted in \cite{seen-to-unseen-acl2023} while its shortcoming lies in that $|\mathcal{C}_{comp}|$ is so insignificant compared with $|\mathcal{C}_{i.d.}|$ that it fails to fully mimic the realistic scenarios, where text data of a relatively large proportion of attribute combinations are unavailable, and hence leads to insufficient assessment to compositional generalization risk (We demonstrate this point in Section~\ref{sec3:analysis}).\par

\begin{table*}[t]
\centering
\resizebox{\textwidth}{!}{
\begin{tabular}{lcc|cccc|cccc|ccc}
\hline
\multirow{2}{*}{\textbf{Method}} &  \multicolumn{2}{c}{\textit{Original}} & \multicolumn{4}{c}{\textit{Hold-Out}} & \multicolumn{4}{c}{\textit{ACD}} & \multicolumn{3}{c}{\textit{Average}}  \\

  & \textit{A}\textsubscript{\textit{i.d.}}$(\uparrow)$& \textit{P}\textsubscript{\textit{i.d.}} $(\downarrow)$
 & \textit{A}\textsubscript{\textit{i.d.}} $(\uparrow)$ & \textit{P}\textsubscript{\textit{i.d.}}$(\downarrow)$ & \textit{A}\textsubscript{\textit{comp}}$(\uparrow)$ &  \textit{P}\textsubscript{\textit{comp}}$(\downarrow)$
 & \textit{A}\textsubscript{\textit{i.d.}} $(\uparrow)$ & \textit{P}\textsubscript{\textit{i.d.}} $(\downarrow)$ & \textit{A}\textsubscript{\textit{comp}} $(\uparrow)$&  \textit{P}\textsubscript{\textit{comp}}$(\downarrow)$
  & \textit{A}\textsubscript{\textit{avg}} $(\uparrow)$& \textit{P}\textsubscript{\textit{avg}} $(\downarrow)$  & \textit{G}\textsubscript{\textit{avg}} $(\downarrow)$ \\
 \hline
 \hline
\textbf{LLM+In-context Learning}\\
\hline
 \textit{LLaMA-2}\footnotesize{~\cite{llama2}}    &$61.53$\% & $27.30$  &$62.61$\%& $25.55$ &$40.82$\%&$23.80$ &$62.98$\%& $28.31$ & $42.11$\% &$24.63$  &$54.01$\% &$25.92$ &$33.97$\% \\
  \textit{ChatGPT}\footnotesize{~\cite{openaiChatGPT}}    &$57.51$\% & $18.03$  &$56.62$\%& $18.29$ &$49.21$\%&$18.49$ &$57.13$\%& $18.27$ & $49.75$\% &$18.22$  &$54.04$\% &$18.26$ &$13.00$\% \\
\hline
\textbf{Decoding-Time based} \\
 \hline
 \textit{PPLM}\footnotesize{~\cite{dathathri2019plug}}    &$40.91$\% & $322.59$  &$41.05$\%& $325.09$ &$40.62$\%&$340.76$ &$42.25$\%& $328.07$ & $39.60$\% &$325.74$  &$40.89$\% &$328.45$ &$3.66$\% \\
\textit{Fudge}\footnotesize{~\cite{yang-klein-2021-fudge}}  &$60.12$\%& $178.51$   &$59.35$\%  & $179.47$ &$42.10$\% &$252.08$  &$57.17$\%& $175.66$ & $41.49$\% &$223.08$  &$52.05$\% &$201.76$ &$28.25$\% \\

\hline
\textbf{Separate-Training based} \\
\hline
\textit{Dis-Lens}\footnotesize{~\cite{gu-etal-2022-distributional}}  &$85.46$\%& $123.72$   &$84.84$\%  & $95.84$ &$55.58$\% &$104.89$  &$85.54$\%&$90.87$& $49.52$\% &$112.60$  &$72.19$\% &$105.58$&$22.30$\%  \\
\textit{Prior}\footnotesize{~\cite{gu-etal-2023-controllable}} &$73.85$\%& $119.91$   &$73.64$\%  & $108.58$ &$49.93$\% &$97.64$  &$78.24$\%&$113.73$& $50.05$\% &$97.63$  &$65.14$\% &$107.50$ &$34.11$\%  \\

\hline
\textbf{Joint-Training based} \\
\hline
\textit{CTRL}\footnotesize{~\cite{keskar2019ctrl}}  &$79.10$\%& $54.17$   &$78.89$\%  & $51.20$ &$75.09$\% &$51.22$  &$77.83$\%&$51.71$& $69.96$\% &$51.28$  &$\mathbf{76.17}$\textbf{\%} &$51.92$ &$\mathbf{7.46}$\%  \\
\textit{CatPrompt}\footnotesize{~\cite{yang-etal-2023-tailor}}  &$63.91$\%& $74.53$   &$63.95$\%  & $73.24$ &$60.32$\% &$69.13$  &$60.53$\%&$98.08$& $48.25$\% &$68.45$  &$59.39$\% &$76.69$ &$12.98$\%  \\
\textit{Con.Prefix}\footnotesize{~\cite{contrastive-prefix}}  &$83.99$\%& $79.29$  &$83.75$\%& $80.49$ &$80.36$\%&$87.19$ &$81.15$\%& $80.71$ &$69.84$\% &$83.90$  &$\mathbf{79.82}$\textbf{\%} &$82.32$ &$\mathbf{8.99}$\%  \\
\textit{DCG}\footnotesize{~\cite{seen-to-unseen-acl2023}} &$79.93$\%& $56.37$   &$79.72$\%  & $62.05$ &$76.66$\% &$64.40$  &$78.43$\%&$57.97$& $67.7$\% &$61.11$  &$\mathbf{76.49}$\textbf{\%} &$60.38$ &$\mathbf{8.76}$\%  \\
\hline 
\end{tabular}
}
\caption{
Averaged overall evaluation results for state-of-the-art baseline approaches on our CompMCTG benchmark (\textit{Hold-Out} testing and \textit{ACD} testing). \textit{A}, \textit{P} and \textit{G} are the abbreviations of accuracy, perplexity, and gap (we explain the meaning of ``gap” in Section~\ref{sec3:evaluation}.) respectively. Subscript \textit{i.d.} and \textit{comp} refer to in-distribution and compositional generalization performance. Each value in this table is the average (Please find the detailed results for each dataset in Appendix~\ref{results on datasets}) of testing performances on four component datasets of CompMCTG: Amazon Review~\cite{amazon_review}, Fyelp~\cite{Fyelp}, YELP~\cite{YELP0,YELP} and Mixture~\cite{Mixture}.
}
\label{tab:compmctg}
\end{table*}

\begin{table}[t]
\centering
\resizebox{0.5\textwidth}{!}{
\begin{tabular}{lcccc}
\hline
\multirow{2}{*}{\textbf{Method}} & \multicolumn{4}{c}{\textit{Few-Shot}} \\
 & \textit{A}\textsubscript{\textit{i.d.}}$(\uparrow)$ & \textit{P}\textsubscript{\textit{i.d.}} $(\downarrow)$ & \textit{A}\textsubscript{\textit{comp}} $(\uparrow)$&  \textit{P}\textsubscript{\textit{comp}}$(\downarrow)$\\
 \hline\hline
 \textbf{LLM+In-context Learning}\\
\hline
\textit{LLaMA-2}\footnotesize{~\cite{llama2}} & $62.78$\% & $26.08$ & $42.99$\% & $23.90$ \\
\textit{ChatGPT}~\cite{openaiChatGPT} & $56.64$\% & $18.62$ & $49.50$\% & $17.71$ \\
\hline
\hline
\textbf{Decoding-Time based} \\
\hline
\textit{PPLM}\footnotesize{~\cite{dathathri2019plug}}  &$43.07$\%& $361.60$ &$40.21$\% &$330.94$  \\
\textit{Fudge}\footnotesize{~\cite{yang-klein-2021-fudge}}  &$58.00$\%& $167.31$ &$40.90$\% &$224.91$  \\
\hline
\hline
\textbf{Separate-Training based} \\
\hline
\textit{Dis-Lens}\footnotesize{~\cite{gu-etal-2022-distributional}}  &$87.81$\%& $95.05$ &$51.47$\% &$116.68$  \\
\textit{Prior}\footnotesize{~\cite{gu-etal-2023-controllable}} &$85.19$\%& $118.97$ &$51.71$\% &$104.16$  \\
\hline
\hline
\textbf{Joint-Training based} \\
\hline
\textit{CTRL}\footnotesize{~\cite{keskar2019ctrl}}  &$77.87$\%& $48.48$ &$65.94$\% &$48.28$  \\
\textit{CatPrompt}\footnotesize{~\cite{yang-etal-2023-tailor}} &$62.47$\%& $163.66$ &$46.23$\% &$130.50$  \\
\textit{Con.Prefix}\footnotesize{~\cite{contrastive-prefix}}  &$79.89$\%& $88.34$ &$57.56$\% &$93.31$  \\
\textit{DCG}\footnotesize{~\cite{seen-to-unseen-acl2023}} &$78.89$\%& $63.22$ &$59.27$\% &$68.14$  \\
\hline 
\end{tabular}
}
\caption{
Averaged overall evaluation results for state-of-the-art baseline approaches on our CompMCTG benchmark (\textit{Few-Shot} testing). Each value in this table is the average of testing performances on four component datasets of CompMCTG: Amazon Review~\cite{amazon_review}, Fyelp~\cite{Fyelp}, YELP~\cite{YELP0,YELP} and Mixture~\cite{Mixture}.
}
\label{tab:few-shot}
\end{table}

\textit{Few-Shot} is the hardest evaluation protocol, in which we guarantee every single attribute appears in the $\mathcal C_{i.d.}$ while minimizing $|\mathcal C_{i.d.}|$, which simulate the scenarios of the low-data regime.
% Moreover, we design \textbf{\textit{Few-Shot}} splits to evaluate the compositional generalization of text generator systems with each attribute appearing at most few times in $\mathcal{C}_{i.d.}$, which simulate the settings of the low-data regime and thus are most challenging.

% To better reflect the capacity of models in cases that $|\mathcal C_{comp}|$ is comparable to $|\mathcal C_{i.d.}|$, we propose a divergence-based split method \textbf{\textit{ACD}}, where we guarantee $|\mathcal{C}_{i.d.}|$ equals to $|\mathcal{C}_{comp}|$.
While in most real-world scenarios, $|\mathcal C_{comp}|$ is comparable to $|\mathcal C_{i.d.}|$.
A crucial issue to this situation is how we divide $\mathcal{C}$ into $\mathcal{C}_{i.d.}$ and $\mathcal{C}_{comp}$ as the exponential complexity of sweeping over all of the eligible possibilities (We discuss this point in Appendix~\ref{appendix:complexity}). 
Thus focusing on a representative subset of them is a feasible solution.
% Randomly sampling half of the attribute combinations as $\mathcal{C}_{i.d.}$ and leaving the rest as $\mathcal{C}_{comp}$ is an intuitive approach, while we argue and demonstrate that such \textbf{\textit{Random Sampling}} method easily results in under-estimation of compositionality gap in Section~\ref{sec3:analysis}. 
Inspired by ~\cite{mcd-iclr2020}, we propose \textit{ACD}, where we keep $|\mathcal C_{i.d.}|=|\mathcal C_{comp}|$ and construct representative splits by maximizing the \textit{\textbf{A}ttribute \textbf{C}ompound \textbf{D}ivergence} between $\mathcal{C}_{i.d.}$ and $\mathcal{C}_{comp}$.
% In comparison, our \textbf{\textit{ACD}} method is to construct representative splits through maximizing the \textit{\textbf{A}ttribute \textbf{C}ompound \textbf{D}ivergence} between $\mathcal{C}_{i.d.}$ and $\mathcal{C}_{comp}$. 
The term \textit{attribute compound} refers to a specific tuple of two attributes: $(A_i^{t_i}, A_j^{t_j}),i\le j, 1\leq t_i \leq a_i, 1\leq t_j \leq a_j$, which characterizes the co-occurrence of two attributes in one attribute combination $c\in\mathcal{C}$. 
Firstly, we calculate the frequency density of the \textit{attribute compound} $(A_i^{t_i}, A_j^{t_j})$ in the combination sets $\mathcal C\in\{\mathcal C_{i.d.}, \mathcal C_{comp}\}$ and obtain two frequency distributions $(f_{\mathcal{C}_{i.d.}}((A_i^{t_i}, A_j^{t_j})))_{i,j,t_i,t_j}$ and $(f_{\mathcal{C}_{comp}}((A_i^{t_i}, A_j^{t_j})))_{i,j,t_i,t_j}$:

{
\small
\begin{equation}
\label{eq:def_freq}
\begin{aligned}
    & f_\mathcal{C}((A_i^{t_i}, A_j^{t_j})) = \frac{\sum_{c\in\mathcal{C}}\mathbb{I}(A_i^{t_i}\in c \land A_j^{t_j}\in c)}{\sum_{c\in\mathcal{C}}\sum_{x\in c,y\in c,x\neq y} 1} \\
   & = \frac{2\sum_{c\in\mathcal{C}}\mathbb{I}(A_i^{t_i}\in c \land A_j^{t_j}\in c)}{m(m-1)|\mathcal{C}|}
\end{aligned}
\end{equation}
}Then we introduce the Chernoff Coefficient $S(P, Q)$~\cite{CHUNG1989280} to measure the scale of similarity between two probability distributions $P$ and $Q$ (i.e., $P=(p_1, p_2, ..., p_n)$ and $Q=(q_1, q_2, ..., q_n)$, $S(P, Q)=\sum_{i=1}^n p_i^\alpha q_i^{1-\alpha}\in [0,1]$)\footnote{$\alpha\in [0,1]$ is a hyperparameter to control our tolerance on the difference between $P$ and $Q$: }. 
Finally, we define the \textit{Attribute Compound Divergence} as $D(P_{i.d.}, P_{comp})=1-S(P_{i.d.}, P_{comp})\in [0,1]$ to measure the divergence between $\mathcal C_{i.d.}$ and $\mathcal C_{comp}$, where distribution $P_{i.d.}$ and $P_{comp}$ represent $(f_{\mathcal{C}_{i.d.}}((A_i^{t_i}, A_j^{t_j})))_{i,j,t_i,t_j}$ and $(f_{\mathcal{C}_{comp}}((A_i^{t_i}, A_j^{t_j})))_{i,j,t_i,t_j}$, respectively.
% Chernoff coefficient~\cite{CHUNG1989280} $S(P,Q)\in [0,1]$ is a conventional tool to measure the similarity between two probability distributions (i.e., $P=(p_1, p_2, ..., p_n)$ and $Q=(q_1, q_2, ..., q_n)$) and correspondingly we use $D(P,Q) = 1-S(P,Q) = 1-\sum_{i=1}^n p_i^\alpha q_i^{1-\alpha}\in [0, 1]$ to measure the divergence between them\footnote{$\alpha\in [0,1]$ is a hyperparameter to control our tolerance on the difference between $P$ and $Q$: }. 
In the real construction of \textit{ACD} splits, we adopt a greedy-based hill climbing algorithm~\cite{russel2010}\footnote{The algorithm pseudo-code is available in Appendix~\ref{sec:appendix-pseduocode}.} to \textbf{sample satisfactory splits which maximize} $D(P_{i.d.}, P_{comp})$.
% where $P_{i.d.}$ and $P_{comp}$ represent the frequency density distributions of \textit{attribute compounds} $(f_{\mathcal{C}_{i.d.}}((A_i^{t_i}, A_j^{t_j})))_{i,j,t_i,t_j}$ and $(f_{\mathcal{C}_{comp}}((A_i^{t_i}, A_j^{t_j})))_{i,j,t_i,t_j}$, respectively.
%($1\leq i \le j \leq m,1\leq t_i \leq a_i, 1\leq t_j \leq a_j$)

Note that for \textit{Amazon Review} and \textit{Mixture} datasets, \textit{ACD} protocol degenerates to \textit{Few-Shot} protocol as these datasets only contain two aspects and we can not optimize the attribute compound divergence in that situation.
% \begin{table}[t]
% \centering
% \resizebox{0.5\textwidth}{!}{
% \begin{tabular}{|lccc|}
% \hline
% \textbf{Dataset} & $m$ & $|\mathcal{C}|$ & size \\
% \hline
% \textit{Amazon Review}\footnotesize{~\cite{amazon_review}}  & $2$ & $12$ & $12000$  \\
% \textit{Mixture}\footnotesize{~\cite{Mixture}} & $2$ & $8$ & $12000$  \\
% \textit{YELP}\footnotesize{~\cite{YELP0,YELP}}  &$3$ & $8$ & $12000$  \\
% \textit{FYelp}\footnotesize{~\cite{Fyelp}} &$4$ & $40$ & $12000$   \\
% \hline 
% \end{tabular}
% }
% \caption{
% Component datasets in our CompMCTG benchmark: $m$ refers to the number of aspects (e.g., sentiment, topic, tense, gender and stuff); $|\mathcal{C}|$ denotes the total number of attribute combinations; size means the number of text data in the dataset.
% }
% \label{tab:dataset_copy}
% \end{table}

\subsection{Baseline and Evaluation Metric}
We select eight representative baseline methods to study: 1) for \textbf{Joint-Training based} methods, we choose \textit{CTRL}~\cite{keskar2019ctrl}, a classic and powerful baseline, \textit{Contrastive Prefix} (\textit{Con.Prefix})~\cite{contrastive-prefix}, \textit{CatPrompt}~\cite{yang-etal-2023-tailor}, and \textit{DCG}~\cite{seen-to-unseen-acl2023}, a related work targeting on reducing the compositionality gap, as our baseline methods, 2) for \textbf{Seperate-Training based}, we select two state-of-the-art baselines: \textit{Distribution-Lens}~\cite{gu-etal-2022-distributional} and \textit{Prior}~\cite{gu-etal-2023-controllable}, 3) for \textbf{Decoding-Time based} methods, we choose \textit{PPLM}~\cite{dathathri2019plug} and \textit{Fudge}~\cite{yang-klein-2021-fudge}. In addition, we adopt \textit{LLaMA-2}~\cite{llama2} and \textit{ChatGPT}~\cite{openaiChatGPT} to study the compositional generalization of large language models (LLMs) with In-context Learning paradigm~\cite{brown2020language}. Following ~\cite{evaluating_llms}, we attach five demonstrations in the input prompt for LLMs to follow. One can find more details about our implementations in Appendix~\ref{sec:Implementation details}. \par
Grounded on the MCTG task, we adopt the evaluation metrics (note that the subfixes ``\textit{i.d.}" and ``\textit{comp}" refer to the in-distribution and compositional testing respectively.) of 1) \textit{ACC}\textsubscript{\textit{i.d.}} and \textit{ACC}\textsubscript{\textit{comp}}: the averaged prediction accuracies\footnote{For each aspect in each dataset, we train a Roberta classifier~\cite{liu2019roberta} to evaluate its accuracy (details in Appendix ~\ref{sec:hyperparameters-classifier}).} for all of the control aspects to measure the \textbf{controllability} of generated text, 2) \textit{PPL}\textsubscript{\textit{i.d.}} and \textit{PPL}\textsubscript{\textit{comp}}: perplexity calculated by GPT-2 Large to measure the \textbf{fluency} of generated text in all of our experiments,
% gap will introduced in the analysis section.
and 3) \textit{Dist-3}: 3-gram distinctness to evaluate the \textbf{diversity} of the text generated by approaches mentioned above. We also adopt \textbf{\textit{Human-evaluation}} to measure the relevance and fluency of the generated text for each approach\footnote{Due to the page limit, please find the result of \textit{Dist-3} and \textit{Human-evaluation} in Appendix~\ref{sec:Experiments of diversity} and ~\ref{human evaluation}.}.  
\label{sec3:baseline}
\subsection{Evaluation Result}
\label{sec3:evaluation}
The main evaluation results on CompMCTG benchmark are shown in Table~\ref{tab:compmctg}, where values in ``\textit{Original}” column refer the performance where text data of all attribute combinations are available in the training set and hence there is no compositional testing; values in ``\textit{Hold-Out}” and ``\textit{ACD}” columns refer to in-distribution and compositional testing performance through the evaluation protocols of ``\textit{Hold-Out}” and ``\textit{ACD}” mentioned in Section~\ref{sec3:construction} respectively; values in ``\textit{A}\textsubscript{\textit{avg}}” and ``\textit{P}\textsubscript{\textit{avg}}” column refer to overall performance which is the arithmetic mean of results under different evaluation protocols mentioned here (\textit{Original}\textsubscript{\textit{i.d.}}, \textit{Hold-Out}\textsubscript{\textit{i.d.}}, \textit{Hold-Out}\textsubscript{\textit{comp}}, \textit{ACD}\textsubscript{\textit{i.d.}} and \textit{ACD}\textsubscript{\textit{comp}}), which are formulated as:

\scriptsize
\begin{equation}
\label{eq: AP_avg}
\begin{aligned}
    &A_{avg} = \frac{1}{5}(A_{i.d.}^{original}+A_{i.d.}^{holdout}+A_{comp}^{holdout}+A_{i.d.}^{acd}+A_{comp}^{acd})\\
    &P_{avg}=\frac{1}{5}(P_{i.d.}^{original}+P_{i.d.}^{holdout}+P_{comp}^{holdout}+P_{i.d.}^{acd}+P_{comp}^{acd})
\end{aligned}
\end{equation}
\normalsize

The ``gap” (\textit{G}\textsubscript{\textit{avg}}) is used to assess the average compositional generalization risk and a lower \textit{G}\textsubscript{\textit{avg}} indicates better robustness under compositional testing, which is formulated as:

\small
\begin{equation}
\label{eq: G_avg}
\begin{aligned}
    G_{avg} &= \frac{1}{2}(G_{holdout}+G_{acd}) \\
    &=\frac{1}{2}(\frac{A_{i.d.}^{holdout}-A_{comp}^{holdout}}{A_{i.d.}^{holdout}}+ \frac{A_{i.d.}^{acd}-A_{comp}^{acd}}{A_{i.d.}^{acd}})
\end{aligned}
\end{equation}
\normalsize
Among all the evaluated baselines, \textbf{joint-training-based} approaches generally exhibit higher attribute accuracy, better fluency (lower perplexity, only inferior to LLM+ICL), and better robustness to compositional testing (lower \textit{G}\textsubscript{\textit{avg}}).
% Nonetheless, We observe that there is a noticeable performance gap between in-distribution testing and compositional testing for joint-training-based approaches ($9.6$\% on average).
% discuss Dis-Lens
Though seperate-training-based methods perform acceptably in in-distribution testing, their performance drops drastically in compositional testing and we discuss the inherent reason for their failures in Appendix~\ref{appendix:seperate training}.
Decoding-time-based methods perform poorly overall, despite PPLM owning the lowest \textit{G}\textsubscript{\textit{avg}}, both its average accuracy and perplexity are unusable.
LLMs can generate more fluent text while the controllability of the generated text ($54.04$\%) falls behind joint-training-based methods ($79.82$\%). At the same time, LLMs (+ICL) also suffer from a large performance drop in compositional testing ($G_{avg}$ is $23.5$\% for LLaMA and ChatGPT).

Additionally, We evaluate all of the baseline approaches with \textit{Few-Shot} evaluation protocol in Table~\ref{tab:few-shot}, to reflect their performance when only limited attribute combinations are available. Again, \textbf{joint-training-based} approaches hold the best average performance and compositional generalization capacity among them. We provide the details of our benchmark in Appendix~\ref{appendix:benchmark}.

\subsection{Insight}
In this section, we conduct analysis experiments to show the effect of our key designs in CompMCTG: 1) the three-dimensional evaluation protocol (\textit{Hold-Out}, \textit{ACD} and \textit{Few-Shot}) and 2) the effectiveness of \textit{ACD} in amplifying the compositional generalization gap.
\label{sec3:analysis}
\paragraph{Compositional gaps with different evaluation protocols.}
\begin{figure}[ht]
    \raggedright
    \includegraphics[width=1.0\linewidth, keepaspectratio=true]{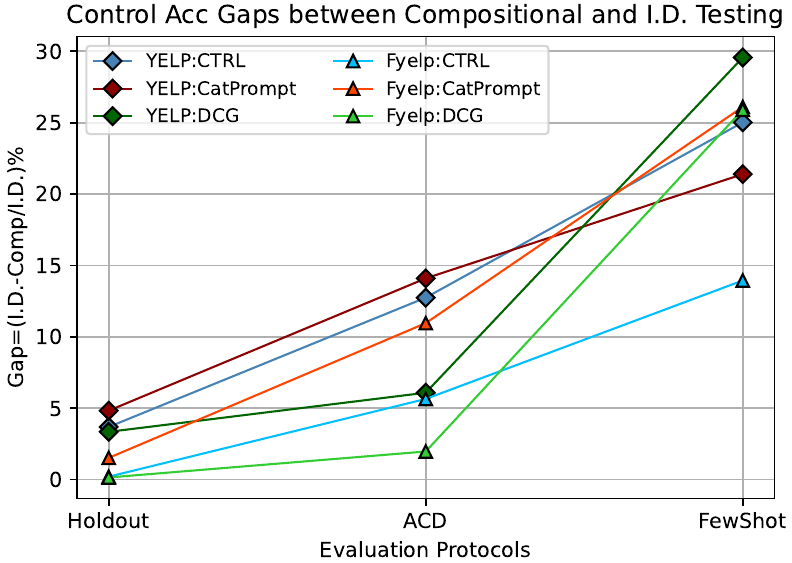}
    \caption{Compositional generalization gap with different evaluation protocols.}
    \label{fig:protocols}
\end{figure}
In Figure~\ref{fig:protocols}, we show compositional gaps ($G = \frac{A_{i.d.}-A_{comp}}{A_{i.d.}}$) for approaches: \textit{CTRL}, \textit{CatPrompt} and \textit{DCG}, with three evaluation protocols on \textit{YELP} and \textit{Fyelp} datasets. We observe that the compositional gaps on the same approach and dataset vary a lot with different evaluation protocols: $G_{holdout}< G_{acd} < G_{fewshot}$ generally holds. Notably, \textit{Hold-Out} can not properly unveil the compositional generalization gap for a specific approach. For instance: On \textit{Fyelp} dataset, \textit{CatPrompt} has the compositional gap of $0.91$\% on \textit{Hold-Out} protocol, while it drastically increases to $10.96$\% on \textit{ACD} protocol. Moreover, different approaches have different preferences for these protocols. By way of example, The compositional gap (e.g., on \textit{Fyelp}) of \textit{DCG} with \textit{ACD} ($1.97$\%) is lower than \textit{CTRL} ($5.95$\%) while its gap with \textit{Few-Shot} ($25.91$\%) is much higher than \textit{CTRL} ($13.95$\%), demonstrating that the deficiency of \textit{DCG} in low-data regime.
Hence jointly leveraging these three evaluation protocols evaluates MCTG approaches more comprehensively.
\paragraph{Does the ACD better unveil the compositional generalization risk in comparison with Random Sampling?}
\begin{figure}[ht]
    \raggedright
    \includegraphics[width=1.0\linewidth, keepaspectratio=true]{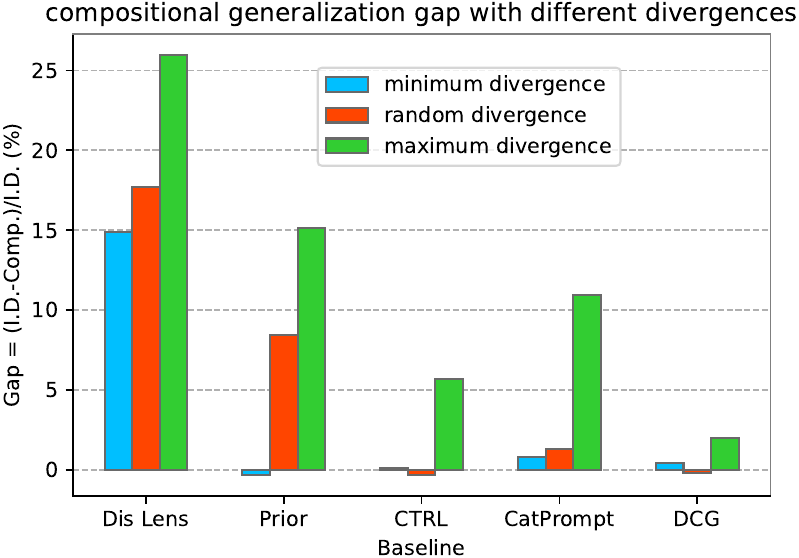}
    \caption{Comparison of compositional gaps between \textit{ACD} (green bars) and two other splitting methods: \textit{Random Sampling} (red bars) and \textit{minimizing the divergence} (blue bars) on five baselines.}
    \label{fig:divergence}
\end{figure}
% The key idea of \textit{ACD} is to \textbf{maximize} the divergence of \textit{attribute compound distributions} between in-distribution and compositional sets. 
To demonstrate the effectiveness of \textit{ACD}, where we maximize the divergence of \textit{attribute compound distributions} between in-distribution and compositional sets, we design two other protocols in which we still keep $|\mathcal C_{i.d.}|=|\mathcal C_{comp}|$: \textit{Random Sampling} (random divergence) and \textit{minimizing the divergence} (minimum divergence).
We compare the compositional gaps among the three protocols (on \textit{Fyelp} dataset) in Figure~\ref{fig:divergence}.
% we compare the compositional gaps (on \textit{Fyelp}) with \textit{ACD} and two other ``Half \& Half” splitting methods: \textit{Random Sampling} and \textit{minimizing the divergence} in Figure~\ref{fig:divergence}.
We observe that gaps of \textit{ACD} are consistently higher than two comparison protocols by large margins. Notably, using baseline approaches of \textit{CTRL} and \textit{DCG}, compositional gaps with \textit{Random Sampling} are \textbf{near zero} while they are $5.65$\% and $1.97$\% with \textit{ACD}. Hence we conclude that \textit{ACD} generally better unveils the compositional generalization risk while \textit{Random Sampling} often causes gross under-estimation of such risk.
\section{Methodlogy: Meta-MCTG}
\label{sec4}
\begin{figure*}[ht]
    \raggedright
    \includegraphics[width=1.0\linewidth, keepaspectratio=true]{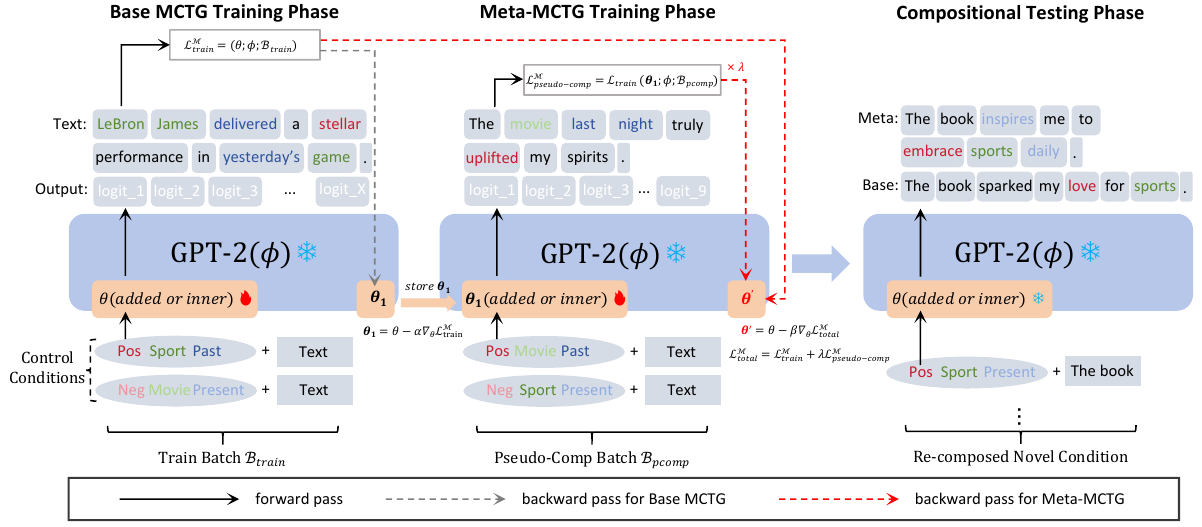}
    \caption{Meta-MCTG: $\theta$ refers to the learnable parameters for encoding control conditions, which could be inner (\textit{CTRL}) or added (\textit{DCG} and \textit{ContraPrefix}). $\phi$, the parameters of LMs, are usually frozen during training (PEFT).}
    \label{fig:metamctg}
\end{figure*}
%Most existing MCTG methods perform well on in-domain combinations. However, they struggle to achieve satisfactory results on compositional combinations. To address this issue, we propose a meta-learning-based approach to improve the performance of existing MCTG methods on compositional combinations. 
In Section~\ref{sec3:analysis}, we observe that joint-training-based (both parameter-efficient fine-tuning based and all-parameter fine-tuning based) baselines generally achieve better overall performance. Nonetheless, there still exist non-negligible compositional generalization gaps for all these baselines, which highly calls for our attention.
To this end, we propose Meta-MCTG, a novel \textbf{Meta}-learning~\cite{maml-icml2017} based \textbf{MCTG} training framework, to further improve compositional generalization capabilities of existing joint-training baselines. The framework is easy to implement and can be directly combined with any joint-training-based methods.
We discuss the design of Meta-MCTG in Section~\ref{sec4:design} and demonstrate its effectiveness through experiment results for Meta-MCTG in combination with three competitive joint-training baselines (CTRL~\cite{keskar2019ctrl}, ContrastivePrefix~\cite{contrastive-prefix} and DCG~\cite{seen-to-unseen-acl2023}) in Section~\ref{sec4:experiment}.
\subsection{Design}
%$\mathcal{L}_{total} = \mathcal{L}_{train} + \lambda \times \mathcal{L}_{p-comp}$
\label{sec4:design}
%Inspired by the meta-learning-based training used in previous work~\cite{conklin-etal-2021-meta}, we aim to train the MCTG models on in-domain combinations to learn knowledge about how to better generalize on compositional combinations.
% MAML~\cite{maml-icml2017}, a popular optimization-based meta-learning method, has been broadly adopted to help deep network generalize~\cite{Li2018MLDG,wang-etal-2021-meta,conklin-etal-2021-meta}.
\paragraph{Overall Motivation}
The overall framework of Meta-MCTG is depicted in Figure~\ref{fig:metamctg}.
We analyze that the failure of generating text satisfying control conditions in compositional testing can be attributed to the over-fitting of language models to local optima of control conditions in the training set. Thus when trained language models are fed with recomposed attribute combinations as the control conditions in the compositional testing (e.g., In Figure~\ref{fig:metamctg}, ``positive-sport-\textbf{present}”), it will potentially encode and distribute those new attribute combinations in the neighbor area of similar ones (e.g., ``positive-sport-\textbf{past}”) that they have seen in the training phase. In this way, previous MCTG approaches fail to generate text that perfectly meets the requirements of all given conditions. As depicted in Figure~\ref{fig:metamctg}, when given the recomposed attribute combination of ``positive-sport-\textbf{present}”, models may generate text like ``The book \textbf{sparked} my love for sports.”, neglecting the ``\textbf{present}” condition (As models only sees ``positive-sport-\textbf{past}” attribute combination in the training phase).

% Inspired by previous meta-learning works targeting generalization~\cite{Li2018MLDG,wang-etal-2021-meta,conklin-etal-2021-meta}, we aim to leverage Model-Agnostic Meta Learning (MAML)~\cite{maml-icml2017} to mitigate the above overfitting problem. Given a batch of training data $\mathcal{B}_{train}=(c_i^{train}, x_i^{train})_{i=1}^m$ (control conditions $c_i^{train}$ and their corresponding text $x_i^{train}$, $m$ refers to the batch size.), we create a batch of ``pseudo-comp” data $\mathcal{B}_{pseudo-comp}=(c_i^{pcomp}, x_i^{pcomp})_{i=1}^m$ selecting from the whole \textbf{training set} $\mathcal{D}_{train}$, where for each pseudo-comp datum, its control condition must be the novel recombination of attributes appearing in the training batch. For instance, in Figure~\ref{fig:metamctg} the pseudo-comp conditions ``\textcolor{blue}{positive}-\textcolor{red}{movie}-\textcolor{blue}{past}” and ``\textcolor{red}{negative}-\textcolor{blue}{sport}-\textcolor{red}{present}” are the recombinations of conditions ``\textcolor{blue}{positive}-\textcolor{blue}{sport}-\textcolor{blue}{past}” and ``\textcolor{red}{negative}-\textcolor{red}{movie}-\textcolor{red}{present}” in the training batch. The pseudo-comp data can be used for mimicking the compositional testing and play a role in navigating the training of models away from the local optima that overfits training conditions.
\paragraph{Meta-MCTG training procedure}
% Here we formalize the whole training procedure of Meta-MCTG. 
Inspired by previous meta-learning works targeting generalization~\cite{Li2018MLDG,wang-etal-2021-meta,conklin-etal-2021-meta}, we aim to leverage Model-Agnostic Meta Learning (MAML)~\cite{maml-icml2017} to mitigate the overfitting problem. 

First of all, given a specific joint-training-based approach $\mathcal{M}$, we denote its training objective as $\mathcal{L}_{train}^\mathcal{M}(\theta;\phi;\mathcal{B})$ where $\theta$ represents the learnable parameters of encoding control conditions, $\phi$ represents the parameters of the language model (e.g., GPT-2), which are frozen during training (Note that in CTRL, $\phi$ is also updated while it still suits for the Meta-MCTG.), and $\mathcal B$ denotes a batch of data. In general, the training objective can be derived as:

\small
\begin{equation} 
\label{obj:l_train}
  \begin{split}
    \min\limits_\theta &\ \mathcal{L}_{train}^\mathcal{M}(\theta;\phi;\mathcal{B}) = \\ 
    \min\limits_\theta &\sum\limits_{(c_i,x_i)\in \mathcal{B}}[-\log p(x_i|c_i;\theta;\phi)] + \mathcal{L}_\mathcal{M}(\theta;\phi;\mathcal{B})
  \end{split}
\end{equation}
\normalsize
The first term refers to the basic LM loss~\cite{radford2018improving} which maximizes the likelihood of generating target text $x_i$ and the second term refers to the auxiliary loss added by baseline $\mathcal{M}$ (e.g., contrastive loss~\cite{contrastive-prefix}).

In the Meta-MCTG framework, we first sample a batch of training data, denoted as $\mathcal{B}_{train}=(c_i^{train}, x_i^{train})_{i=1}^m$ and a batch of pseudo-comp data, denoted as $\mathcal{B}_{pcomp}=(c_i^{pcomp}, x_i^{pcomp})_{i=1}^m$ where $\{c_i^{train}\}_{i=1}^m \cap \{c_i^{pcomp}\}_{i=1}^m = \emptyset$ and each attribute combination of $\{c_i^{pcomp}\}_{i=1}^m$ must be the recombination of single attributes appearing in the $\{c_i^{train}\}_{i=1}^m$. 
For instance, in Figure~\ref{fig:metamctg} the pseudo-comp conditions ``\textcolor{blue}{positive}-\textcolor{red}{movie}-\textcolor{blue}{past}” and ``\textcolor{red}{negative}-\textcolor{blue}{sport}-\textcolor{red}{present}” are the recombinations of conditions ``\textcolor{blue}{positive}-\textcolor{blue}{sport}-\textcolor{blue}{past}” and ``\textcolor{red}{negative}-\textcolor{red}{movie}-\textcolor{red}{present}” in the training batch.

% Given a specific baseline approach, we denote its overall loss function as  $\mathcal{L}(x;\theta)$, where $\theta$ represents the model's learnable parameters and $x$ is a training sample. During the training phase, we denote the data of each training batch as $X_{train}=\{(x^t_i)_{1\le i \le N}\}$, and the combination of attribute labels it contains as $\mathcal C_{train}=\{(c^t_j)_{1\le j\le N_t}\}, c^t_j\in \mathcal C_{i.d.}$. Based on $\mathcal C_{train}$, we sample a support batch $X_{support}=\{(x^s _i)_{1\le i\le N}\}$ from the training set, which contains the combination of attribute labels $\mathcal C_{support}=\{(c^s_j)_{1\le j\le N_s}\}, c^s_j\in \mathcal C_{i.d.}$, where $N$ represents the size of a training batch, $N_t$ represents the number of combinations in the training batch, and $N_s$ represents the number of combinations in the support batch. The relationship between $\mathcal C_{train}$ and $\mathcal C_{support}$ is defined by the following equation:
% \begin{align*}
%      & \mathcal C_{train}\in \mathcal C_{i.d.}, \mathcal C_{support}\in \mathcal C_{i.d.}\\
%      & \mathcal C_{train}\cap \mathcal C_{support} = \emptyset \\
%      & \{attribute|\exists c\in \mathcal C_{support}, attribute\in c\} \subseteq \\
%      & \{attribute|\exists c\in \mathcal C_{train}, attribute\in c\}
% \end{align*}
We train model on $\mathcal B_{train}$ and perform one step of gradient descent to update $\theta$ with Objective~\ref{obj:l_train} ($\alpha$ is the learning-rate):
% \begin{align}
%     \mathcal L_{train}(\theta) = \frac{1}{N}\sum_{i=1}^N\mathcal L(x^t_i; \theta) \label{eq:train_loss}
% \end{align}

\small
\begin{align}
\label{eq:update_first}
    \theta_1 = \theta - \alpha\nabla_\theta \mathcal{L}_{train}^\mathcal{M}(\theta;\phi;\mathcal{B}_{train})
\end{align}
\normalsize
Then we maintain $\theta$ unchanged in the original model, temporarily store $\theta_1$ to a backup model, and feed $\mathcal{B}_{pcomp}$ to the backup model to obtain the loss on pseudo-comp data:

\small
\begin{equation} 
\label{obj:pseudo-comp}
  \begin{split}
    &\mathcal{L}_{pseudo-comp}^\mathcal{M}(\theta;\phi;\mathcal{B}_{pcomp}) = \mathcal{L}_{train}^\mathcal{M}(\theta_1;\phi;\mathcal{B}_{pcomp})\\
    &=\mathcal{L}_{train}^\mathcal{M}(\theta-\alpha\nabla_\theta \mathcal{L}_{train}^\mathcal{M}(\theta;\phi;\mathcal{B}_{train});\phi;\mathcal{B}_{pcomp})
  \end{split}
\end{equation}
\normalsize
According to the construction of $\mathcal{B}_{pcomp}$, we use $\mathcal{L}_{pseudo-comp}^\mathcal{M}(\theta;\phi;\mathcal{B}_{pcomp})$ to simulate the compositional generalization scenario, evaluating the compositional generalization capacity of model updated by Eq~\ref{eq:update_first}. We hope the updated model (with $\theta_1$) performs as well as possible on these pseudo-comp data rather than merely overfitting $\mathcal{B}_{train}$. Taking both the original training Objective~\ref{obj:l_train} and the compositional generalization Objective~\ref{obj:pseudo-comp}
into consideration, Meta-MCTG is to minimize the following objective:

\small
\begin{equation} 
\label{obj:total}
  \begin{split}
    &\mathcal{L}_{total}^\mathcal{M}(\theta;\phi;\mathcal{B}_{train};\mathcal{B}_{pcomp}) =\\
    & \mathcal{L}_{train}^\mathcal{M}(\theta;\phi;\mathcal{B}_{train}) + \lambda\mathcal{L}_{pseudo-comp}^\mathcal{M}(\theta;\phi;\mathcal{B}_{pcomp})\\
  \end{split}
\end{equation}
\normalsize
Where $\lambda$ is a hyper-parameter to make a trade-off between the above two terms.
% Next, we store the updated parameters $\theta_1$ into a backup model and use it to train the support batch, obtaining the support loss $\mathcal L_{support}$. The total loss of the model is the sum of the training loss and the support loss, which is denoted as Eq.\ref{eq:support_loss} and Eq.\ref{eq:total_loss}.
% \begin{align}
%     \mathcal L_{support}(\theta) = \frac{1}{N}\sum_{i=1}^N\mathcal L(x^s_i; \theta_1) \label{eq:support_loss}
% \end{align}
% \begin{align}
%     \mathcal L_{total}(\theta) = \mathcal L_{train}(\theta) + \lambda\mathcal L_{support}(\theta) \label{eq:total_loss}
% \end{align}
Finally, we perform one step of gradient descent to update $\theta$ in the original model with Objective~\ref{obj:total}:

\small
\begin{align}
\label{eq:update_final}
    \theta^\prime = \theta - \beta\nabla_\theta \mathcal{L}_{total}^\mathcal{M}(\theta;\phi;\mathcal{B}_{train};\mathcal{B}_{pcomp}) 
\end{align}
\normalsize
Where $\beta$ is the learning rate. We summarize the pseudo-code of the Meta-MCTG training procedure in Algorithm~\ref{alg:meta-learning} in Appendix~\ref{sec:appendix-pseduocode}. 
\subsection{Experiment Results and Analysis}
\label{sec4:experiment}

\begin{table*}[t]
\centering
\resizebox{\textwidth}{!}{
\begin{tabular}{lcc|cc|cc|cc|cc|cc}
\hline
\multirow{3}{*}{\textbf{Method}}  & \multicolumn{4}{c}{\textbf{Fyelp}} & \multicolumn{2}{c}{\textbf{Amazon}} & \multicolumn{4}{c}{\textbf{YELP}}  &\multicolumn{2}{c}{\textbf{Mixture}} \\
 & \multicolumn{2}{c}{\textit{Hold-Out}}& \multicolumn{2}{c}{\textit{ACD}}
 & \multicolumn{2}{c}{\textit{Hold-Out}} & \multicolumn{2}{c}{\textit{Hold-Out}}& \multicolumn{2}{c}{\textit{ACD}}
  & \multicolumn{2}{c}{\textit{Hold-Out}}\\
 &  \textit{A}\textsubscript{\textit{comp}}$(\uparrow)$ &  \textit{P}\textsubscript{\textit{comp}}$(\downarrow)$
 & \textit{A}\textsubscript{\textit{comp}} $(\uparrow)$&  \textit{P}\textsubscript{\textit{comp}}$(\downarrow)$
  &  \textit{A}\textsubscript{\textit{comp}} $(\uparrow)$&  \textit{P}\textsubscript{\textit{comp}}$(\downarrow)$
   &  \textit{A}\textsubscript{\textit{comp}}$(\uparrow)$ &  \textit{P}\textsubscript{\textit{comp}}$(\downarrow)$
 & \textit{A}\textsubscript{\textit{comp}} $(\uparrow)$&  \textit{P}\textsubscript{\textit{comp}}$(\downarrow)$
  &  \textit{A}\textsubscript{\textit{comp}} $(\uparrow)$&  \textit{P}\textsubscript{\textit{comp}}$(\downarrow)$\\
 \hline
\hline
\textit{CTRL}\footnotesize{~\cite{keskar2019ctrl}}  &$68.29$\%  & $45.61$ &$65.31$\% &$45.86$  &$77.89$\%&$37.02$& $82.02$\% &$73.74$  &$74.63$\% &$75.46$ &$71.82$\% & $47.46$  \\
\textit{Meta-CTRL} \footnotesize{\textbf{(Ours)}}  &$\mathbf{68.69}$\textbf{\%}  & $46.42$ &$\mathbf{65.77}$\textbf{\%} &$46.01$  &$\mathbf{78.78}$\textbf{\%}&$37.30$& $\mathbf{83.85}$\textbf{\%} &$68.94$  &$\mathbf{78.27}$\textbf{\%} &$78.11$ &$\mathbf{72.83}$\textbf{\%} & $46.20$ \\
\hline
\textit{Con.Prefix}\footnotesize{~\cite{contrastive-prefix}} &$67.50$\%& $52.32$ &$63.93$\%&$49.78$ &$87.58$\%& $44.36$ &$92.79$\% &$132.21$  &$88.84$\% &$128.87$ &$71.91$\% & $138.93$ \\
\textit{Meta-Con.Prefix} \footnotesize{\textbf{(Ours)}} &$\mathbf{67.75}$\textbf{\%}& $52.62$ &$\mathbf{64.06}$\textbf{\%}&$49.12$ &$\mathbf{87.69}$\textbf{\%}& $43.89$ &$\mathbf{94.06}$\textbf{\%} &$130.66$  &$\mathbf{90.40}$\textbf{\%} &$132.19$ &$\mathbf{73.11}$\textbf{\%} & $140.53$  \\
\hline
\textit{DCG}\footnotesize{~\cite{seen-to-unseen-acl2023}} &$66.39$\%& $53.52$ &$64.71$\%&$53.67$ &$84.51$\%& $47.09$ &$80.61$\% &$69.87$  &$75.72$\% &$82.08$ &$76.32$\% & $71.20$\\
\textit{Meta-DCG} \footnotesize{\textbf{(Ours)}} &$66.36$\%& $53.04$ &$\mathbf{64.84}$\textbf{\%}&$53.58$ &$\mathbf{85.11}$\textbf{\%}& $47.77$ &$\mathbf{81.15}$\textbf{\%} &$72.32$  &$\mathbf{75.88}$\textbf{\%} &$84.58$ &$\mathbf{79.15}$\textbf{\%} & $65.68$ \\
\hline 
\end{tabular}
}
\caption{
Experiment results of \textit{CTRL}, \textit{ContraPrefix}, and \textit{DCG} with Meta-MCTG training in compositional testing.
}
\label{tab:meta-mctg-comp}
\end{table*}

\begin{table*}[t]
\centering
\resizebox{\textwidth}{!}{
\begin{tabular}{lcc|cc|cc|cc|cc|cc}
\hline
\multirow{3}{*}{\textbf{Method}}  & \multicolumn{4}{c}{\textbf{Fyelp}} & \multicolumn{2}{c}{\textbf{Amazon}} & \multicolumn{4}{c}{\textbf{YELP}}  &\multicolumn{2}{c}{\textbf{Mixture}} \\
 & \multicolumn{2}{c}{\textit{Hold-Out}}& \multicolumn{2}{c}{\textit{ACD}}
 & \multicolumn{2}{c}{\textit{Hold-Out}} & \multicolumn{2}{c}{\textit{Hold-Out}}& \multicolumn{2}{c}{\textit{ACD}}
  & \multicolumn{2}{c}{\textit{Hold-Out}}\\
 &  \textit{A}\textsubscript{\textit{i.d.}}$(\uparrow)$ &  \textit{P}\textsubscript{\textit{i.d.}}$(\downarrow)$
 & \textit{A}\textsubscript{\textit{comp}} $(\uparrow)$&  \textit{P}\textsubscript{\textit{i.d.}}$(\downarrow)$
  &  \textit{A}\textsubscript{\textit{i.d.}} $(\uparrow)$&  \textit{P}\textsubscript{\textit{i.d.}}$(\downarrow)$
   &  \textit{A}\textsubscript{\textit{i.d.}}$(\uparrow)$ &  \textit{P}\textsubscript{\textit{i.d.}}$(\downarrow)$
 & \textit{A}\textsubscript{\textit{i.d.}} $(\uparrow)$&  \textit{P}\textsubscript{\textit{i.d.}}$(\downarrow)$
  &  \textit{A}\textsubscript{\textit{i.d.}} $(\uparrow)$&  \textit{P}\textsubscript{\textit{i.d.}}$(\downarrow)$\\
 \hline
\hline
\textit{CTRL}\footnotesize{~\cite{keskar2019ctrl}}  &$69.43$\%  & $45.95$ &$69.22$\% &$45.60$  &$80.52$\%&$37.43$& $85.16$\% &$72.20$  &$85.52$\% &$76.06$ &$80.56$\% & $48.82$  \\
\textit{Meta-CTRL} \footnotesize{\textbf{(Ours)}}  &$\mathbf{69.51}$\textbf{\%}  & $46.16$ &$\mathbf{69.45}$\textbf{\%} &$45.50$  &$80.26$\%&$37.31$& $\mathbf{85.76}$\textbf{\%} &$69.05$  &$\mathbf{86.11}$\textbf{\%} &$70.95$ &$80.08$\% & $46.42$ \\
\hline
\textit{Con.Prefix}\footnotesize{~\cite{contrastive-prefix}} &$67.84$\%& $52.48$ &$63.40$\%&$53.11$ &$87.56$\%& $43.97$ &$94.40$\% &$136.04$  &$91.82$\% &$141.15$ &$83.88$\% & $96.46$ \\
\textit{Meta-Con.Prefix} \footnotesize{\textbf{(Ours)}} &$\mathbf{67.90}$\textbf{\%}& $52.40$ &$\mathbf{64.19}$\textbf{\%}&$52.84$ &$87.43$\%& $43.93$ &$\mathbf{94.42}$\textbf{\%} &$136.42$  &$\mathbf{91.86}$\textbf{\%} &$136.39$ &$\mathbf{84.24}$\textbf{\%} & $97.66$  \\
\hline
\textit{DCG}\footnotesize{~\cite{seen-to-unseen-acl2023}} &$66.49$\%& $53.50$ &$66.01$\%&$53.29$ &$84.71$\%& $47.20$ &$82.43$\% &$70.28$  &$80.12$\% &$82.96$ &$83.69$\% & $91.80$\\
\textit{Meta-DCG} \footnotesize{\textbf{(Ours)}} &$\mathbf{66.50}$\textbf{\%}& $53.16$ &$\mathbf{66.23}$\textbf{\%}&$52.92$ &$\mathbf{84.78}$\textbf{\%}& $47.55$ &$82.07$\% &$70.01$  &$\mathbf{80.57}$\textbf{\%} &$82.04$ &$83.50$\% & $83.39$ \\
\hline 
\end{tabular}
}
\caption{
Experiment results of \textit{CTRL}, \textit{ContraPrefix} and \textit{DCG} with Meta-MCTG training in in-distribution testing.
}
\label{tab:meta-mctg-id}
\end{table*}

\begin{table}[t]
\centering
\resizebox{0.5\textwidth}{!}{
\begin{tabular}{lc|c|c|c|c|c}
\hline
\multirow{2}{*}{\textbf{Method}}  & \multicolumn{2}{c}{\textbf{Fyelp}} & \multicolumn{1}{c}{\textbf{Amazon}} & \multicolumn{2}{c}{\textbf{YELP}}  &\multicolumn{1}{c}{\textbf{Mixture}} \\
 & \multicolumn{1}{c}{\textit{Hold-Out}}& \multicolumn{1}{c}{\textit{ACD}}
 & \multicolumn{1}{c}{\textit{Hold-Out}} & \multicolumn{1}{c}{\textit{Hold-Out}}& \multicolumn{1}{c}{\textit{ACD}}
  & \multicolumn{1}{c}{\textit{Hold-Out}}\\
 \hline
\hline
\textit{CTRL}\footnotesize{~\cite{keskar2019ctrl}}  &$1.64$\% &$5.65$\% &$3.27$\% &$3.69$\% &$12.73$\% &$10.85$\% \\
\textit{Meta-CTRL} \footnotesize{\textbf{(Ours)}}  &$\mathbf{0.89}$\textbf{\%} &$\mathbf{5.3}$\textbf{\%} &$\mathbf{1.84}$\textbf{\%} &$\mathbf{2.23}$\textbf{\%} &$\mathbf{9.1}$\textbf{\%} &$\mathbf{9.05}$\textbf{\%} \\
\hline
\textit{Con.Prefix}\footnotesize{~\cite{contrastive-prefix}} &$0.5$\% &$-0.84$\% &$-0.02$\% &$1.71$\% &$3.25$\% &$14.27$\% \\
\textit{Meta-Con.Prefix} \footnotesize{\textbf{(Ours)}} &$\mathbf{0.22}$\textbf{\%} &$0.2$\% &$\mathbf{-0.3}$\textbf{\%} &$\mathbf{0.38}$\textbf{\%} &$\mathbf{1.59}$\textbf{\%} &$\mathbf{13.21}$\textbf{\%} \\
\hline
\textit{DCG}\footnotesize{~\cite{seen-to-unseen-acl2023}} &$0.15$\% &$1.97$\% &$0.24$\% &$2.21$\% &$5.49$\% &$8.81$\% \\
\textit{Meta-DCG} \footnotesize{\textbf{(Ours)}} &$0.21$\% &$2.1$\% &$\mathbf{-0.39}$\textbf{\%} &$\mathbf{1.12}$\textbf{\%} &$5.82$\% &$\mathbf{5.21}$\textbf{\%} \\
\hline 
\end{tabular}
}
\caption{
Compositional generalization gap of \textit{CTRL}, \textit{ContraPrefix} and \textit{DCG} with Meta-MCTG training.
}
\label{tab:meta-mctg-gap}
\end{table}

\paragraph{Experiment Results of Meta-MCTG}
We train \textit{CTRL}, \textit{ContrastivePrefix} and \textit{DCG} with the Meta-MCTG algorithm and aim to demonstrate that Meta-MCTG can generally improve their compositional generalization capacity. The compositional testing results for all four datasets are shown in Table~\ref{tab:meta-mctg-comp}\footnote{We do not apply Meta-MCTG to \textit{Few-Shot} settings, for we can not construct $\mathcal{B}_{pseudo-comp}$ when each attribute only appears once in $\mathcal{C}_{i.d.}$.}. For most cases ($94.4$\% of the total), we can observe that baseline approaches trained with Meta-MCTG have an obvious improvement in compositional testing performance on controllability of generated text (i.e., attribute accuracy) over the original versions (by at most $3.64$\%).
Besides, the introduction of the Meta-MCTG framework has almost no impact on text fluency (i.e., perplexity).
We additionally show the in-distribution testing results in Table~\ref{tab:meta-mctg-id}, demonstrating that Meta-MCTG nearly has no negative effect on in-distribution testing. Instead, it improves the in-distribution testing over the original baselines on $72.2$\% cases. We also provide a separate analysis of the compositional generalization gap variations for each dataset and protocol before and after incorporating the Meta-MCTG framework in Table~\ref{tab:meta-mctg-gap}, where the gap is calculated by $gap=\frac{A_{i.d.}-A_{comp}}{A_{i.d.}}$. From the results, it can be observed that in the majority of cases, the Meta-MCTG framework is able to reduce the compositional generalization gap.

\begin{figure}[ht]
    \raggedright
    \includegraphics[width=0.99\linewidth, keepaspectratio=true]{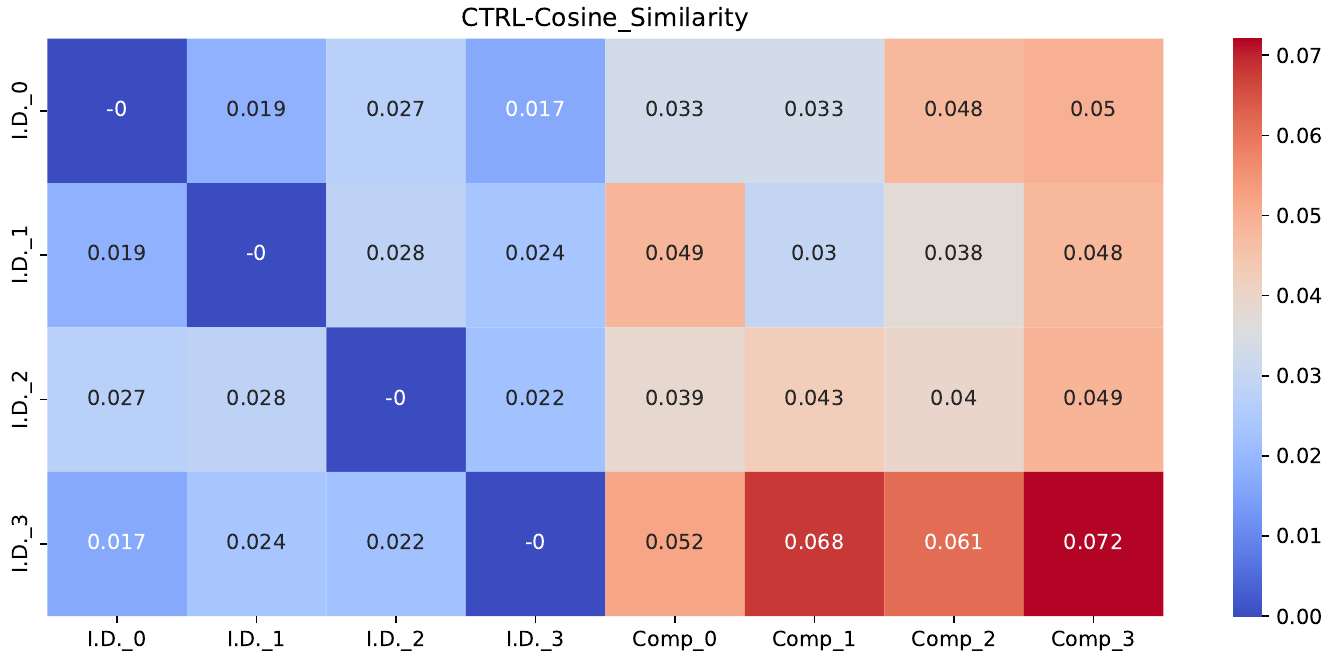}
    \caption{Difference of the distances ($d = 1 - cos<h_1, h_2>$) between attribute combinations in the representation space ($h_1, h_2$) with Meta-CTRL and the origin version of CTRL.}
    \label{fig:analysis-meta}
\end{figure}

\paragraph{Visualization and Case Study}
Previously we hypothesize that Meta-MCTG mitigates the problem that overfitted baseline approaches distribute recomposed novel attribute combinations in the neighbor of in-distribution ones in the representation space. We now calculate the difference in the distance of any two attribute combinations of the original version of baselines and baselines trained with Meta-MCTG. An example result for CTRL is shown in Figure~\ref{fig:analysis-meta}. 
We observe that nearly all of the distances between $\mathcal{C}_{i.d.}$ and $\mathcal{C}_{comp}$ increase with Meta-MCTG and are notably larger than the distances within $\mathcal{C}_{i.d.}$. The results demonstrate that Meta-MCTG can distribute the hidden representations of attribute combinations more sparsely and thus possibly make them more distinguishable. Calculation details and more relevant results are available in Appendix~\ref{appendix:analysis exps}. Besides, we also present \textbf{case study} to compare the generation results of the original version of baselines and baselines trained with Meta-MCTG in Appendix~\ref{appendix: case study}, highlighting the better controllability of the latter ones.

\section{Conclusion}
We propose CompMCTG, the first holistic benchmark targeting compositional generalization for Multi-Aspect Controllable Text Generation (MCTG), and conduct extensive experiments on CompMCTG with eight representative MCTG baselines and two LLM baselines, unveiling noticeable compositional generalization risk in them and demonstrating the effectiveness of CompMCTG. In addition, we propose Meta-MCTG, a framework incorporating meta-learning into the MCTG training process to improve its compositional generalization ability, which can be combined with any joint-training-based MCTG methods.
\section*{Limitations}
Our proposed Meta-MCTG framework improves the compositional generalization performance of MCTG methods in most scenarios. However, when attribute combinations of data in the training set are extremely scarce (e.g., the \textit{Few-Shot} protocol in CompMCTG), we cannot build the pseudo-comp batch to utilize the Meta-MCTG framework. Besides, though Meta-MCTG is generally effective, current MCTG methods still have considerable room for improvement in compositional generalization. Both of these limitations will be areas for our future research.

Frankly speaking, the experimental workload of \textit{Hold-Out} protocol in the CompMCTG benchmark is overly cumbersome, and the average results in our main table do not include \textit{Few-Shot}, which we believe are discrepancies. For researchers with limited resources who want to follow our work, we recommend focusing on the performance of models under the \textit{ACD} and \textit{Few-Shot} protocols. These protocols are relatively more challenging and facilitate distinguishing models based on different capabilities.

\section*{Ethics Statement}
Multi-aspect controllable text generation is widely used in social media. However, improper use can cause serious negative effects, such as using this technology to spread inappropriate remarks (political attributes) or create rumors. Therefore this kind of technology should be subject to certain regulations. 
% In human evaluation, we pay annotators \$0.1 for each piece of data.

\section*{Acknowledgements}
We sincerely thank all the anonymous reviewers for their constructive comments.
% Tianqi Fundings
% Zhaoyi Fundings
This work is supported by the National Science Fund for Excellent Young Scholars under Grant 62222212 and the National Key R\&D Program of China (No.2021ZD0111801).
% Zhaoyi Li and Defu Lian was supported by grants from the National Key R\&D Program of China (No.2021ZD0111801).

\bibliography{anthology,custom}
\bibliographystyle{acl_natbib}

\clearpage
\appendix
\section*{Appendix}
\addcontentsline{toc}{section}{Appendix}
\section{Datasets}
\label{appendix-datasets}
% \begin{table}[t]
% \centering
% \resizebox{0.5\textwidth}{!}{
% \begin{tabular}{|lcccc|}
% \hline
% \textbf{Dataset} & \textit{Mixture}  & \textit{Amazon Review} & \textit{YELP} & \textit{Fyelp} \\
% \hline
% $m$ & $2$ & $2$ & $3$  & $4$ \\
% $|\mathcal{C}|$ & $8$ & $12$ & $8$ & $40$ \\
% \hline 
% \end{tabular}
% }
% \caption{
% Component datasets of CompMCTG benchmark: $m$ is the number of aspects (e.g., sentiment, topic, tense, and stuff); $|\mathcal{C}|$ is the number of attribute combinations.
% }
% \label{tab:dataset}
% \end{table}

We select a shopping review dataset: \textit{Amazon Review}~\cite{amazon_review}, a mixture of movie(IMDB~\cite{imdb}), tablet, automobile(Sentube~\cite{Sentube}) and hotel(OpenNER~\cite{OpenNER}) review dataset: \textit{Mixture}~\cite{Mixture}, and two restaurant review datasets: \textit{YELP}~\cite{YELP0,YELP} and \textit{FYelp}~\cite{Fyelp}. In this section, we mainly introduce the four datasets that make up our benchmark as mentioned above. 

\paragraph{Fyelp}
Following previous work~\cite{yang-etal-2023-tailor, huang-etal-2023-extensible, Fyelp}, we adopt the widely used \textit{Fyelp} dataset, which contains restaurant reviews with the sentiment (positive and negative), the cuisine (American, Mexican, Asian, Bar, and dessert), and the gender (Male and Female). To evaluate the extensibility of methods, we add one additional aspect of constraints: the tense (Past and Present)~\cite{ficler-goldberg-2017-controlling}, where its label is automatically extracted from the reviews. Thus far, the \textit{Fyelp} dataset is the one with the largest scale of attribute combinations in our benchmark. In total, there are $2\times 2\times 5\times 2=40$ possible attribute combinations.
\paragraph{Amazon Review}
Amazon Review~\cite{amazon_review} is a dataset containing reviews for Amazon products, which is widely used in previous academic works around text rewriting, controllable text generation, and stuff~\cite{li-tuzhilin-2019-towards, Fyelp, pmlr-v202-zhou23g}. Following ~\cite{Fyelp}, we process the dataset and label the data with two aspects: the sentiment (positive and negative) and the topic (Books, Clothing, Music, Electronics, Movies and Sports) with the meta-data in the original Amazon Review\footnote{\url{https://jmcauley.ucsd.edu/data/amazon/}} dataset. Hence there are $2 \times 6 = 12$ different attribute combinations.
\paragraph{YELP}
YELP business reviews dataset~\cite{YELP} contains the three aspects of attributes: the tense (Past and Present), the sentiment (positive and negative), and the person (singular and plural). We process the dataset in alignment with ~\cite{john-etal-2019-disentangled} and ~\cite{control_emnlp_2020} and randomly re-split the whole dataset for our usage. There are $2\times 2\times 2=8$ different attribute combinations in this dataset.
\paragraph{Mixture}
Mixture is the combination of three individual datasets: IMDb~\cite{imdb} (movie reviews) OpenNER~\cite{OpenNER} (hotel reviews) and SenTube~\cite{Sentube} (tablet and automobile reviews), constructed by ~\cite{Mixture}. Hence each datum in Mixture has two aspects of attributes: sentiment (positive and negative) and topic (movie, hotel, tablet, and automobile) and there are in total $2\times4=8$ possible attribute combinations.

We summarize all details and statistics of these datasets in Table~\ref{tab:datasets-details}. 
\begin{table*}[t]
\setlength\tabcolsep{15pt}
\centering
\begin{tabular}{l|ccccc}
\hline
\multirow{2}{*}{\textbf{Dataset}} & \multirow{2}{*}{$m$} & \multirow{2}{*}{$|\mathcal C|$}  & \multicolumn{2}{c}{\textbf{Classifier}} & \textbf{Generator}\\
 & & & Train & Development & Train\\
 \hline
 \textit{Fyelp} & 4 & 40 & 34000 & 6000 & 70000\\
 \hline
 \textit{Amazon} & 2 & 12 & 153000 & 27000 & 120000\\
 \hline
 \textit{Yelp} & 3 & 8 & 20400 & 3600 & 24000\\
 \hline
 \textit{Mixture} & 2 & 8 & 3624 & 640 & 4800\\
\hline
\end{tabular}
\caption{
Information of the datasets in our CompMCTG Benchmark. $m$ is the number of aspects (e.g., sentiment, topic, tense, and stuff); $|\mathcal{C}|$ is the number of attribute combinations. "Classifier" refers to the size of the data used for training the classifier. We split the data into training and development sets at a ratio of 8.5:1.5 based on this. "Generator" refers to the size of the data used for training the generative model. The data for each attribute combination is uniformly distributed across all sub-datasets (i.e., Train and Development of "Classifier" and Train of "Generator").
}
\label{tab:datasets-details}
\end{table*}

\section{Details of CompMCTG Benchmark}
\label{appendix:benchmark}
\subsection{Two Types of Testing}
Our CompMCTG benchmark contains four datasets: \textit{Fyelp}, \textit{Amazon}, \textit{Yelp}, and \textit{Mixture}. For each dataset, we divide it into two disjoint subsets: in-distribution set and the compositional set. The in-distribution set contains the data that is visible during training, while the compositional set contains the data that is not visible during training. The sets of attribute combinations contained in the in-distribution set and compositional set are defined as $\mathcal C_{i.d.}$ and $\mathcal C_{comp}$, respectively. We first train the model on the in-distribution set and then there are two types of testing. The first type that involves generating text attribute combinations from $\mathcal C_{i.d.}$ is referred to as \textbf{\textit{in-distribution testing}}, which tests the model's learning ability within the distribution of the training data. The second type that involves generating text with attribute combinations from $\mathcal C_{comp}$ is referred to as \textit{\textbf{compositional testing}}, which tests the model's compositional generalization ability beyond the distribution of the training data.
\subsection{Datasets Details}
For each dataset, the total number of data points $N$, the number of attribute combinations $|\mathcal C|$, and the number of data points per attribute combinations $N_i$ are related as $N=|\mathcal C|\times N_i$ (the data points per attribute combination are equal for all datasets). 

For the \textit{Hold-Out} protocol, we define the in-distribution set as the subset obtained by removing one attribute combination from the total dataset. Therefore, the size of the in-distribution set for the \textit{Hold-Out} protocol is given by $N\times (|\mathcal C|-1)/|\mathcal C|$.

For the $ACD$ protocol, we designed it such that the ratio between the in-distribution set and the compositional set is 1:1. Therefore, the size of the in-distribution set for the \textit{ACD} protocol is $N/2$.

For the \textit{Few-Shot} protocol, our requirement for the in-distribution set is: 1) Each individual attribute must appear at least once, and 2) The total number of attribute combinations should be minimal. Therefore, for the \textit{Few-Shot} protocol, the number attribute combinations in the in-distribution set is equal to the number of attributes in the aspect with the most attributes. Let's assume that the aspect with the most attributes in the dataset contains $M$ attributes. In this case, the size of the in-distribution set for the \textit{Few-Shot} protocol is $N\times M/|\mathcal C|$.

Take the \textit{Fyelp} dataset as an example. The total number of data points for training generator is $70000$, and the number of attribute combinations is $2\times2\times2\times5=40$. Therefore, $N=70000$, $|\mathcal C|=40$, $M=5$. Hence, the size of the in-distribution set for the \textit{Hold-Out} protocol is $70000\times(40-1)/40=68250$, for the \textit{ACD} protocol is $70000/2=35000$, and for the \textit{Few-Shot} protocol is $70000\times5/40=8750$. Similarly, we can calculate the corresponding sizes of the in-distribution sets for the other three datasets.

\subsection{Why Few-Shot not in Average Results?}
In Table~\ref{tab:compmctg}, the calculation of the Average does not include results from the \textit{Few-Shot} protocol. There are two reasons for this approach: 1) According to the design principles of our \textit{ACD} and \textit{Few-Shot} protocols, the partitioning results for the datasets \textit{Mixture} and \textit{Amazon} are consistent between \textit{ACD} and \textit{Few-Shot}; 2) The difficult level of the \textit{Few-Shot} protocol is relatively high for current models, and we aim to present this category as a direction for future research within the community.

\subsection{Results in CompMCTG}
As previously mentioned, the results in Table~\ref{tab:compmctg} and Table~\ref{tab:few-shot} represent the average outcomes across four datasets. In fact, for each dataset, the results for each protocol are derived from the average of multiple experiments.

For the \textit{Hold-Out} protocol, we define it as randomly selecting one attribute combination from the complete dataset. In order to eliminate bias during the experiments, we iterate over all attribute combinations, and the final result for each dataset in the \textit{Hold-Out} protocol is the average of all these results.

For the \textit{ACD} protocol, we maximize attribute divergence to partition the datasets. In our experiments, there are usually multiple optimal partitioning methods, hence we also average over all cases for the final results.

Similarly, for the \textit{Few-Shot} protocol, we partition the datasets by maximizing attribute divergence and take the average of all optimal partitioning results.

We present the number of partitioning methods included in different protocols across four datasets in Table~\ref{tab:split_num}.

\begin{table}[t]
\centering
\resizebox{0.5\textwidth}{!}{
\begin{tabular}{|lcccc|}
\hline
\textbf{Dataset} & \textit{Original} & \textit{Hold-Out} & \textit{ACD} & \textit{Few-Shot} \\
\hline
\textit{Fyelp}  & $1$ & $40$ & $10$ & $2$ \\
\textit{Amazon} & $1$ & $12$ & $-$ & $10$  \\
\textit{YELP}  & $1$ & $8$ & $10$ & $8$  \\
\textit{Mixture} & $1$ & $8$ & $-$ & $8$  \\
\hline 
\end{tabular}
}
\caption{
The number of partitioning methods included in different protocols across four datasets in CompMCTG Benchmark.
}
\label{tab:split_num}
\end{table}

\section{Complexity discussion}
\label{appendix:complexity}
In this section, we discuss the complexity of sweeping over all possibilities for ``Half\&Half” splitting (i.e.,$|\mathcal C_{i.d.}|=|\mathcal C_{comp}|$) in Section~\ref{sec3:construction}.
Following the denotations in Section~\ref{sec3:construction}: $m$ refers to the number of different aspects; $\mathcal{A}_i,(1\leq i\leq m)$ is the set of attribute values for the $i$-th aspect;  $\min_{1\leq i \leq m}|\mathcal{A}_i|=a$; the total number of attribute combinations is $\mathcal{O}(a^m)$.

Sweeping over all possible ``Half\&Half” splitting methods requires $\mathcal{O}(\tbinom{a^m}{a^m/2})$ kinds of situations, which can be derived as follows (using Stirling's formula~\cite{Robbins1955ARO}):
\begin{equation*}
\begin{split}
    \tbinom{a^m}{a^m/2} &=  \frac{(a^m)!}{(\frac{a^m}{2})!\cdot(\frac{a^m}{2})!} \approx \frac{ \sqrt{2\pi a^m}\cdot (\frac{a^m}{e})^{a^m}} {\pi a^m\cdot (\frac{a^m}{2e})^{a^m}} \\
    & =  \frac{ \sqrt{2\pi a^m}\cdot 2^{a^m}} {\pi a^m}
\end{split}
\end{equation*}
Hence $\mathcal{O}(\tbinom{a^m}{a^m/2}) \approx \mathcal{O}(\frac{ \sqrt{2\pi a^m}\cdot 2^{a^m}} {\pi a^m}) = \mathcal{O}((2-\eta)^{a^m})$ where $\eta \rightarrow 0$. This complexity is exponential to $a^m$ and thus unacceptable, which highly calls for an effective sampling strategy (i.e., \textit{ACD} in Section~\ref{sec3:construction}).
\section{Implementation Details}
\label{sec:Implementation details}
Our implementation is based on Hugging Face Transformer models\footnote{\url{https://github.com/huggingface/transformers}} and we use GPT-2 Medium as our backbone for all baselines (except two LLM baselines). In this section, we provide all the hyperparameters for the baselines and our Meta-MCTG method, as well as the training hyperparameters for the classifiers used for evaluation.

First of all, we unify the settings for all experiments during the generation phase. Following previous work~\cite{gu-etal-2022-distributional, gu-etal-2023-controllable}, we use the 35 prompts from PPLM~\cite{dathathri2019plug} for testing. For all MCTG baselines, we generate 10 texts for each prompt and each attribute combination, each text with a length of 50, and we adopt topk=200, topp=1.0, and temperature=1.0. For two LLM baselines, due to time and financial costs, we generate only one text for each prompt and each attribute combination. All experiments are completed on an NVIDIA A100 (80G) GPU.
\subsection{MCTG Baselines}
\label{sec:hyperparameters-generator}

\paragraph{Fudge}
Fudge~\cite{yang-klein-2021-fudge} uses a future discriminator to guide the GPT-2 for the generation. Following previous work~\cite{seen-to-unseen-acl2023}, for each dataset, we train a  Multilayer Perceptron (MLP) of dimension $d_{embd}\times m$ as the future discriminator, where $d_{embd}$ is the embedding dimension of GPT-2 Medium, and m is the number of all attribute combinations in the dataset. We set batch size to 8, epoch to 5, and learning rate to 3e-5 in the training phase for all datasets and all protocols. As for the generation, we set control strength $\alpha$ to 20 for all datasets and all settings.
\paragraph{PPLM}
PPLM~\cite{dathathri2019plug} uses a discriminator to calculate gradient to update the states of a language model and guide the model to generate texts with a certain attribute. We train a Multilayer Perceptron of dimension $d_{embd}\times m$ as the discriminator-like fudge to guide the model. For each dataset and each protocol, we set the batch size to 8, epoch to 5, and learning rate to 3e-5 in the training phase. As for the generation, we followed the hyperparameters in \citet{dathathri2019plug}. We set $\gamma$ to 1.5, num-iterations to 3, num-samples to 10, stepsize to 0.03, window-length to 5, fusion-kl-scale to 0.01, and fusion-gm-scale to 0.99.

\paragraph{Distributional Lens}
During the training phase, we follow all the hyperparameters of the original work~\cite{gu-etal-2022-distributional}, with the only change made to the number of training steps. We sweep across training steps from \{2000,4000,6000, ...,30000\} and select the minimum number of steps for convergence as our experimental setup. We summarize it in the Table \ref{tab:lens-training-params}. In the generation phase, for simplicity and fairness, we set all aspect weights to 1, and all other settings are consistent with the original paper.
\begin{table}[t]
\centering
\resizebox{0.5\textwidth}{!}{
\begin{tabular}{|lcccc|}
\hline
\textbf{Dataset} & \textit{Original} & \textit{Hold-Out} & \textit{ACD} & \textit{Few-Shot} \\
\hline
\textit{Fyelp}  & $8000$ & $8000$ & $4000$ & $4000$ \\
\textit{Amazon} & $6000$ & $6000$ & $-$ & $4000$  \\
\textit{YELP}  & $4000$ & $4000$ & $6000$ & $8000$  \\
\textit{Mixture} & $10000$ & $10000$ & $-$ & $10000$  \\
\hline 
\end{tabular}
}
\caption{
Training steps of different datasets and different protocols in Distributional Lens~\cite{gu-etal-2022-distributional}.
}
\label{tab:lens-training-params}
\end{table}

\paragraph{Prior}
Proposed by~\cite{gu-etal-2023-controllable}, this method is based on the model trained in \citet{gu-etal-2022-distributional}, with the training loss of the Normalizing Flows added for further training. Therefore, during the training phase, we further train based on all models trained by method \citet{gu-etal-2022-distributional}, with the hyperparameters consistent with the original work and only a change made to the number of training steps. Like experiments in \citet{gu-etal-2022-distributional}, we sweep across training steps from \{5000, 10000, ..., 50000\} and select the minimum number of steps for convergence as our experimental setup. We summarize it in the Table \ref{tab:prior-training-params}. In the generation phase, we find that aspect weights setting to 1 for the \textit{Fyelp} dataset do not yield satisfactory results. Therefore, we attempt to adjust the aspect weights on this dataset and finally set weights to [12,4,24,12] corresponding to aspect ["sentiment", "gender", "cuisine", "tense"] and std to 0.1. For the other three datasets, we set weight to 1 for all aspects and set std to 1.
\begin{table}[t]
\centering
\resizebox{0.5\textwidth}{!}{
\begin{tabular}{|lcccc|}
\hline
\textbf{Dataset} & \textit{Original} & \textit{Hold-Out} & \textit{ACD} & \textit{Few-Shot} \\
\hline
\textit{Fyelp}  & $30000$ & $30000$ & $30000$ & $30000$ \\
\textit{Amazon} & $30000$ & $30000$ & $-$ & $30000$  \\
\textit{YELP}  & $5000$ & $5000$ & $5000$ & $5000$  \\
\textit{Mixture} & $30000$ & $30000$ & $-$ & $30000$  \\
\hline 
\end{tabular}
}
\caption{
Training steps of different datasets and different protocols in Prior Control~\cite{gu-etal-2023-controllable}.
}
\label{tab:prior-training-params}
\end{table}

\paragraph{Catprompt}
As this is a naive method derived from \citet{yang-etal-2023-tailor}, there is no clear experiment setup for reference. We sweep across prompt length from \{10,20,40,60,80,100,120\}, selecting the length with the best test results for each attribute as our experimental hyperparameters. The specific results are as follows.  For the \textit{Fyelp} dataset, in the non-FewShot protocols, we set prompt length to 120, batch size to 16, epochs to 20, and learning rate to 5e-5, and in the FewShot protocol, we set prompt length to 100, batch size to 16, epochs to 40, and learning rate to 5e-5. For the \textit{Amazon} dataset, we set prompt length to 10, batch size to 16, epochs to 5, and learning rate to 5e-5 for all settings. For the \textit{YELP} dataset, in the non-FewShot protocols, we set prompt length to 20, batch size to 16, epochs to 20, and learning rate to 5e-5, and in the FewShot protocol, we set prompt length to 20, batch size to 16, epochs to 40, and learning rate to 5e-5. For the \textit{Mixture} dataset, we set prompt length to 10, batch size to 16, epochs to 50, and learning rate to 5e-5 for all settings.
\paragraph{DCG}
Following previous work~\cite{seen-to-unseen-acl2023}, for all settings across all datasets, prompt length is set to 50 (where attribute prompt length is set to 6 and task prompt length is set to 44), the disentanglement loss weight is set to 0.1, the batch size is set to 8, and the number of Pseudo Combinations is set to 7. For the setting of epochs, we set epochs to 3 for dataset \textit{Fyelp} and \textit{Amazon}, epochs to 8 for dataset \textit{YELP}, and epochs to 7 for dataset \textit{Mixture}. And for all datasets and protocols, we set the learning rate to 7.5e-5.
\paragraph{CTRL}
Following previous work~\cite{seen-to-unseen-acl2023}, we concatenate multi-attribute control codes with training datasets to fine-tune the GPT-2. Since we find that \textit{CTRL} is not sensitive to hyperparameters, we set the batch size to 8, epochs to 5, and learning rate to 3e-5, which converges well for all datasets and protocols.
\paragraph{Contrastive Prefix-Tuning}
Following previous work~\cite{contrastive-prefix}, we set each attribute's prefix length to 10. For the dataset \textit{Fyelp} and \textit{Amazon}, we set the batch size to 8 and epochs to 2 for all protocols. For the dataset \textit{YELP}, we set the batch size to 8 and epochs to 5 for all protocols. For the dataset \textit{Mixture}, we set the batch size to 8 and epochs to 5 for non-FewShot protocols. For the FewShot protocol of the dataset \textit{Mixture}, we set the batch size to 8 and the epoch to 10. And for all datasets and protocols, we set the learning rate to 3e-5.
\subsection{LLM Baselines and Prompts}
In this section, we introduce the LLMs we use in Section~\ref{sec3:evaluation} and the prompt template we used for In-Context Learning.
\paragraph{Prompt} Following~\cite{evaluating_llms}, we use \textit{5-shot} in context learning prompt template to evaluate the compositional generalization capacity of LLMs regarding ICL. Namely, we insert five demonstrations (Input, Output) for each time of controllable generation. Here is our prompt template:
\begin{lstlisting}
\\5-shot in-context-learning 
\\prompt template
"Task: write a sentence that meets the requirement of input control conditions.
Below are some examples (Input, Output) for the task:
Input: <attribute combination 1>. 
Output: <text 1>   # demonstration_1
Input: <attribute combination 2>. 
Output: <text 2>   # demonstration_2
Input: <attribute combination 3>. 
Output: <text 3>   # demonstration_3
Input: <attribute combination 4>. 
Output: <text 4>   # demonstration_4
Input: <attribute combination 5>. 
Output: <text 5>   # demonstration_5
Input: <testing attribute combination>.
Output: <a head of text>"  \\ generation
\end{lstlisting}
For in-distribution testing, we insert five demonstrations that share the control conditions (in the attribute combinations) with the testing one.
For compositional testing, we uniformly sample five demonstrations (of different attribute combinations) from the whole training set.

Another point that is worth noting is that we encode the control conditions in a standard format (e.g., in Yelp we use “cuisine-0” to represent Asian cuisine, “cuisine-1” to represent Mexican cuisine, “gender-0” to represent gender Male, “gender-1” to represent gender Female and so on). The underlying reason is that we aim to test the LLM's ability to understand the relationship between control attributes and target text content, as well as their capacity to generalize to new combinations of previously seen control attributes.
\paragraph{LLM} For LLaMA-2~\cite{llama2}, we adopt the version of “LLaMA-2-7B-hf”\footnote{\url{https://huggingface.co/meta-llama/Llama-2-7b-hf}}. Our generation configuration is following the default configuration provided by Meta:
\begin{lstlisting}
\\LLaMA-2-7B generation configuration
GEN_CONFIGS["llama2-7b"]={
  "bos_token_id": 1,
  "do_sample": True,
  "eos_token_id": 2,
  "pad_token_id": 0,
  "temperature": 0.6,
  "max_length": 50,
  "top_p": 0.9,
  "transformers_version": "4.31.0.dev0"
}
\end{lstlisting}
For ChatGPT~\cite{openaiChatGPT}, we use the OpenAI-api\footnote{\url{https://openai.com/blog/openai-api}} and adpot the version of “gpt-3.5-turbo-0613”. The default generation configuration is as follows:
\begin{lstlisting}
\\gpt-3.5 generation configuration
GEN_CONFIGS["gpt-3.5-turbo-0613"]={
  "temperature": 1.0,
  "max_length": 50,
  "top_p": 0.9,
  "openai_version": "0.28.0"
}
\end{lstlisting}
\paragraph{Cost} For the evaluation of LLaMA-2-7B, we do experiments on a NVIDIA A100 GPU for around 60 hours. For the evaluation of ChatGPT, we spend around 3.5e7 tokens in total, costing 70 dollars.
\label{llms and prompts}
\subsection{Classifiers}
\label{sec:hyperparameters-classifier}
To avoid the impact of domain differences among different datasets on the accuracy of the classifier, we train a classifier using Roberta-Large~\cite{liu2019roberta} for each aspect of each dataset. We sweep over batch sizes from \{4,8,16,32,64,128,256,512,1024\} and epochs from \{1,2,3,4,5,6,7,8,9,10\}, choosing the settings that yield the highest accuracy on the test set as our experimental configuration. The specific configuration results and the performance of the classifiers on the test set for all datasets and all attribute aspects are shown in Table \ref{tab:classifiers-param}.
\begin{table}[t]
\centering
\resizebox{0.5\textwidth}{!}{
\begin{tabular}{|lcccc|}
\hline
\textbf{Dataset} & \textbf{Aspect} & \textbf{Batch} & \textbf{Epochs} & \textbf{Accuracy}\\
\hline
\multirow{4}{*}{\textit{Fyelp}} & \textit{Sentiment} & $512$ & $5$ & $98.68\%$ \\
 & \textit{Gender} & $512$ & $3$ & $70.68\%$ \\
 & \textit{Cuisine} & $64$ & $4$ & $77.97\%$ \\
 & \textit{Tense} & $32$ & $4$ & $88.57\%$ \\
\hline
\multirow{2}{*}{\textit{Amazon}} & \textit{Sentiment} & $128$ & $5$ & $98.41\%$  \\
 & \textit{Topic} & $64$ & $5$ & $92.84\%$\\
\hline
\multirow{3}{*}{\textit{YELP}}  & \textit{Sentiment} & $1024$ & $5$ & $97.11\%$  \\
 & \textit{Person} & $32$ & $8$ & $99.42\%$ \\
 & \textit{Tense} & $256$ & $3$ & $99.78\%$ \\
\hline
\multirow{2}{*}{\textit{Mixture}} & \textit{Sentiment} & $128$ & $4$ & $84.37\%$  \\
 & \textit{Topic} & $512$ & 8 & $98.59\%$ \\
\hline 
\end{tabular}
}
\caption{
The specific configuration and the performance of the classifiers used in our benchmark.
}
\label{tab:classifiers-param}
\end{table}

\subsection{Meta-MCTG}
\label{sec:hyperparameters-meta-MCTG}
In the experiments of Meta-MCTG, we select the three best-performing joint-training-based methods from the baselines, namely \textit{CTRL}~\cite{keskar2019ctrl}
, \textit{DCG}~\cite{seen-to-unseen-acl2023}, and \textit{Contrastive Prefix}~\cite{qian-etal-2022-controllable}. For different datasets and protocols in our benchmark, we search $\lambda$ from \{0.01,0.05,0.1,0.2\} based on the original experimental hyperparameters, and further refine the value of $\lambda$ based on the results. For the majority of cases, we opt for $\lambda$ to be 0.01. For the learning rate $\beta$ in all MCTG experiments, we set $\beta$ to be the same as the learning rate $\alpha$ of each baseline.

% \begin{table}[t]
% \centering
% \resizebox{0.5\textwidth}{!}{
% \begin{tabular}{|l|cc|c|cc|c|}
% \hline
% \multirow{2}{*}{\textbf{Method}} & \multicolumn{2}{c|}{\textit{Fyelp}} & \textit{Amazon} & \multicolumn{2}{c|}{\textit{YELP}} & \textit{Mixture} \\
%  & \textit{Hold.} & \textit{ACD} & \textit{Hold.} & \textit{Hold.} & \textit{ACD} & \textit{Hold.}\\
% \hline
% \textit{CTRL} & 0.01 & 0.01 & 0.005 & 0.01 & 0.01 & 0.01 \\
% \hline
% \textit{Con.P.} & 0.01 & 0.1 & 0.01 & 0.04 & 0.2 & 0.05 \\
% \hline
% \textit{DCG} & 0.01 & 0.2 & 0.01 & 0.007 & 0.1 & 0.006 \\
% \hline 
% \end{tabular}
% }
% \caption{
% The selection of $\lambda$ in different datasets and settings in the Meta-MCTG experiments. "Hold." represents the Hold-Out protocol and "\textit{Con.P.}" represents the baseline \textit{Contrastive Prefix}.
% }
% \label{tab:hyperparameters-meta-MCTG}
% \end{table}

\section{Evaluation on diversity}
\label{sec:Experiments of diversity}
Following previous work~\cite{li-etal-2016-diversity}, we use distinctness to measure the generated text's diversity. For each text, we calculate 3-grams named Dist-3 to evaluate distinctness. We choose to conduct diversity evaluation on the data under the three protocols of \textit{Original}, \textit{Hold-Out}, and \textit{ACD}. The whole results are shown in Table \ref{tab:compmctg-diversity}.

\begin{table*}[t]
\centering
\resizebox{\textwidth}{!}{
\begin{tabular}{lc|cc|cc|c}
\hline
\multirow{2}{*}{\textbf{Method}} & \textit{Original} & \multicolumn{2}{c|}{\textit{Hold-Out}} & \multicolumn{2}{c|}{\textit{ACD}} & \multicolumn{1}{c}{\textit{Average}}\\
 & \textit{Dist-3}\textsubscript{\textit{i.d.}}$(\uparrow)$ & \textit{Dist-3}\textsubscript{\textit{i.d.}} & \textit{Dist-3}\textsubscript{\textit{comp}} & \textit{Dist-3}\textsubscript{\textit{i.d.}} & \textit{Dist-3}\textsubscript{\textit{comp}} & \textit{Dist-3}\textsubscript{\textit{avg}}\\
\hline
\hline
\textbf{LLM+In-context Learning}\\
\hline
\textit{LLaMA-2}~\cite{llama2} & 0.587 & 0.430 & 0.577 & 0.456 & 0.451 & 0.500 \\
\textit{ChatGPT}~\cite{openaiChatGPT} & 0.611 & 0.408 & 0.660 & 0.451 & 0.457 & 0.517\\
\hline\hline
\textbf{Decoding-Time based} \\
\hline
\textit{Fudge}~\cite{yang-klein-2021-fudge} & 0.656 & 0.652 & 0.621 & 0.625 & 0.587 & 0.628 \\
\textit{PPLM}~\cite{dathathri2019plug} & 0.697 & 0.622 & 0.694 & 0.621 & 0.617 & 0.650 \\
\hline
\hline
\textbf{Separate-Training based} \\
\hline
\textit{Dis-Lens}~\cite{gu-etal-2022-distributional} & 0.473 & 0.466 & 0.462 & 0.454 & 0.427 & 0.456  \\
\textit{Prior}~\cite{gu-etal-2023-controllable} & 0.573 & 0.547 & 0.548 & 0.539 & 0.540 & 0.549  \\
\hline
\hline
\textbf{Joint-Training based} \\
\hline
\textit{CTRL}~\cite{keskar2019ctrl} & 0.625 & 0.623 & 0.634 & 0.616 & 0.622 & 0.624\\
\textit{CatPrompt}~\cite{yang-etal-2023-tailor} & 0.642 & 0.636 & 0.656 & 0.677 & 0.688 & 0.660  \\
\textit{Con.Prefix}~\cite{qian-etal-2022-controllable} & 0.701 & 0.696 & 0.727 & 0.682 & 0.717 & 0.705 \\
\textit{DCG}~\cite{seen-to-unseen-acl2023} & 0.677 & 0.694 & 0.716 & 0.675 & 0.695 & 0.691 \\
\hline
\hline 
\end{tabular}
}
\caption{
Averaged overall evaluation results of \textbf{diversity} for state-of-the-art baseline approaches on our CompMCTG benchmark (\textit{Hold-Out} testing and \textit{ACD testing}). Subscript \textit{i.d.} and \textit{comp} refer to in-distribution and compositional generalization performance.
}
\label{tab:compmctg-diversity}
\end{table*}

\section{Human Evaluation}
\label{human evaluation}
Following previous work~\cite{zhang-song-2022-discup, zhong2023airdecoding}, we evaluate generated texts from two aspects: \textbf{Relevance} (\textbf{R}) which reflects the degree of achievement for the desired control attribute combination and \textbf{Fluency} (\textbf{F}) which evaluates the text's fluency. Unlike automated evaluation, where the accuracy of individual attributes is measured and averaged, human evaluation directly scores the satisfaction of the given control condition (attribute combination). For each dataset and baseline in each protocol (\textit{Original, HoldOut, and ACD}), we randomly sample 10 texts (for \textit{HoldOut} and \textit{ACD}, we sample 10 texts from in-distribution result and 10 texts from compositional result) and employ three annotators to score them on the two metrics on a scale from 1 (very bad) to 5 (very good). Finally, we calculate the average of these scores and get the final result shown in Table \ref{tab:human-evaluation}. We can find that the results of human evaluation are consistent with the results of automated evaluation.

\subsection{Specific Scoring Guidelines}
In this subsection, we provide specific scoring guidelines for each human evaluation metric.
\paragraph{Relevance}
\begin{itemize}[leftmargin=*]
    \item $5$: The generated texts are perfectly aligned with the desired attribute combination.
    \item $4$: The generated texts are very related to the desired attribute combination.
    \item $3$: The generated texts are related to the desired attribute combination. At most one attribute does not match.
    \item $2$: The generated texts are less related to the desired attribute combination. At most two attributes do not match.
    \item $1$: The generated texts are not aligned with the desired attribute combination. None of the attributes meet the requirements.
\end{itemize}

\paragraph{Fluency}
\begin{itemize}[leftmargin=*]
    \item $5$: The generated texts are grammatically correct, fluent, and easy to understand.
    \item $4$: The generated texts are grammatically correct, but slightly less smooth, yet still easily understandable.
    \item $3$: The generated texts have a few grammar errors, but do not hinder understanding.
    \item $2$: The generated texts have a few grammar errors and are not very easy to understand.
    \item $1$: The generated texts have many grammar errors, lack coherence, and are difficult to understand.
\end{itemize}
\subsection{Inter-Annotator Agreement Score}
We also use \textbf{Fleiss'Kappa coefficient}~\cite{fleiss1971measuring} to measure the inter-annotator agreement score for each human evaluation metric. The result is shown in Table \ref{tab:human-evaluation-kappa}.

\section{Case Study}
\label{appendix: case study}
In this section, we show some specific generation examples, primarily to compare the difference in generation results before and after using the Meta-MCTG framework. Cases in this section are from the compositional result of \textit{ACD} protocol of dataset \textit{Fyelp}. The specific results are shown in Table \ref{tab:case-study}.

\begin{table*}[t]
\centering
\resizebox{\textwidth}{!}{
\begin{tabular}{lcc|cccc|cccc|cc}
\hline
\multirow{2}{*}{\textbf{Method}} &  \multicolumn{2}{c}{\textit{Original}} & \multicolumn{4}{c}{\textit{Hold-Out}} & \multicolumn{4}{c}{\textit{ACD}} & \multicolumn{2}{c}{\textit{Average}}  \\

  & \textit{R}\textsubscript{\textit{i.d.}}$(\uparrow)$& \textit{F}\textsubscript{\textit{i.d.}} $(\uparrow)$
 & \textit{R}\textsubscript{\textit{i.d.}} $(\uparrow)$ & \textit{F}\textsubscript{\textit{i.d.}}$(\uparrow)$ & \textit{R}\textsubscript{\textit{comp}}$(\uparrow)$ &  \textit{F}\textsubscript{\textit{comp}}$(\uparrow)$
 & \textit{R}\textsubscript{\textit{i.d.}} $(\uparrow)$ & \textit{F}\textsubscript{\textit{i.d.}} $(\uparrow)$ & \textit{R}\textsubscript{\textit{comp}} $(\uparrow)$&  \textit{F}\textsubscript{\textit{comp}}$(\uparrow)$
  & \textit{R}\textsubscript{\textit{avg}} $(\uparrow)$& \textit{F}\textsubscript{\textit{avg}} $(\uparrow)$ \\
 \hline
 \hline
\textbf{LLM+In-Context Learning}\\
\hline
 \textit{LLaMA-2}\footnotesize{~\cite{llama2}}    &3.12 & 4.56  &3.23 & 4.48 &2.37 & 4.43 & 3.31& 4.60 & 2.22 &4.59  &2.85 &4.53  \\
  \textit{ChatGPT}\footnotesize{~\cite{openaiChatGPT}} & 2.89 & 4.78 & 2.86 & 4.75 & 2.47 & 4.81 & 2.75 & 4.88 & 2.57 & 4.74 & 2.71 & \textbf{4.79}  \\
\hline
\textbf{Decoding-Time based} \\
 \hline
 \textit{PPLM}\footnotesize{~\cite{dathathri2019plug}}    & 2.07 & 1.12 & 2.22 & 1.07 & 2.01 & 1.09 & 2.16 & 1.14 & 1.82 & 1.03 &2.06 & 1.09 \\
\textit{Fudge}\footnotesize{~\cite{yang-klein-2021-fudge}}  & 2.88 & 2.35 & 2.68 & 2.13 &  2.07 & 1.87 & 2.59 & 1.90 & 1.97 & 2.24 & 2.44 & 2.10\\

\hline
\textbf{Separate-Training based} \\
\hline
\textit{Dis-Lens}\footnotesize{~\cite{gu-etal-2022-distributional}}  & 4.24 & 2.86 & 4.10 & 3.12 & 2.55 & 3.01 & 4.44 & 3.21 & 2.42 & 2.91 &3.55 & 3.02 \\
\textit{Prior}\footnotesize{~\cite{gu-etal-2023-controllable}} & 3.67 & 2.96 & 3.53 & 3.04 & 2.42 & 3.20 & 3.78 & 3.03 & 2.39 & 3.24 & 3.16 & 3.09  \\

\hline
\textbf{Joint-Training based} \\
\hline
\textit{CTRL}\footnotesize{~\cite{keskar2019ctrl}}  & 3.98 & 3.87 & 3.78 & 3.92 & 3.75 & 3.94 & 3.80 & 3.81 & 3.55 & 3.84 & 3.77 & 3.88\\
\textit{CatPrompt}\footnotesize{~\cite{yang-etal-2023-tailor}}  & 3.23 & 3.52 & 3.27 & 3.49 & 3.04 & 3.58 & 3.01 & 3.07 & 2.45 & 3.61 & 3.00 & 3.45   \\
\textit{Con.Prefix}\footnotesize{~\cite{contrastive-prefix}}  & 4.22 & 3.44 & 4.19 & 3.40 & 4.01 & 3.13 & 4.15 & 3.23 & 3.52 & 3.12 & \textbf{4.02} & 3.26  \\
\textit{DCG}\footnotesize{~\cite{seen-to-unseen-acl2023}} & 3.92 & 3.80 & 3.90 & 3.68 & 3.84 & 3.64 & 3.88 & 3.83 & 3.39 & 3.73 & 3.79 & 3.74  \\
\hline 
\end{tabular}
}
\caption{
Averaged overall \textbf{human evaluation} results for state-of-the-art baseline approaches on our CompMCTG benchmark (\textit{Hold-Out} testing and \textit{ACD} testing). "R" refers to metric "Relevance" and "F" refers to metric "Fluency". Subscript \textit{i.d.} and \textit{comp} refer to in-distribution and compositional generalization performance.
}
\label{tab:human-evaluation}
\end{table*}

\begin{table*}[t]
\centering
\resizebox{\textwidth}{!}{
\begin{tabular}{lcc|cccc|cccc}
\hline
\multirow{2}{*}{\textbf{Method}} &  \multicolumn{2}{c}{\textit{Original}} & \multicolumn{4}{c}{\textit{Hold-Out}} & \multicolumn{4}{c}{\textit{ACD}}\\

  & \textit{R}\textsubscript{\textit{i.d.}}$(\uparrow)$& \textit{F}\textsubscript{\textit{i.d.}} $(\uparrow)$
 & \textit{R}\textsubscript{\textit{i.d.}} $(\uparrow)$ & \textit{F}\textsubscript{\textit{i.d.}}$(\uparrow)$ & \textit{R}\textsubscript{\textit{comp}}$(\uparrow)$ &  \textit{F}\textsubscript{\textit{comp}}$(\uparrow)$
 & \textit{R}\textsubscript{\textit{i.d.}} $(\uparrow)$ & \textit{F}\textsubscript{\textit{i.d.}} $(\uparrow)$ & \textit{R}\textsubscript{\textit{comp}} $(\uparrow)$&  \textit{F}\textsubscript{\textit{comp}}$(\uparrow)$ \\
 \hline
 \hline
\textbf{LLM+In-context Learning}\\
\hline
 \textit{LLaMA-2}\footnotesize{~\cite{llama2}}    &0.823 & 0.805  &0.834 & 0.816 &0.840 & 0.809 & 0.825& 0.833 & 0.836 &0.824  \\
  \textit{ChatGPT}\footnotesize{~\cite{openaiChatGPT}} & 0.811 & 0.814 & 0.805 & 0.843 & 0.827 & 0.840 & 0.829 & 0.860 & 0.851 & 0.837  \\
\hline
\textbf{Decoding-Time based} \\
 \hline
 \textit{PPLM}\footnotesize{~\cite{dathathri2019plug}}    & 0.910 & 0.908 & 0.887 & 0.893 & 0.828 & 0.839 & 0.834 & 0.890 & 0.887 & 0.836  \\
\textit{Fudge}\footnotesize{~\cite{yang-klein-2021-fudge}}  & 0.845 & 0.814 & 0.838 & 0.829 &  0.845 & 0.789 & 0.830 & 0.892 & 0.846 & 0.837 \\

\hline
\textbf{Separate-Training based} \\
\hline
\textit{Dis-Lens}\footnotesize{~\cite{gu-etal-2022-distributional}}  & 0.923 & 0.898 & 0.914 & 0.887 & 0.791 & 0.867 & 0.910 & 0.879 & 0.801 & 0.882\\
\textit{Prior}\footnotesize{~\cite{gu-etal-2023-controllable}} & 0.858 & 0.838 & 0.835& 0.846 & 0.837 & 0.821 & 0.845 & 0.883 & 0.826 & 0.818 \\

\hline
\textbf{Joint-Training based} \\
\hline
\textit{CTRL}\footnotesize{~\cite{keskar2019ctrl}}  & 0.830 & 0.808 & 0.845 & 0.794 & 0.815 & 0.829 & 0.810 & 0.822 & 0.816 & 0.815\\
\textit{CatPrompt}\footnotesize{~\cite{yang-etal-2023-tailor}}  & 0.782 & 0.804 & 0.793 & 0.811 & 0.824 & 0.815 & 0.806 & 0.785 & 0.823 & 0.836  \\
\textit{Con.Prefix}\footnotesize{~\cite{contrastive-prefix}}  & 0.898 & 0.843 & 0.904 & 0.826 & 0.876 & 0.837 & 0.879 & 0.841 & 0.844 & 0.820  \\
\textit{DCG}\footnotesize{~\cite{seen-to-unseen-acl2023}} & 0.857 & 0.886 & 0.854 & 0.874 & 0.818 & 0.825 & 0.857 & 0.867 &  0.834& 0.826  \\
\hline 
\end{tabular}
}
\caption{
Averaged overall \textbf{Fleiss'Kappa coefficient} of human evaluation results for state-of-the-art baseline approaches on our CompMCTG benchmark (\textit{Hold-Out} testing and \textit{ACD} testing). "R" refers to the Kappa coefficient of metric "Relevance" and "F" refers to the Kappa coefficient of metric "Fluency". Subscript \textit{i.d.} and \textit{comp} refer to in-distribution and compositional generalization performance.
}
\label{tab:human-evaluation-kappa}
\end{table*}

\begin{table*}[t]
\centering
% \setlength\tabcolsep{5.6pt}
% \resizebox{\textwidth}{!}{
\begin{tabular}{|m{2cm}|m{1.7cm}|m{10.8cm}|}
\hline
\textbf{Method} & \textbf{Comb.} & \textbf{Generated Texts}\\
\hline
\textit{CTRL} & \textcolor{red}{\textit{Negative}}$\quad$ \textcolor{yellow}{\textit{Female}}$\quad$ \textcolor{blue}{\textit{Bar}}$\quad\quad$ \textcolor{green}{\textit{Past}} & \textbf{Furthermore}, \textcolor{red}{where's the shrimp?} there are currently seven shrimp clusters. thank god there are still two clusters and that's where we ordered our food... at this location. despite this, we \textcolor{green}{waited} over 45 minutes. the server put me on \\
\hline
\textit{CTRL+Meta} & \textcolor{red}{\textit{Negative}}$\quad$ \textcolor{yellow}{\textit{Female}}$\quad$ \textcolor{blue}{\textit{Bar}}$\quad\quad$ \textcolor{green}{\textit{Past}} & \textbf{Furthermore}, this tiny place \textcolor{red}{fell} short of its potential. The food \textcolor{green}{was} enormous but \textcolor{red}{not up to par}. I \textcolor{green}{ordered} an \textcolor{blue}{oyster benedict} which \textcolor{green}{was} \textcolor{red}{disappointing}, and the \textcolor{blue}{wine} \textcolor{green}{was} \textcolor{red}{subpar}. I still find myself having \textcolor{red}{nightmares} about that dessert, which \textcolor{green}{was} made with \textcolor{red}{stale} \\
\hline

\textit{CTRL} & \textcolor{red}{\textit{Positive}}$\quad$ \textcolor{yellow}{\textit{Female}}$\quad$ \textcolor{blue}{\textit{Asian}}$\quad\quad$ \textcolor{green}{\textit{Past}} & \textbf{The movie} i \textcolor{green}{saw} called \" dark star \" and it really \textcolor{red}{impressed} me. the chef in the movie is \textcolor{red}{so nice} and did his chef impression. this restaurant \textcolor{green}{was} amazing to watch. we \textcolor{green}{had} the eggplant korma and beef teriyaki. \\
\hline
\textit{CTRL+Meta} & \textcolor{red}{\textit{Positive}}$\quad$ \textcolor{yellow}{\textit{Female}}$\quad$ \textcolor{blue}{\textit{Asian}}$\quad\quad$ \textcolor{green}{\textit{Past}} & \textbf{The movie} \textcolor{green}{started} and we \textcolor{green}{were} getting ready for bed but the \textcolor{blue}{korean} guys \textcolor{green}{got} us a lot of food. \textcolor{yellow}{my husband} and i both \textcolor{green}{got} the \textcolor{red}{delicious} beef \textcolor{blue}{kimchi rice} dish and both \textcolor{red}{really liked} it and \textcolor{green}{were} \textcolor{red}{very happy} with it. we \textcolor{green}{ended} \\
\hline

\textit{DCG} & \textcolor{red}{\textit{Positive}}$\quad$ \textcolor{yellow}{\textit{Female}}$\ \ $ \textcolor{blue}{\textit{dessert}}$\quad$ \textcolor{green}{\textit{Past}} & \textbf{In brief}, we are so \textcolor{red}{glad} we \textcolor{green}{went} here. it's always as cold as winter to get the fruit or veg for me ( some frozen squash, maybe a banana, too ). They have everything we go for here, and they\\
\hline
\textit{DCG+Meta} & \textcolor{red}{\textit{Positive}}$\quad$ \textcolor{yellow}{\textit{Female}}$\ \ $ \textcolor{blue}{\textit{dessert}}$\quad$ \textcolor{green}{\textit{Past}} & \textbf{In brief}, this \textcolor{green}{was} an \textcolor{red}{awesome} place. \textcolor{green}{Forget} the size of it, which i really \textcolor{green}{found} to be little too large, this \textcolor{green}{was} \textcolor{red}{SO GOOD}. We \textcolor{green}{stopped} in for breakfast and \textcolor{green}{decided} to try the \textcolor{blue}{sweet omelet pancakes}. \textcolor{yellow}{My husband} and\\
\hline

\textit{DCG} & \textcolor{red}{\textit{Negative}}$\quad$ \textcolor{yellow}{\textit{Male}}$\ \ $ \textcolor{blue}{\textit{Mexican}}$\quad$ \textcolor{green}{\textit{Present}} & \textbf{More importantly}, they have no toilet paper. would \textcolor{red}{NEVER EVER} order coffee or soda here.! they also \textcolor{green}{give} you a coupon for soup to go. \textcolor{red}{not the best}. everyone \textcolor{green}{is} \textcolor{red}{rude}. it \textcolor{green}{is} a \textcolor{red}{crowded} place. what gives there drive is that\\
\hline
\textit{DCG+Meta} & \textcolor{red}{\textit{Negative}}$\quad$ \textcolor{yellow}{\textit{Male}}$\ \ $ \textcolor{blue}{\textit{Mexican}}$\quad$ \textcolor{green}{\textit{Present}} & \textbf{More importantly}, the food \textcolor{red}{isn 't good enough} for me. \textcolor{yellow}{my girlfriend's} favorite \textcolor{blue}{taco} out of the bunch, \textcolor{blue}{Taco Linguini}, \textcolor{green}{is} supposed to be good but she \textcolor{red}{never} saw it ; dang there you guys. my salsa \textcolor{green}{is} really a \textcolor{red}{letdown}. \textcolor{green}{It's} too \textcolor{red}{bland} and \textcolor{red}{lacks} the right kick\\
\hline

\textit{Con.P.} & \textcolor{red}{\textit{Negative}}$\quad$ \textcolor{yellow}{\textit{Female}}$\ \ $ \textcolor{blue}{\textit{Asian}}$\quad$ \textcolor{green}{\textit{Past}} & \textbf{The last time} I \textcolor{green}{went} to a restaurant in town for sushi I \textcolor{green}{was} happy with the time but \textcolor{green}{was} \textcolor{red}{disappointed} the broth was chalky with \textcolor{blue}{soy sauce} and \textcolor{blue}{rice}. The temperature \textcolor{green}{was} \textcolor{red}{extreme} and the restaurant \textcolor{green}{had} no food prepared that \textcolor{green}{looked} appealing even when I \\
\hline
\textit{Con.P.+Meta} & \textcolor{red}{\textit{Negative}}$\quad$ \textcolor{yellow}{\textit{Female}}$\ \ $ \textcolor{blue}{\textit{Asian}}$\quad$ \textcolor{green}{\textit{Past}} & \textbf{The last time} I \textcolor{green}{was} to see the \textcolor{blue}{sushi place} here I \textcolor{green}{felt} \textcolor{red}{poor}. \textcolor{yellow}{My boyfriend} and I \textcolor{green}{felt} \textcolor{red}{uneasy} entering our table, so we \textcolor{green}{were} at all to begin with and he \textcolor{green}{waited} outside to eat lunch all the way until we \textcolor{green}{were} seated. The food \textcolor{green}{was} \textcolor{red}{bad}\\
\hline 

\textit{Con.P.} & \textcolor{red}{\textit{Positive}}$\quad$ \textcolor{yellow}{\textit{Male}}$\ \ $ \textcolor{blue}{\textit{American}} \textcolor{green}{\textit{Past}} & \textbf{The book} is \textcolor{red}{well} written and \textcolor{red}{well} planned with lots of really \textcolor{red}{delicious-to-and-simple} recipes and an in depth look at the last few years in the region with some \textcolor{red}{wonderful} photos and interesting twists on local food. Many thanks to my husband for\\
\hline
\textit{Con.P.+Meta} & \textcolor{red}{\textit{Positive}}$\quad$ \textcolor{yellow}{\textit{Male}}$\ \ $ \textcolor{blue}{\textit{American}} \textcolor{green}{\textit{Past}} & \textbf{The book} \textcolor{green}{commenced} with the account of a \textcolor{blue}{baseball-loving American} daycare worker in a center for immigrant families on Thanksgiving. "Every day, this \textcolor{yellow}{gentle man}, with his \textcolor{red}{warm smile}, \textcolor{green}{taught} the children that their most vital abilities resided within them\\
\hline 
\end{tabular}
% }
\caption{
A case study of the state-of-the-art baselines before and after incorporating the Meta-MCTG training framework. Different attribute words are marked with their corresponding colors. The text in bold represents the prompt. ``Comb.” means attribute combination and ``Con.P.” represents the baseline ContrastivePrefix.
}
\label{tab:case-study}
\end{table*}

\section{Algorithm Pseudo-Code}
\label{sec:appendix-pseduocode}
We conclude the pseudo-code of constructing ACD splits in Algorithm~\ref{alg:ACD} and the pseudo-code of Meta-MCTG training in Algorithm~\ref{alg:meta-learning}.

Following the denotations in Section~\ref{sec3:construction}: $m$ refers to the number of different aspects; $\mathcal{A}_i,(1\leq i\leq m)$ is the set of attribute values for the $i$-th aspect;  $\min_{1\leq i \leq m}|\mathcal{A}_i|=a$; the total number of attribute combinations is $\mathcal{O}(a^m)$. The time complexity of Algorithm~\ref{alg:ACD} (Greedily constructing ACD splits) is $\mathcal{O}(T_1\cdot T_2\cdot a^m)$ (linearly increasing with $a^m$) which is much better than $\mathcal{O}((2-\epsilon)^{a^m}), (\epsilon \leftarrow 0)$ (exponentially increasing with $a^m$) in Appendix~\ref{appendix:complexity}.
\begin{algorithm}
    \caption{Constructing ACD splits}
    \begin{algorithmic}[1]
        \REQUIRE Attribute combination set $\mathcal{C}$.
        \REQUIRE Divergence function $D(\cdot,\cdot)$.
        \REQUIRE Maximum step $T_1, T_2$, maximum divergence threshold $\eta\in(0,1)$.
        \STATE Initialization: current step $t_1 = 0$; maximum divergence $d_m = 0$.
        \STATE A set of ACD splits \textit{result} = $\emptyset$.
        \WHILE{$t_1<T_1$}
        \STATE $t_1 = t_1 + 1$
        \STATE Randomly split $\mathcal{C}$ into $\mathcal{C}_{i.d.}$ and $\mathcal{C}_{comp}$ where $|\mathcal{C}_{i.d.}|=|\mathcal{C}_{comp}|$.
        \STATE $t_2 = 0$
        \STATE Compute current divergence $d$:\\
        $d = D(\mathcal{C}_{i.d.}, \mathcal{C}_{comp})$.
        \STATE Update maximum divergence: $d_m = d$.
        \WHILE{$t_2 < T_2$}
        \STATE $t_2 = t_2+1$
        \STATE  $c_1 = None$.
        \FOR{$c \in \mathcal{C}_{i.d.}$}
        \IF{$d_m < D(\mathcal{C}_{i.d.} - \{c\}, \mathcal{C}_{comp} + \{c\})$}
        \STATE $c_1 = c$.
        \STATE $d_m = D(\mathcal{C}_{i.d.} - \{c\}, \mathcal{C}_{comp} + \{c\})$.
        \STATE \textbf{break}
        \ENDIF
        \ENDFOR
        \IF{$c_1 == None$}
        \STATE \textbf{continue}
        \ENDIF
        \STATE $\mathcal{C}_{i.d.} = \mathcal{C}_{i.d.} - \{c_1\}$.
        \STATE $\mathcal{C}_{comp} = \mathcal{C}_{comp} + \{c_1\}$.
        \FOR{$c \in \mathcal{C}_{comp}$}
        \IF{$d_m < D(\mathcal{C}_{i.d.} + \{c\}, \mathcal{C}_{comp} - \{c\})$}
        \STATE $d_m = D(\mathcal{C}_{i.d.} + \{c\}, \mathcal{C}_{comp} - \{c\})$.
        \STATE $\mathcal{C}_{i.d.} = \mathcal{C}_{i.d.} + \{c_1\}$.
        \STATE $\mathcal{C}_{comp} = \mathcal{C}_{comp} - \{c_1\}$.
        \STATE \textbf{break}
        \ENDIF
        \ENDFOR
        \ENDWHILE
        \FOR{$d_m \geq \eta$}
        \STATE Add ($\mathcal{C}_{i.d.}$,$\mathcal{C}_{comp}$) into \textit{result}.
        \ENDFOR
        \ENDWHILE
        \RETURN \textit{result}
    \end{algorithmic}
    \label{alg:ACD}
\end{algorithm}

\begin{algorithm}
    \caption{Meta-MCTG}
    \begin{algorithmic}[1]
        \REQUIRE Training set $\mathcal{D}_{train}$
        \REQUIRE Base Method $\mathcal{M}$
        \REQUIRE Learning rate $\alpha,\beta$, batch size $m$
        \WHILE{not done}
        \STATE Sample $m$ data as the training batch $\mathcal{B}_{train}=(c_i^{train}, x_i^{train})_{i=1}^m$ from $\mathcal{D}_{train}$.
        \STATE Construct pseudo-compositional batch $\mathcal{B}_{pcomp}=(c_i^{pcomp}, x_i^{comp})_{i=1}^m$ by sampling another $m$ data from $\mathcal{D}_{train}$, where $\{c_i^{train}\}_{i=1}^m \cap \{c_i^{pcomp}\}_{i=1}^m = \emptyset$ while each single attribute condition in $\mathcal{B}_{pseudo-comp}$ must appear in the $\mathcal{B}_{train}$.
        \STATE Compute training loss  $\mathcal{L}_{train}^\mathcal{M}$ through Objective~\ref{obj:l_train}.
        \STATE Compute $\theta_1$ through Equation~\ref{eq:update_first}. (while not really update $\theta$ to $\theta_1$)
        \STATE Temporarily use $\theta_1$ in the language model.
        \STATE Compute pseduo compositional generalization loss $\mathcal{L}_{p-comp}^\mathcal{M}$ through Objective~\ref{obj:pseudo-comp}.
        \STATE Compute total loss $\mathcal{L}_{total}^\mathcal{M} $ through Objective~\ref{obj:total}.
        \STATE Update $\theta$ to $\theta'$ through Equation~\ref{eq:update_final}
        \ENDWHILE
        
    \end{algorithmic}
    \label{alg:meta-learning}
\end{algorithm}

\section{Additional Results}
\subsection{Why do Separate-Training-based methods perform badly in compositional testing?}
\label{appendix:seperate training}
In this section, we briefly discuss the reasons why the seperate-training-based MCTG methods fail in compositional testing. We take \textit{Dis-Lens}~\cite{gu-etal-2022-distributional} as an example to illustrate. This type of method encodes each single attribute data into a latent vector space, and then constructs the intersection of different attribute latent vector areas through loss function constraints, and finally guides GPT-2 to generate multi-aspect text by searching for the intersection of different attribute spaces. The essential reason why this method can work is that the training dataset itself has multiple attributes. For example, the data corresponding to the latent space intersection constructed with positive emotion data and sports theme data actually has these two attributes. Therefore, when using a multi-attribute dataset to train the latent vector space, the attribute combinations corresponding to the constrained intersection space are the attribute combinations contained in the training set, and will not produce attribute combinations that do not exist in the training set.

Specifically, we use a \textit{Few-Shot} split of the dataset \textit{Mixture} to conduct experiments, reducing the dimensionality of hidden vectors to a two-dimensional plane through PCA and performing visualization processing. 
There are four attribute combinations in the training set which are "Negative-movies", "Negative-opener", "Negative-tablets", and "Positive-auto". 
\begin{figure}[ht]
    \raggedright
    \includegraphics[width=0.99\linewidth, keepaspectratio=true]{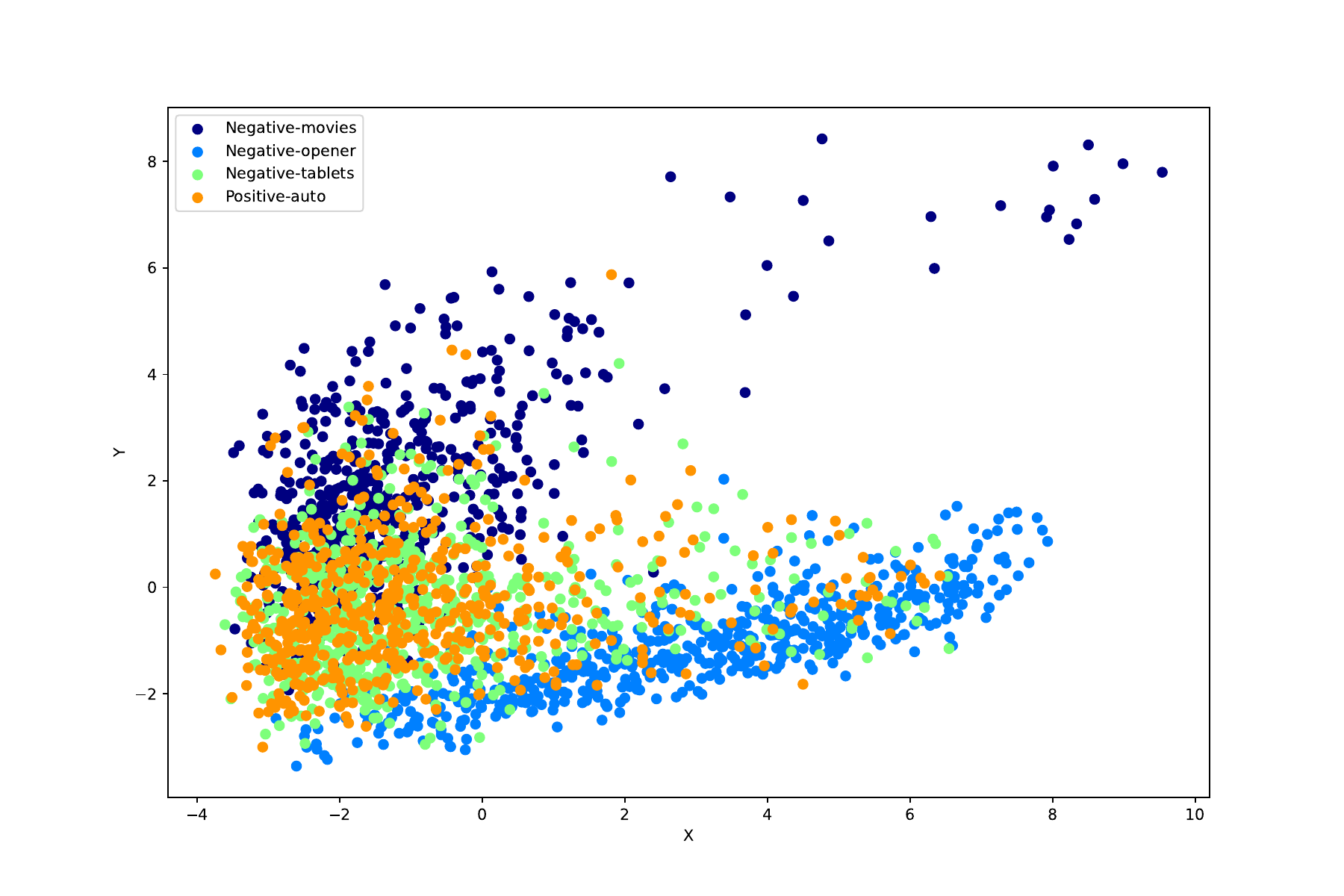}
    \caption{Visualization of \textit{Dis-lens} in \textit{Mixture} dataset before training with multi-aspect label.}
    \label{fig:lens-visual-multi-before}
\end{figure}
\begin{figure}[ht]
    \raggedright
    \includegraphics[width=0.99\linewidth, keepaspectratio=true]{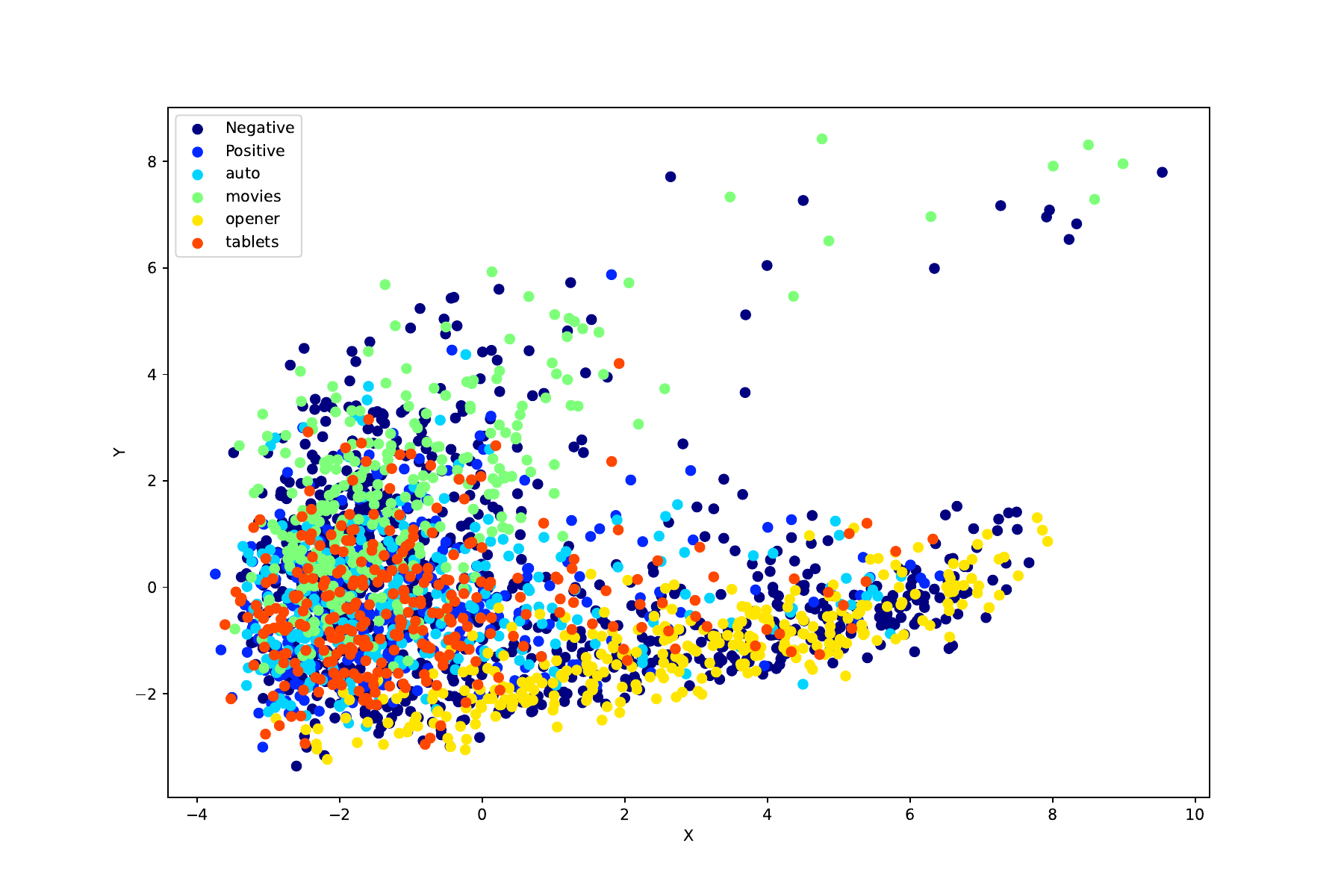}
    \caption{Visualization of \textit{Dis-lens} in \textit{Mixture} dataset before training with single-aspect label.}
    \label{fig:lens-visual-single-before}
\end{figure}
\begin{figure}[ht]
    \raggedright
    \includegraphics[width=0.99\linewidth, keepaspectratio=true]{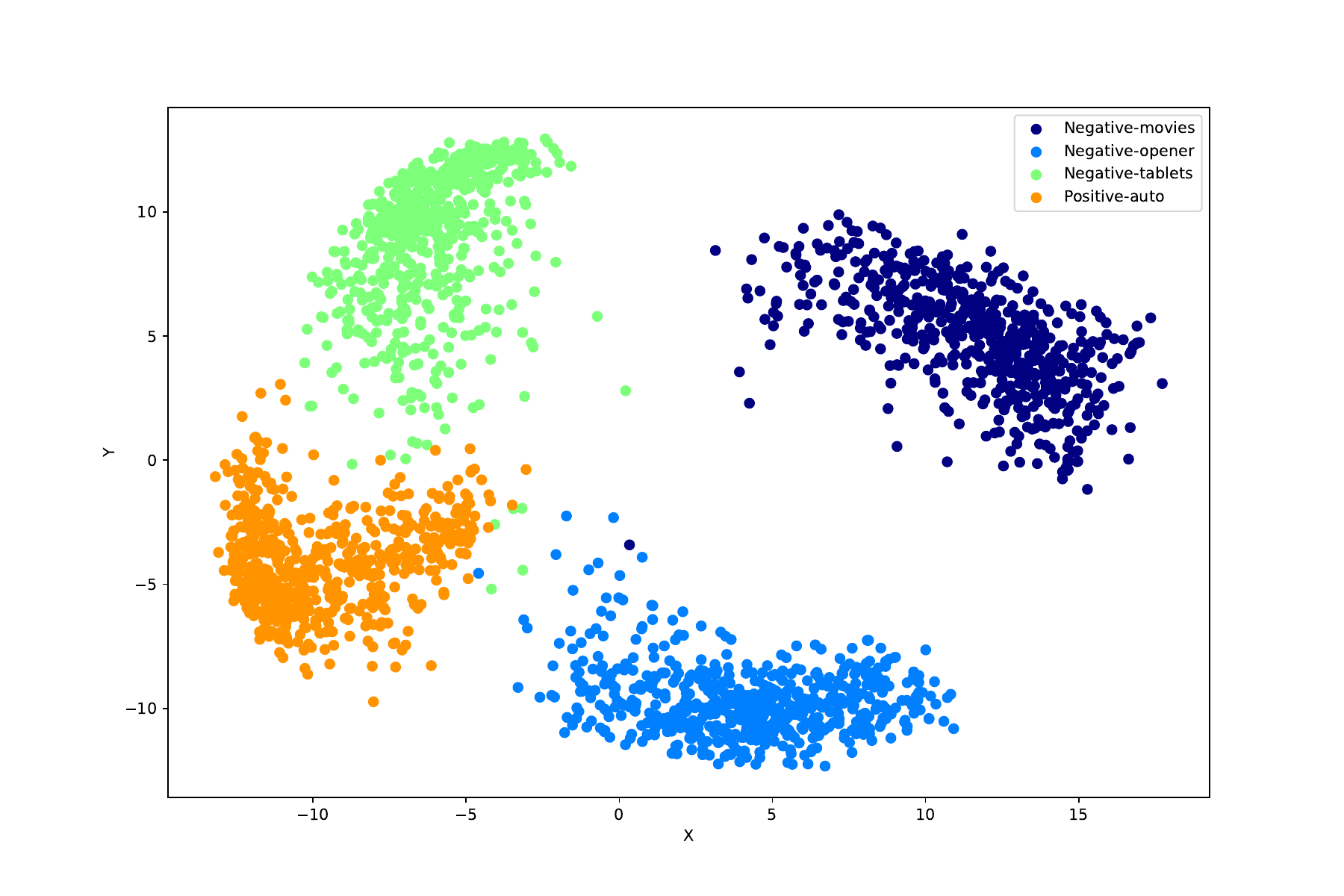}
    \caption{Visualization of \textit{Dis-lens} in \textit{Mixture} dataset after training with multi-aspect label.}
    \label{fig:lens-visual-multi-after}
\end{figure}
\begin{figure}[ht]
    \raggedright
    \includegraphics[width=0.99\linewidth, keepaspectratio=true]{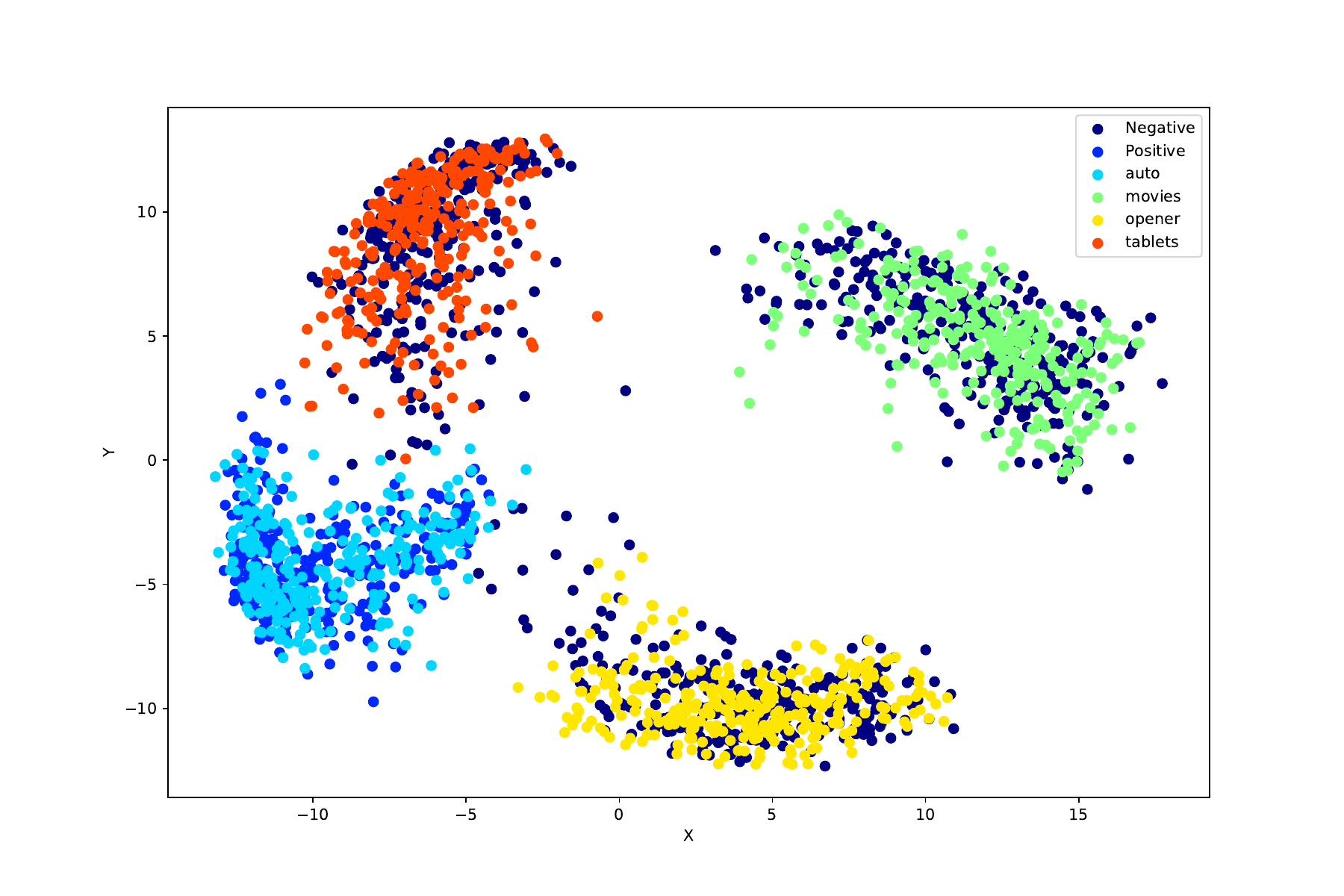}
    \caption{Visualization of \textit{Dis-lens} in \textit{Mixture} dataset after training with single-aspect label.}
    \label{fig:lens-visual-single-after}
\end{figure}
% The visualization results before training are shown in Figure \ref{fig:lens-visual-multi-before} and Figure \ref{fig:lens-visual-single-before}. 
% Figure \ref{fig:lens-visual-multi-before} is marked with multi-aspect labels, and Figure \ref{fig:lens-visual-single-before} is marked with single-aspect labels. The visualization results after training are shown in Figure \ref{fig:lens-visual-multi-after} and Figure \ref{fig:lens-visual-single-after}.
Figure \ref{fig:lens-visual-multi-before}, \ref{fig:lens-visual-single-before} depict the results of pre-training visualizations, while Figure \ref{fig:lens-visual-multi-after}, \ref{fig:lens-visual-single-after} show the results of post-training's counterpart. Figure \ref{fig:lens-visual-multi-before}, \ref{fig:lens-visual-multi-after} are annotated with multi-aspect labels, whereas Figure \ref{fig:lens-visual-single-before}, \ref{fig:lens-visual-single-after} are annotated with single-aspect labels.
From these four figures, we can find that after training, the hidden vector spaces corresponding to different single attributes have converged, and the intersection of four multi-attribute latent vector spaces has been formed.
However, through Figure \ref{fig:lens-visual-multi-after}, it can be found that these four intersections exactly correspond to the four attribute combinations contained in the training set, and the intersection of the latent vector spaces of the four compositional attribute combinations ("Negative-auto", "Positive-movies", "Positive-opener", and "Positive-tablets") in Figure \ref{fig:lens-visual-single-after} basically does not exist. This explains why such methods fail in compositional testing.

\subsection{Analysis Experiments}
\label{appendix:analysis exps}
In this section, we conduct visualization experiments on the Meta-MCTG framework we proposed, indirectly verifying its effectiveness. Considering that the joint-training-based MCTG methods tend to overfit the control parameters to the \textit{in-distribution} (I.D.) attribute combinations, this implies that for \textit{compositional} (Comp.) attribute combinations, their control parameters are relatively close to those of \textit{in-distribution}. Therefore, we approach this from the perspective of control parameters, calculating the $L1$ distance $L1_{base}, L1_{meta}$ and cosine similarity $Cos_{base}, Cos_{meta}$ between the control parameters before and after the introduction of the Meta-MCTG framework, and use the difference $diff_{L1}=\frac{L1_{meta}-L1_{base}}{L1_{meta}}\times 100$, $diff_{Cos}=-\frac{Cos_{meta}-Cos_{base}}{Cos_{meta}}\times 100$ between the two as the data for visualization. 

\begin{figure*}[ht]
    \raggedright
    \includegraphics[width=0.99\linewidth, keepaspectratio=true]{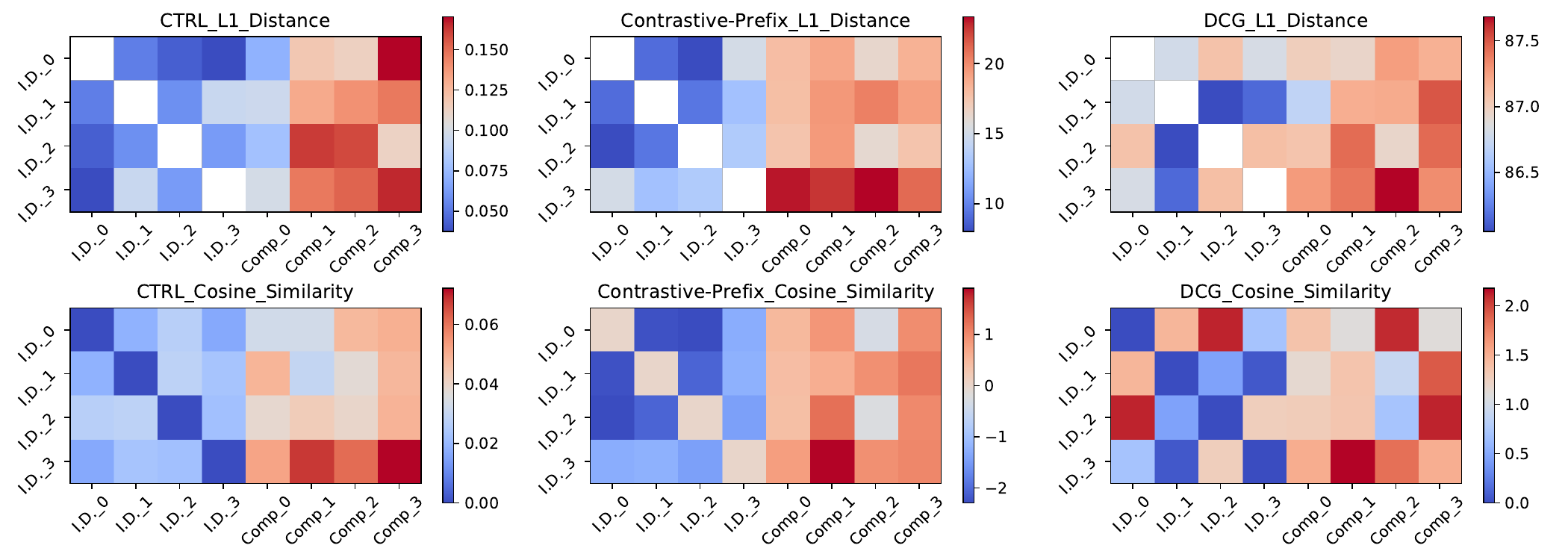}
    \caption{Difference of the distances ($d_{cos}=1-cos<h_1,h_2>$, $d_{l1}=|h_1-h_2|$) between attribute combinations in the representation space ($h_1, h_2$) with \textit{Meta-CTRL} \textit{Meta Contrastive Prefix}, \textit{Meta-DCG} and the origin version of \textit{CTRL, Contrastive Prefix, DCG} in dataset \textit{YELP}.}
    \label{fig:analysis-meta-yelp}
\end{figure*}

\begin{figure*}[ht]
    \raggedright
    \includegraphics[width=0.99\linewidth, keepaspectratio=true]{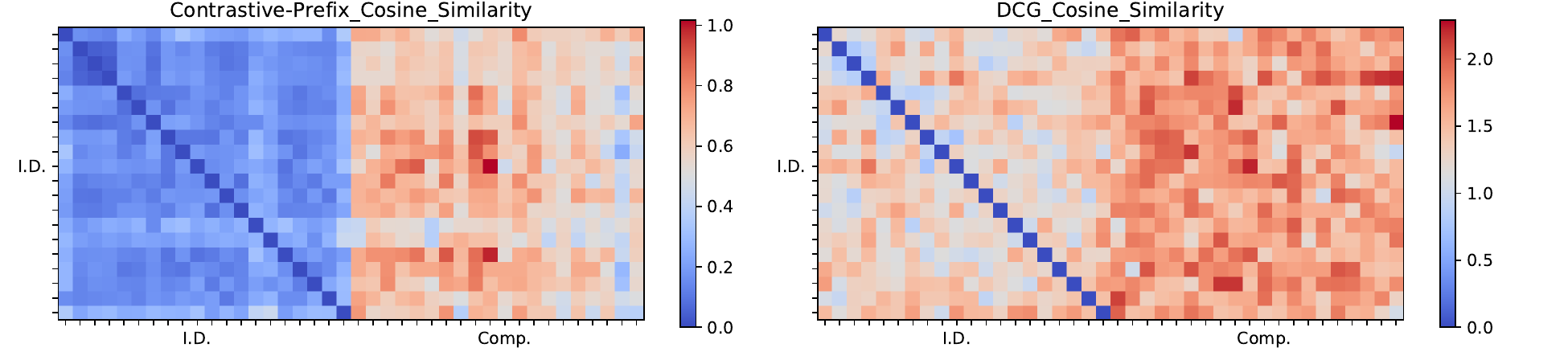}
    \caption{Difference of the distances ($d_{cos}=1-cos<h_1, h_2>$) between attribute combinations in the representation space ($h_1, h_2$) with \textit{Meta Contrastive Prefix}, \textit{Meta-DCG} and the origin version of \textit{Contrastive Prefix, DCG} in dataset \textit{Fyelp}.}
    \label{fig:analysis-meta-fyelp}
\end{figure*}

We select \textit{CTRL}~\cite{keskar2019ctrl}
, \textit{DCG}~\cite{seen-to-unseen-acl2023}, and \textit{Contrastive Prefix}~\cite{qian-etal-2022-controllable} and conduct our visualization experiments on \textit{ACD} protocol of \textit{YELP}~\cite{YELP} and \textit{Fyelp}~\cite{Fyelp} datasets. For \textit{CTRL}, we use the mean embeddings of its attribute tokens (i.e., control codes) as the control parameters. For \textit{DCG}, we use the mean embedding obtained by encoding the attribute tokens through a fully connected layer as the control parameters. For \textit{Contrastive Prefix}, we use the mean embedding of the prefix keys and prefix values of the corresponding attributes in the last layer of the GPT-2 as the control parameters. On the \textit{YELP} dataset, there are a total of 8 attribute combinations, including 4 in-distribution and 4 compositional. For the control parameters under 8 control conditions, we compute the difference $diff_{L1}$ and $diff_{Cos}$ between each pair and obtain two $4\times 8$ heatmaps for each baseline. Similarly, for the \textit{Fyelp} dataset, we can get two $20\times 40$ heatmaps for each baseline. The results are shown in Figure \ref{fig:analysis-meta-yelp} and Figure \ref{fig:analysis-meta-fyelp}. The visual results show that the control parameters after the Meta-MCTG training framework can better distinguish between the in-distribution and compositional parts, thus confirming the effectiveness of the Meta-MCTG framework.


\subsection{Detailed Results on the Single Dataset}
In this section, we provide detailed experimental results of all baselines (eight MCTG baselines and two LLMs) in CompMCTG Benchmark in 4 datasets. In these tables, the first column contains the protocol, including \textit{Original}, \textit{HoldOut}, \textit{ACD}, and \textit{FewShot} (\textit{Amazon} and \textit{Mixture} datasets do not have \textit{ACD} protocol). \textit{Holdout}, \textit{ACD}, and \textit{FewShot} respectively divide the \textit{in-distribution} (\textit{I.D.}) results and \textit{compositional} (\textit{Comp.}) results. The second column is the method name and the next two to four columns are the accuracy of the corresponding attributes. Specifically, $Acc_s, Acc_g, Acc_c, Acc_t$ in \textit{Fyelp} are the accuracy of sentiment, gender, cuisine, and tense respectively. $Acc_s, Acc_t$ in \textit{Amazon} are the accuracy of sentiment and topic respectively. $Acc_s, Acc_p, Acc_t$ in \textit{YELP} are the accuracy of sentiment, person, and tense respectively. $Acc_s, Acc_t$ in \textit{Mixture} are the accuracy of sentiment and topic respectively. $Acc_{avg}$ is the average of accuracy in previous columns. $Acc_{gap}$ is calculated from the \textit{I.D.} and \textit{Comp.} of the corresponding protocol. The specific formula is $Acc_{gap} = \frac{Acc_{I.D.}-Acc_{Comp.}}{Acc_{I.D.}} \times 100\%$. $PPL$ represents perplexity and $Dist3$ is the value of 3-ngrams. All the detailed experimental results can be found in Table \ref{main-results-Fyelpv3-ctrl} to Table \ref{main-results-Mixture-gpt-3.5-anony}.

\begin{table*}\centering
\begin{tabularx}{\linewidth}{|l*{9}{X}|}
\hlineB{2.5}
\textbf{Protocol} & \textbf{Method} & $\boldsymbol{Acc_s}$ & $\boldsymbol{Acc_g}$ & $\boldsymbol{Acc_c}$ & $\boldsymbol{
Acc_t}$ & $\boldsymbol{Acc_{avg}}$ & $\boldsymbol{Acc_{gap}}$ & $\boldsymbol{PPL \downarrow}$ & $\boldsymbol{Dist3}$\\
\hline\hline
\multirow{1}{*}{\textit{Original}} & \textit{CTRL} & 88.28 & 60.13 & 60.38 & 67.29 & 69.02 & - & 45.69 & 0.675 \\
\hline
\multirow{1}{*}{\textit{HoldOut-I.D.}} & \textit{CTRL} & 88.42 & 60.88 & 60.53 & 67.89 & 69.43 & \multirow{2}{*}{1.64} & 45.95 & 0.675 \\
\multirow{1}{*}{\textit{HoldOut-Comp.}} & \textit{CTRL} & 87.88 & 59.65 & 59.02 & 66.61 & 68.29 & & 45.61 & 0.676 \\
\hline
\multirow{1}{*}{\textit{ACD-I.D.}} & \textit{CTRL} & 87.83 & 60.25 & 59.45 & 69.35 & 69.22 & \multirow{2}{*}{5.65} & 45.60 & 0.684\\
\multirow{1}{*}{\textit{ACD-Comp.}} & \textit{CTRL} & 87.00 & 55.35 & 58.93 & 59.95 & 65.31 & & 45.86 & 0.678 \\
\hline
\multirow{1}{*}{\textit{FewShot-I.D.}} & \textit{CTRL} & 84.06 & 70.03 & 54.71 & 69.11 & 69.48 & \multirow{2}{*}{13.95} & 45.01 & 0.683\\
\multirow{1}{*}{\textit{FewShot-Comp.}} & \textit{CTRL} & 82.37 & 48.35 & 55.75 & 52.70 & 59.79 & & 44.33 & 0.684\\
\hlineB{2.5}
\end{tabularx}
\caption{\label{main-results-Fyelpv3-ctrl}
The result of baseline \textit{CTRL}~\cite{keskar2019ctrl} in dataset \textit{Fyelp}.
}
\end{table*}

\begin{table*}
\centering
\begin{tabularx}{\linewidth}{|l*{7}{X}|}
\hlineB{2.5}
\textbf{Protocol} & \textbf{Method} & $\boldsymbol{Acc_s}$ & $\boldsymbol{Acc_t}$ & $\boldsymbol{Acc_{avg}}$ & $\boldsymbol{Acc_{gap}}$ & $\boldsymbol{PPL \downarrow}$ & $\boldsymbol{Dist3}$\\
\hline\hline
\textit{Original} & \textit{CTRL} & 88.43 & 71.76 & 80.10 & - & 37.97 & 0.731\\
\hline
\textit{HoldOut-I.D.} & \textit{CTRL} & 88.77 & 72.00 & 80.39 & \multirow{2}{*}{2.67} & 37.87 & 0.734\\
\textit{HoldOut-Comp.} & \textit{CTRL} & 86.55 & 69.93 & 78.24 &  & 38.10 & 0.736\\
\hline
\textit{FewShot-I.D.} & \textit{CTRL} & 88.60 & 70.29 & 79.45 & \multirow{2}{*}{9.13} & 37.40 & 0.734\\
\textit{FewShot-Comp.} & \textit{CTRL} & 76.53 & 67.87 & 72.20 &  & 37.50 & 0.740\\
\hlineB{2.5}
\end{tabularx}
\caption{\label{main-results-Amazon-ctrl}
The result of baseline \textit{CTRL}~\cite{keskar2019ctrl} in dataset \textit{Amazon}.
}
\end{table*}

\begin{table*}
\setlength\tabcolsep{6.9pt}
\centering
\begin{tabularx}{\linewidth}{|l*{8}{X}|}
\hlineB{2.5}
\textbf{Protocol} & \textbf{Method} & $\boldsymbol{Acc_s}$ & $\boldsymbol{Acc_p}$ & $\boldsymbol{Acc_t}$ & $\boldsymbol{Acc_{avg}}$ & $\boldsymbol{Acc_{gap}}$ & $\boldsymbol{PPL \downarrow}$ & $\boldsymbol{Dist3}$\\
\hline\hline
\multirow{1}{*}{\textit{Original}} & \textit{CTRL} & 90.07 & 75.71 & 89.82 & 85.20 & - & 84.94 & 0.356\\
\hline
\multirow{1}{*}{\textit{HoldOut-I.D.}} & \textit{CTRL} & 91.47 & 74.28 & 89.72 & 85.16 & \multirow{2}{*}{3.69} & 72.20 & 0.360\\
\multirow{1}{*}{\textit{HoldOut-Comp.}} & \textit{CTRL} & 89.89 & 69.00 & 87.18 & 82.02 &  & 73.74 & 0.368\\
\hline
\multirow{1}{*}{\textit{ACD-I.D.}} & \textit{CTRL} & 91.76 & 74.35 & 90.46 & 85.52 & \multirow{2}{*}{12.73} & 76.06  & 0.348\\
\multirow{1}{*}{\textit{ACD-Comp.}} & \textit{CTRL} & 88.06 & 55.81 & 80.03 & 74.63 &  & 75.46 & 0.359\\
\hline
\multirow{1}{*}{\textit{FewShot-I.D.}} & \textit{CTRL} & 90.05 & 76.55 & 89.73 & 85.44 & \multirow{2}{*}{25.02} & 63.72 & 0.269\\
\multirow{1}{*}{\textit{FewShot-Comp.}} & \textit{CTRL} & 81.90 & 47.54 & 62.73 & 64.06 &  & 64.74 & 0.338\\
\hlineB{2.5}
\end{tabularx}
\caption{\label{main-results-Yelp-ctrl}
The result of baseline \textit{CTRL}~\cite{keskar2019ctrl} in dataset \textit{YELP}.
}
\end{table*}

\begin{table*}
\centering
\begin{tabularx}{\linewidth}{|l*{7}{X}|}
\hlineB{2.5}
\textbf{Protocol} & \textbf{Method} & $\boldsymbol{Acc_s}$ & $\boldsymbol{Acc_{tc}}$ & $\boldsymbol{Acc_{avg}}$ & $\boldsymbol{Acc_{gap}}$ & $\boldsymbol{PPL \downarrow}$ & $\boldsymbol{Dist3}$\\
\hline\hline
\textit{Original} & \textit{CTRL} & 76.14 & 88.04 & 82.09 & - & 48.11 & 0.736\\
\hline
\textit{HoldOut-I.D.} & \textit{CTRL} & 72.45 & 88.66 & 80.56 & \multirow{2}{*}{10.85} & 48.82 & 0.723\\
\textit{HoldOut-Comp.} & \textit{CTRL} & 66.46 & 77.18 & 71.82 &  & 47.46 & 0.755\\
\hline
\textit{FewShot-I.D.} & \textit{CTRL} & 68.71 & 85.51 & 77.11 & \multirow{2}{*}{12.19} & 47.79 & 0.699\\
\textit{FewShot-Comp.} & \textit{CTRL} & 61.21 & 74.20 & 67.71 &  & 46.31 & 0.709\\
\hlineB{2.5}
\end{tabularx}
\caption{\label{main-results-Mixture-ctrl}
The result of baseline \textit{CTRL}~\cite{keskar2019ctrl} in dataset \textit{Mixture}.
}
\end{table*}

\begin{table*}
\centering
\begin{tabularx}{\linewidth}{|l*{9}{X}|}
\hlineB{2.5}
\textbf{Protocol} & \textbf{Method} & $\boldsymbol{Acc_s}$ & $\boldsymbol{Acc_g}$ & $\boldsymbol{Acc_c}$ & $\boldsymbol{
Acc_t}$ & $\boldsymbol{Acc_{avg}}$ & $\boldsymbol{Acc_{gap}}$ & $\boldsymbol{PPL \downarrow}$ & $\boldsymbol{Dist3}$\\
\hline\hline
\multirow{1}{*}{\textit{Original}} & \textit{CatPro} & 84.65 & 54.43 & 53.72 & 63.91 & 64.18 & - & 70.58 & 0.726\\
\hline
\multirow{1}{*}{\textit{HoldOut-I.D.}} & \textit{CatPro} & 84.45 & 54.76 & 56.80 & 64.64 & 65.16 & \multirow{2}{*}{0.91} & 69.71 & 0.726\\
\multirow{1}{*}{\textit{HoldOut-Comp.}} & \textit{CatPro} & 83.82 & 54.07 & 56.04 & 64.36 & 64.57 &  & 69.48 & 0.725 \\
\hline
\multirow{1}{*}{\textit{ACD-I.D.}} & \textit{CatPro} & 83.45 & 54.04 & 47.33 & 61.21 & 61.51 & \multirow{2}{*}{10.96} & 69.30 & 0.735\\
\multirow{1}{*}{\textit{ACD-Comp.}} & \textit{CatPro} & 71.26 & 50.11 & 35.36 & 62.35 & 54.77 &  & 63.83  & 0.750\\
\hline
\multirow{1}{*}{\textit{FewShot-I.D.}} & \textit{CatPro} & 79.31 & 66.71 & 37.54 & 63.00 & 61.64 & \multirow{2}{*}{26.10} & 70.94 & 0.741\\
\multirow{1}{*}{\textit{FewShot-Comp.}} & \textit{CatPro} & 46.04 & 48.28 & 24.11 & 63.75 & 45.55 &  & 68.16 & 0.740\\
\hlineB{2.5}
\end{tabularx}
\caption{\label{main-results-Fyelpv3-catprompt}
The result of baseline \textit{CatPrompt}~\cite{yang-etal-2023-tailor} in dataset \textit{Fyelp}.
}
\end{table*}

\begin{table*}
\centering
\begin{tabularx}{\linewidth}{|l*{7}{X}|}
\hlineB{2.5}
\textbf{Protocol} & \textbf{Method} & $\boldsymbol{Acc_s}$ & $\boldsymbol{Acc_t}$ & $\boldsymbol{Acc_{avg}}$ & $\boldsymbol{Acc_{gap}}$ & $\boldsymbol{PPL \downarrow}$ & $\boldsymbol{Dist3}$\\
\hline\hline
\textit{Original} & \textit{CatPro} & 82.31 & 60.88 & 71.60 & - & 55.08 & 0.734\\
\hline
\textit{HoldOut-I.D.} & \textit{CatPro} & 83.00 & 56.99 & 70.00 & \multirow{2}{*}{9.89} & 57.50 & 0.701\\
\textit{HoldOut-Comp.} & \textit{CatPro} & 72.86 & 53.29 & 63.08 &  & 50.39 & 0.727\\
\hline
\textit{FewShot-I.D.} & \textit{CatPro} & 77.95 & 44.64 & 61.30 & \multirow{2}{*}{35.42} & 55.63 & 0.658\\
\textit{FewShot-Comp.} & \textit{CatPro} & 48.22 & 30.96 & 39.59 &  & 41.59 & 0.717\\
\hlineB{2.5}
\end{tabularx}
\caption{\label{main-results-Amazon-catprompt}
The result of baseline \textit{CatPrompt}~\cite{yang-etal-2023-tailor} in dataset \textit{Amazon}.
}
\end{table*}

\begin{table*}
\centering
\begin{tabularx}{\linewidth}{|l*{8}{X}|}
\hlineB{2.5}
\textbf{Protocol} & \textbf{Method} & $\boldsymbol{Acc_s}$ & $\boldsymbol{Acc_p}$ & $\boldsymbol{Acc_t}$ & $\boldsymbol{Acc_{avg}}$ & $\boldsymbol{Acc_{gap}}$ & $\boldsymbol{PPL \downarrow}$ & $\boldsymbol{Dist3}$\\
\hline\hline
\multirow{1}{*}{\textit{Original}} & \textit{CatPro} & 78.93 & 51.43 & 75.43 & 68.60 & - & 83.96 & 0.467\\
\hline
\multirow{1}{*}{\textit{HoldOut-I.D.}} & \textit{CatPro} & 76.04 & 51.67 & 74.86 & 67.52 & \multirow{2}{*}{4.83} & 86.92 & 0.462\\
\multirow{1}{*}{\textit{HoldOut-Comp.}} & \textit{CatPro} & 70.68 & 50.18 & 71.93 & 64.26 &  & 86.79 & 0.467\\
\hline
\multirow{1}{*}{\textit{ACD-I.D.}} & \textit{CatPro} & 72.24 & 52.88 & 73.23 & 66.12 & \multirow{2}{*}{14.10} & 118.02 & 0.634\\
\multirow{1}{*}{\textit{ACD-Comp.}} & \textit{CatPro} & 47.54 & 49.75 & 73.12 & 56.80 &  & 105.37 & 0.657\\
\hline
\multirow{1}{*}{\textit{FewShot-I.D.}} & \textit{CatPro} & 79.86 & 57.07 & 84.21 & 73.71 & \multirow{2}{*}{21.39} & 378.69 & 0.448 \\
\multirow{1}{*}{\textit{FewShot-Comp.}} & \textit{CatPro} & 45.43 & 49.73 & 78.65 & 57.94  &  & 349.24 & 0.585\\
\hlineB{2.5}
\end{tabularx}
\caption{\label{main-results-Yelp-catprompt}
The result of baseline \textit{CatPrompt}~\cite{yang-etal-2023-tailor} in dataset \textit{YELP}.
}
\end{table*}

\begin{table*}
\centering
\begin{tabularx}{\linewidth}{|l*{7}{X}|}
\hlineB{2.5}
\textbf{Protocol} & \textbf{Method} & $\boldsymbol{Acc_s}$ & $\boldsymbol{Acc_{tc}}$ & $\boldsymbol{Acc_{avg}}$ & $\boldsymbol{Acc_{gap}}$ & $\boldsymbol{PPL \downarrow}$ & $\boldsymbol{Dist3}$\\
\hline\hline
\textit{Original} & \textit{CatPro} & 51.61 & 50.86 & 51.24 & - & 88.51 & 0.641\\
\hline
\textit{HoldOut-I.D.} & \textit{CatPro} & 51.53 & 54.67 & 53.10 & \multirow{2}{*}{7.01} & 79.25 & 0.654\\
\textit{HoldOut-Comp.} & \textit{CatPro} & 50.36 & 48.39 & 49.38 &  & 69.87 & 0.705\\
\hline
\textit{FewShot-I.D.} & \textit{CatPro} & 54.52 & 51.91 & 53.22 & \multirow{2}{*}{21.42} & 149.37 & 0.679\\
\textit{FewShot-Comp.} & \textit{CatPro} & 53.11 & 30.52 & 41.82 &  & 63.00 & 0.629\\
\hlineB{2.5}
\end{tabularx}
\caption{\label{main-results-Mixture-catprompt}
The result of baseline \textit{CatPrompt}~\cite{yang-etal-2023-tailor} in dataset \textit{Mixture}.
}
\end{table*}

\begin{table*}
\centering
\begin{tabularx}{\linewidth}{|l*{9}{X}|}
\hlineB{2.5}
\textbf{Protocol} & \textbf{Method} & $\boldsymbol{Acc_s}$ & $\boldsymbol{Acc_g}$ & $\boldsymbol{Acc_c}$ & $\boldsymbol{
Acc_t}$ & $\boldsymbol{Acc_{avg}}$ & $\boldsymbol{Acc_{gap}}$ & $\boldsymbol{PPL \downarrow}$ & $\boldsymbol{Dist3}$\\
\hline\hline
\multirow{1}{*}{\textit{Original}} & \textit{DCG} & 90.18 & 56.68 & 56.50 & 62.34 & 66.43 & - & 53.31 & 0.688\\
\hline
\multirow{1}{*}{\textit{HoldOut-I.D.}} & \textit{DCG} & 90.09 & 56.33 & 57.21 & 62.33 & 66.49 & \multirow{2}{*}{0.15} & 53.50 & 0.702\\
\multirow{1}{*}{\textit{HoldOut-Comp.}} & \textit{DCG} & 90.29 & 56.39 & 57.00 & 61.88 & 66.39 &  & 53.52 & 0.704\\
\hline
\multirow{1}{*}{\textit{ACD-I.D.}} & \textit{DCG} & 90.07 & 55.55 & 56.44 & 61.96 & 66.01 & \multirow{2}{*}{1.97} & 53.29 & 0.702\\
\multirow{1}{*}{\textit{ACD-Comp.}} & \textit{DCG} & 89.73 & 55.04 & 54.99 & 59.07 & 64.71 &  & 53.67 & 0.704\\
\hline
\multirow{1}{*}{\textit{FewShot-I.D.}} & \textit{DCG} & 89.00 & 68.26 & 50.37 & 65.63 & 68.32 & \multirow{2}{*}{25.91} & 53.30 & 0.704\\
\multirow{1}{*}{\textit{FewShot-Comp.}} & D\textit{}CG & 57.34 & 49.02 & 41.68 & 54.42 & 50.62 &  & 52.82 & 0.695\\
\hlineB{2.5}
\end{tabularx}
\caption{\label{main-results-Fyelpv3-dcg}
The result of baseline \textit{DCG}~\cite{seen-to-unseen-acl2023} in dataset \textit{Fyelp}.
}
\end{table*}

\begin{table*}
\centering
\begin{tabularx}{\linewidth}{|l*{7}{X}|}
\hlineB{2.5}
\textbf{Protocol} & \textbf{Method} & $\boldsymbol{Acc_s}$ & $\boldsymbol{Acc_t}$ & $\boldsymbol{Acc_{avg}}$ & $\boldsymbol{Acc_{gap}}$ & $\boldsymbol{PPL \downarrow}$ & $\boldsymbol{Dist3}$\\
\hline\hline
\textit{Original} & \textit{DCG} & 91.00 & 77.95 & 84.48 & - & 46.66 & 0.723\\
\hline
\textit{HoldOut-I.D.} & \textit{DCG} & 91.13 & 78.29 & 84.71 & \multirow{2}{*}{0.24} & 47.20 & 0.727\\
\textit{HoldOut-Comp.} & \textit{DCG} & 91.50 & 77.52 & 84.51 &  & 47.09 & 0.723\\
\hline
\textit{FewShot-I.D.} & \textit{DCG} & 91.66 & 76.63 & 84.15 & \multirow{2}{*}{18.86} & 48.05 & 0.727\\
\textit{FewShot-Comp.} & \textit{DCG} & 69.86 & 66.70 & 68.28 &  & 48.36 & 0.720\\
\hlineB{2.5}
\end{tabularx}
\caption{\label{main-results-Amazon-dcg}
The result of baseline \textit{DCG}~\cite{seen-to-unseen-acl2023} in dataset \textit{Amazon}.
}
\end{table*}

\begin{table*}
\centering
\begin{tabularx}{\linewidth}{|l*{8}{X}|}
\hlineB{2.5}
\textbf{Protocol} & \textbf{Method} & $\boldsymbol{Acc_s}$ & $\boldsymbol{Acc_p}$ & $\boldsymbol{Acc_t}$ & $\boldsymbol{Acc_{avg}}$ & $\boldsymbol{Acc_{gap}}$ & $\boldsymbol{PPL \downarrow}$ & $\boldsymbol{Dist3}$\\
\hline\hline
\multirow{1}{*}{\textit{Original}} & \textit{DCG} & 95.75 & 66.57 & 91.07 & 84.46 & - & 57.08 & 0.706\\
\hline
\multirow{1}{*}{\textit{HoldOut-I.D.}} & \textit{DCG} & 94.49 & 64.33 & 90.38 & 83.07 & \multirow{2}{*}{3.35} & 79.05 & 0.703\\
\multirow{1}{*}{\textit{HoldOut-Comp.}} & \textit{DCG} & 94.50 & 58.75 & 87.61 & 80.29 &  & 80.58 & 0.721\\
\hline
\multirow{1}{*}{\textit{ACD-I.D.}} & \textit{DCG} & 92.64 & 61.59 & 88.79 & 81.01 & \multirow{2}{*}{6.09} & 79.86 & 0.668\\
\multirow{1}{*}{\textit{ACD-Comp.}} & \textit{DCG} & 88.06 & 57.90 & 82.28 & 76.08 &  & 84.30 & 0.686\\
\hline
\multirow{1}{*}{\textit{FewShot-I.D.}} & \textit{DCG} & 90.82 & 62.21 & 85.93 & 79.65 & \multirow{2}{*}{29.57} & 93.66 & 0.510\\
\multirow{1}{*}{\textit{FewShot-Comp.}} & \textit{DCG} & 55.15 & 52.51 & 60.63 & 56.10 &  & 111.03 & 0.653\\
\hlineB{2.5}
\end{tabularx}
\caption{\label{main-results-Yelp-dcg}
The result of baseline \textit{DCG}~\cite{seen-to-unseen-acl2023} in dataset \textit{YELP}.
}
\end{table*}

\begin{table*}
\centering
\begin{tabularx}{\linewidth}{|l*{7}{X}|}
\hlineB{2.5}
\textbf{Protocol} & \textbf{Method} & $\boldsymbol{Acc_s}$ & $\boldsymbol{Acc_{tc}}$ & $\boldsymbol{Acc_{avg}}$ & $\boldsymbol{Acc_{gap}}$ & $\boldsymbol{PPL \downarrow}$ & $\boldsymbol{Dist3}$\\
\hline\hline
\textit{Original} & \textit{DCG} & 72.07 & 96.61 & 84.34 & - & 68.44 & 0.592\\
\hline
\textit{HoldOut-I.D.} & \textit{DCG} & 73.86 & 95.35 & 84.61 & \multirow{2}{*}{10.83} & 68.45 & 0.645\\
\textit{HoldOut-Comp.} & \textit{DCG} & 56.64 & 94.25 & 75.45 &  & 76.41 & 0.715\\
\hline
\textit{FewShot-I.D.} & \textit{DCG} & 71.64 & 95.21 & 83.43 & \multirow{2}{*}{25.58} & 57.87 & 0.603\\
\textit{FewShot-Comp.} & \textit{DCG} & 40.34 & 83.83 & 62.09 &  & 60.33 & 0.670\\
\hlineB{2.5}
\end{tabularx}
\caption{\label{main-results-Mixture-dcg}
The result of baseline \textit{DCG}~\cite{seen-to-unseen-acl2023} in dataset \textit{Mixture}.
}
\end{table*}

\begin{table*}
\centering
\begin{tabularx}{\linewidth}{|l*{9}{X}|}
\hlineB{2.5}
\textbf{Protocol} & \textbf{Method} & $\boldsymbol{Acc_s}$ & $\boldsymbol{Acc_g}$ & $\boldsymbol{Acc_c}$ & $\boldsymbol{
Acc_t}$ & $\boldsymbol{Acc_{avg}}$ & $\boldsymbol{Acc_{gap}}$ & $\boldsymbol{PPL \downarrow}$ & $\boldsymbol{Dist3}$\\
\hline\hline
\multirow{1}{*}{\textit{Original}} & \textit{Fudge} & 67.49 & 51.45 & 37.07 & 59.73 & 53.94 & - & 223.31 & 0.732\\
\hline
\multirow{1}{*}{\textit{HoldOut-I.D.}} & \textit{Fudge} & 67.09 & 51.45 & 37.15 & 59.71 & 53.85 & \multirow{2}{*}{22.54} & 221.77 & 0.726 \\
\multirow{1}{*}{\textit{HoldOut-Comp.}} & \textit{Fudge} & 49.61 & 48.80 & 20.91 & 47.50 & 41.71 &  & 269.55 & 0.728\\
\hline
\multirow{1}{*}{\textit{ACD-I.D.}} & \textit{Fudge} & 67.44 & 48.58 & 36.64 & 60.15 & 53.20 & \multirow{2}{*}{24.02} & 213.12 & 0.705\\
\multirow{1}{*}{\textit{ACD-Comp.}} & \textit{Fudge} & 51.01 & 50.34 & 19.17 & 41.17 & 40.42 &  & 239.45 & 0.718\\
\hline
\multirow{1}{*}{\textit{FewShot-I.D.}} & \textit{Fudge} & 70.83 & 79.46 & 25.80 & 45.54 & 55.41 & \multirow{2}{*}{26.06} & 208.09 & 0.666 \\
\multirow{1}{*}{\textit{FewShot-Comp.}} & \textit{Fudge} & 47.87 & 45.30 & 20.27 & 50.44 & 40.97 &  & 282.25 & 0.490\\
\hlineB{2.5}
\end{tabularx}
\caption{\label{main-results-Fyelpv3-fudge}
The result of baseline \textit{Fudge}~\cite{yang-klein-2021-fudge} in dataset \textit{Fyelp}.
}
\end{table*}

\begin{table*}
\centering
\begin{tabularx}{\linewidth}{|l*{7}{X}|}
\hlineB{2.5}
\textbf{Protocol} & \textbf{Method} & $\boldsymbol{Acc_s}$ & $\boldsymbol{Acc_t}$ & $\boldsymbol{Acc_{avg}}$ & $\boldsymbol{Acc_{gap}}$ & $\boldsymbol{PPL \downarrow}$ & $\boldsymbol{Dist3}$\\
\hline\hline
\textit{Original} & \textit{Fudge} & 65.40 & 47.64 & 56.52 & - & 185.96 & 0.743\\
\hline
\textit{HoldOut-I.D.} & \textit{Fudge} & 64.71 & 47.49 & 56.10 & \multirow{2}{*}{38.89} & 192.16 & 0.738\\
\textit{HoldOut-Comp.} & \textit{Fudge} & 51.81 & 16.74 & 34.28 &  & 188.13 & 0.786\\
\hline
\textit{FewShot-I.D.} & \textit{Fudge} & 64.16 & 54.30 & 59.23 & \multirow{2}{*}{41.53} & 206.58 & 0.722\\
\textit{FewShot-Comp.} & \textit{Fudge} & 52.05 & 17.21 & 34.63 &  & 175.48 & 0.772\\
\hlineB{2.5}
\end{tabularx}
\caption{\label{main-results-Amazon-fudge}
The result of baseline \textit{Fudge}~\cite{yang-klein-2021-fudge} in dataset \textit{Amazon}.
}
\end{table*}

\begin{table*}
\centering
\begin{tabularx}{\linewidth}{|l*{8}{X}|}
\hlineB{2.5}
\textbf{Protocol} & \textbf{Method} & $\boldsymbol{Acc_s}$ & $\boldsymbol{Acc_p}$ & $\boldsymbol{Acc_t}$ & $\boldsymbol{Acc_{avg}}$ & $\boldsymbol{Acc_{gap}}$ & $\boldsymbol{PPL \downarrow}$ & $\boldsymbol{Dist3}$\\
\hline\hline
\multirow{1}{*}{\textit{Original}} & \textit{Fudge} & 63.68 & 93.79 & 84.57 & 80.68 & - & 104.33 & 0.667\\
\hline
\multirow{1}{*}{\textit{HoldOut-I.D.}} & \textit{Fudge} & 63.09 & 93.59 & 83.55 & 80.08 & \multirow{2}{*}{34.12} & 99.90 & 0.656\\
\multirow{1}{*}{\textit{HoldOut-Comp.}} & \textit{Fudge} & 50.39 & 55.25 & 52.64 & 52.76 & & 355.48 & 0.717\\
\hline
\multirow{1}{*}{\textit{ACD-I.D.}} & \textit{Fudge} & 53.24 & 86.00 & 74.31 & 71.18 & \multirow{2}{*}{24.23} & 86.50 & 0.609\\
\multirow{1}{*}{\textit{ACD-Comp.}} & \textit{Fudge} & 55.39 & 54.55 & 51.86 & 53.93 &  & 297.18 & 0.636\\
\hline
\multirow{1}{*}{\textit{FewShot-I.D.}} & \textit{Fudge} & 58.32 & 87.32 & 71.32 & 72.32 & \multirow{2}{*}{29.29} & 58.13 & 0.481\\
\multirow{1}{*}{\textit{FewShot-Comp.}} & \textit{Fudge} & 50.24 & 51.70 & 51.48 & 51.14 & & 261.71 & 0.578\\
\hlineB{2.5}
\end{tabularx}
\caption{\label{main-results-Yelp-fudge}
The result of baseline \textit{Fudge}~\cite{yang-klein-2021-fudge} in dataset \textit{YELP}.
}
\end{table*}

\begin{table*}
\centering
\begin{tabularx}{\linewidth}{|l*{7}{X}|}
\hlineB{2.5}
\textbf{Protocol} & \textbf{Method} & $\boldsymbol{Acc_s}$ & $\boldsymbol{Acc_{tc}}$ & $\boldsymbol{Acc_{avg}}$ & $\boldsymbol{Acc_{gap}}$ & $\boldsymbol{PPL \downarrow}$ & $\boldsymbol{Dist3}$\\
\hline\hline
\textit{Original} & \textit{Fudge} & 56.00 & 42.64 & 49.32 & - & 200.42 & 0.483\\
\hline
\textit{HoldOut-I.D.} & \textit{Fudge} & 54.22 & 40.51 & 47.37 & \multirow{2}{*}{16.34} & 204.05 & 0.487\\
\textit{HoldOut-Comp.} & \textit{Fudge} & 51.96 & 27.29 & 39.63 &  & 195.15 & 0.254\\
\hline
\textit{FewShot-I.D.} & \textit{Fudge} & 51.89 & 38.15 & 45.02 & \multirow{2}{*}{18.15} & 196.42 & 0.465\\
\textit{FewShot-Comp.} & \textit{Fudge} & 48.65 & 25.05 & 36.85 &  & 180.19 & 0.221\\
\hlineB{2.5}
\end{tabularx}
\caption{\label{main-results-Mixture-fudge}
The result of baseline \textit{Fudge}~\cite{yang-klein-2021-fudge} in dataset \textit{Mixture}.
}
\end{table*}

\begin{table*}
\centering
\begin{tabularx}{\linewidth}{|l*{9}{X}|}
\hlineB{2.5}
\textbf{Protocol} & \textbf{Method} & $\boldsymbol{Acc_s}$ & $\boldsymbol{Acc_g}$ & $\boldsymbol{Acc_c}$ & $\boldsymbol{
Acc_t}$ & $\boldsymbol{Acc_{avg}}$ & $\boldsymbol{Acc_{gap}}$ & $\boldsymbol{PPL \downarrow}$ & $\boldsymbol{Dist3}$\\
\hline\hline
\multirow{1}{*}{\textit{Original}} & \textit{Lens} & 96.89 & 59.31 & 77.23 & 70.77 & 76.05 & - & 51.09 & 0.555\\
\hline
\multirow{1}{*}{\textit{HoldOut-I.D.}} & \textit{Lens} & 94.53 & 60.30 & 78.33 & 71.19 & 76.09 & \multirow{2}{*}{11.87} & 52.63 & 0.562\\
\multirow{1}{*}{\textit{HoldOut-Comp.}} & \textit{Lens} & 77.03 & 56.05 & 78.23 & 56.93 & 67.06 &  & 52.59 & 0.556\\
\hline
\multirow{1}{*}{\textit{ACD-I.D.}} & \textit{Lens} & 94.15 & 62.34 & 76.83 & 76.22 & 77.39 & \multirow{2}{*}{25.95} & 54.63 & 0.526\\
\multirow{1}{*}{\textit{ACD-Comp.}} & \textit{Lens} & 60.80 & 57.27 & 51.68 & 59.49 & 57.31 &  & 54.15 & 0.469\\
\hline
\multirow{1}{*}{\textit{FewShot-I.D.}} & \textit{Lens} & 97.00 & 70.00 & 74.29 & 84.80 & 81.52 & \multirow{2}{*}{36.73} & 50.69 & 0.539\\
\multirow{1}{*}{\textit{FewShot-Comp.}} & \textit{Lens} & 63.60 & 50.63 & 34.18 & 57.92 & 51.58 &  & 50.25 & 0.501\\
\hlineB{2.5}
\end{tabularx}
\caption{\label{main-results-Fyelpv3-lens}
The result of baseline \textit{Lens}~\cite{gu-etal-2022-distributional} in dataset \textit{Fyelp}.
}
\end{table*}

\begin{table*}
\centering
\begin{tabularx}{\linewidth}{|l*{7}{X}|}
\hlineB{2.5}
\textbf{Protocol} & \textbf{Method} & $\boldsymbol{Acc_s}$ & $\boldsymbol{Acc_t}$ & $\boldsymbol{Acc_{avg}}$ & $\boldsymbol{Acc_{gap}}$ & $\boldsymbol{PPL \downarrow}$ & $\boldsymbol{Dist3}$\\
\hline\hline
\textit{Original} & \textit{Lens} & 91.67 & 81.52 & 86.60 & - & 68.33 & 0.666\\
\hline
\textit{HoldOut-I.D.} & \textit{Lens} & 91.68 & 83.31 & 87.50 & \multirow{2}{*}{47.78} & 69.95 & 0.660\\
\textit{HoldOut-Comp.} & \textit{Lens} & 48.26 & 43.12 & 45.69 &  & 130.07 & 0.663\\
\hline
\textit{FewShot-I.D.} & \textit{Lens} & 90.86 & 81.40 & 86.13 & \multirow{2}{*}{49.92} & 71.27 & 0.650\\
\textit{FewShot-Comp.} & \textit{Lens} & 48.85 & 37.40 & 43.13 &  & 198.37 & 0.587\\
\hlineB{2.5}
\end{tabularx}
\caption{\label{main-results-Amazon-lens}
The result of baseline \textit{Lens}~\cite{gu-etal-2022-distributional} in dataset \textit{Amazon}.
}
\end{table*}

\begin{table*}
\centering
\begin{tabularx}{\linewidth}{|l*{8}{X}|}
\hlineB{2.5}
\textbf{Protocol} & \textbf{Method} & $\boldsymbol{Acc_s}$ & $\boldsymbol{Acc_p}$ & $\boldsymbol{Acc_t}$ & $\boldsymbol{Acc_{avg}}$ & $\boldsymbol{Acc_{gap}}$ & $\boldsymbol{PPL \downarrow}$ & $\boldsymbol{Dist3}$\\
\hline\hline

\multirow{1}{*}{\textit{Original}} & \textit{Lens} & 79.54 & 96.75 & 93.36 & 89.88 & - & 265.42 & 0.284\\
\hline
\multirow{1}{*}{\textit{HoldOut-I.D.}} & \textit{Lens} & 71.74 & 96.77 & 95.47 & 87.99 & \multirow{2}{*}{36.73} & 121.94 & 0.232\\
\multirow{1}{*}{\textit{HoldOut-Comp.}} & \textit{Lens} & 51.54 & 64.75 & 50.71 & 55.67 & & 122.77 & 0.231\\
\hline
\multirow{1}{*}{\textit{ACD-I.D.}} & \textit{Lens} & 83.83 & 90.26 & 96.14 & 90.08 & \multirow{2}{*}{47.59} & 121.54 & 0.228\\
\multirow{1}{*}{\textit{ACD-Comp.}} & \textit{Lens} & 48.78 & 52.94 & 39.92 & 47.21 &  & 121.13 & 0.233\\
\hline
\multirow{1}{*}{\textit{FewShot-I.D.}} & \textit{Lens} & 98.54 & 89.25 & 97.25 & 95.01 & \multirow{2}{*}{36.07} & 142.18 & 0.212\\
\multirow{1}{*}{\textit{FewShot-Comp.}} & \textit{Lens} & 62.87 & 58.14 & 61.20 & 60.74 & & 141.35 & 0.271\\
\hlineB{2.5}
\end{tabularx}
\caption{\label{main-results-Yelp-lens}
The result of baseline \textit{Lens}~\cite{gu-etal-2022-distributional} in dataset \textit{YELP}.
}
\end{table*}

\begin{table*}
\centering
\begin{tabularx}{\linewidth}{|l*{7}{X}|}
\hlineB{2.5}
\textbf{Protocol} & \textbf{Method} & $\boldsymbol{Acc_s}$ & $\boldsymbol{Acc_{tc}}$ & $\boldsymbol{Acc_{avg}}$ & $\boldsymbol{Acc_{gap}}$ & $\boldsymbol{PPL \downarrow}$ & $\boldsymbol{Dist3}$\\
\hline\hline
\textit{Original} & \textit{Lens} & 83.11 & 95.46 & 89.29 & - & 110.04 & 0.387\\
\hline
\textit{HoldOut-I.D.} & \textit{Lens} & 82.14 & 93.37 & 87.76 & \multirow{2}{*}{38.58} & 138.82 & 0.410\\
\textit{HoldOut-Comp.} & \textit{Lens} & 52.00 & 55.79 & 53.90 &  & 114.13 & 0.397\\
\hline
\textit{FewShot-I.D.} & \textit{Lens} & 81.41 & 95.72 & 88.57 & \multirow{2}{*}{43.05} & 116.04 & 0.410\\
\textit{FewShot-Comp.} & \textit{Lens} & 49.36 & 51.52 & 50.44 &  & 76.73 & 0.418\\
\hlineB{2.5}
\end{tabularx}
\caption{\label{main-results-Mixture-lens}
The result of baseline \textit{Lens}~\cite{gu-etal-2022-distributional} in dataset \textit{Mixture}.
}
\end{table*}

\begin{table*}
\centering
\begin{tabularx}{\linewidth}{|l*{9}{X}|}
\hlineB{2.5}
\textbf{Protocol} & \textbf{Method} & $\boldsymbol{Acc_s}$ & $\boldsymbol{Acc_g}$ & $\boldsymbol{Acc_c}$ & $\boldsymbol{
Acc_t}$ & $\boldsymbol{Acc_{avg}}$ & $\boldsymbol{Acc_{gap}}$ & $\boldsymbol{PPL \downarrow}$ & $\boldsymbol{Dist3}$\\
\hline\hline
\multirow{1}{*}{\textit{Original}} & \textit{Prior} & 72.43 & 52.02 & 48.39 & 63.58 & 59.11 & - & 72.14 & 0.602\\
\hline
\multirow{1}{*}{\textit{HoldOut-I.D.}} & \textit{Prior} & 70.82 & 51.96 & 46.51 & 64.13 & 58.36 & \multirow{2}{*}{6.37} & 73.95 & 0.607\\
\multirow{1}{*}{\textit{HoldOut-Comp.}} & \textit{Prior} & 63.56 & 50.79 & 43.58 & 60.62 & 54.64 &  & 73.91 & 0.609\\
\hline
\multirow{1}{*}{\textit{ACD-I.D.}} & \textit{Prior} & 72.96 & 54.53 & 47.62 & 71.36 & 61.62 & \multirow{2}{*}{15.14} & 79.37 & 0.624\\
\multirow{1}{*}{\textit{ACD.Comp.}} & \textit{Prior} & 68.42 & 48.29 & 48.26 & 44.20 & 52.29 &  & 79.10 & 0.627\\
\hline
\multirow{1}{*}{\textit{FewShot-I.D.}} & \textit{Prior} & 98.11 & 73.89 & 55.83 & 86.86 & 78.67 & \multirow{2}{*}{32.54} & 84.29 & 0.643\\
\multirow{1}{*}{\textit{FewShot-Comp.}} & \textit{Prior} & 59.07 & 47.37 & 48.67 & 57.18 & 53.07 &  & 83.13 & 0.576\\
\hlineB{2.5}
\end{tabularx}
\caption{\label{main-results-Fyelpv3-prior}
The result of baseline \textit{Prior}~\cite{gu-etal-2023-controllable} in dataset \textit{Fyelp}.
}
\end{table*}

\begin{table*}
\centering
\begin{tabularx}{\linewidth}{|l*{7}{X}|}
\hlineB{2.5}
\textbf{Protocol} & \textbf{Method} & $\boldsymbol{Acc_s}$ & $\boldsymbol{Acc_t}$ & $\boldsymbol{Acc_{avg}}$ & $\boldsymbol{Acc_{gap}}$ & $\boldsymbol{PPL \downarrow}$ & $\boldsymbol{Dist3}$\\
\hline\hline
\textit{Original} & \textit{Prior} & 82.02 & 82.90 & 82.46 & - & 86.79 & 0.647\\
\hline
\textit{HoldOut-I.D.} & \textit{Prior} & 83.78 & 79.46 & 81.62 & \multirow{2}{*}{40.74} & 86.93 & 0.644\\
\textit{HoldOut-Comp.} & \textit{Prior} & 25.76 & 70.98 & 48.37 &  & 84.02 & 0.650\\
\hline
\textit{FewShot-I.D.} & \textit{Prior} & 96.91 & 78.99 & 87.95 & \multirow{2}{*}{40.11} & 93.00 & 0.643\\
\textit{FewShot-Comp.} & \textit{Prior} & 54.43 & 50.90 & 52.67 &  & 93.80 & 0.648\\
\hlineB{2.5}
\end{tabularx}
\caption{\label{main-results-Amazon-prior}
The result of baseline \textit{Prior}~\cite{gu-etal-2023-controllable} in dataset \textit{Amazon}.
}
\end{table*}

\begin{table*}
\centering
\begin{tabularx}{\linewidth}{|l*{8}{X}|}
\hlineB{2.5}
\textbf{Protocol} & \textbf{Method} & $\boldsymbol{Acc_s}$ & $\boldsymbol{Acc_p}$ & $\boldsymbol{Acc_t}$ & $\boldsymbol{Acc_{avg}}$ & $\boldsymbol{Acc_{gap}}$ & $\boldsymbol{PPL \downarrow}$ & $\boldsymbol{Dist3}$\\
\hline\hline
\multirow{1}{*}{\textit{Original}} & \textit{Prior} & 70.96 & 65.11 & 82.93 & 73.00 & - & 124.68 & 0.477\\
\hline
\multirow{1}{*}{\textit{HoldOut-I.D.}} & \textit{Prior} & 73.48 & 63.91 & 80.62 & 72.67 & \multirow{2}{*}{24.99} & 68.44 & 0.379\\
\multirow{1}{*}{\textit{HoldOut-Comp.}} & \textit{Prior} & 55.89 & 51.18 & 56.46 & 54.51 & & 65.61 & 0.398\\
\hline
\multirow{1}{*}{\textit{ACD-I.D.}} & \textit{Prior} & 79.93 & 68.35 & 82.45 & 76.91 & \multirow{2}{*}{39.11} & 82.68 & 0.347\\
\multirow{1}{*}{\textit{ACD-Comp.}} & \textit{Prior} & 48.45 & 51.56 & 40.48 & 46.83 &  & 72.61 & 0.344\\
\hline
\multirow{1}{*}{\textit{FewShot-I.D.}} & \textit{Prior} & 89.68 & 77.07 & 96.21 & 87.65 & \multirow{2}{*}{39.92} & 98.73 & 0.287\\
\multirow{1}{*}{\textit{FewShot-Comp.}} & \textit{Prior} & 53.36 & 51.62 & 53.00 & 52.66 & & 94.69 & 0.345\\
\hlineB{2.5}
\end{tabularx}
\caption{\label{main-results-Yelp-prior}
The result of baseline \textit{Prior}~\cite{gu-etal-2023-controllable} in dataset \textit{YELP}.
}
\end{table*}

\begin{table*}
\centering
\begin{tabularx}{\linewidth}{|l*{7}{X}|}
\hlineB{2.5}
\textbf{Protocol} & \textbf{Method} & $\boldsymbol{Acc_s}$ & $\boldsymbol{Acc_{tc}}$ & $\boldsymbol{Acc_{avg}}$ & $\boldsymbol{Acc_{gap}}$ & $\boldsymbol{PPL \downarrow}$ & $\boldsymbol{Dist3}$\\
\hline\hline
\textit{Original} & \textit{Prior} & 77.79 & 83.89 & 80.84 & - & 196.01 & 0.565\\
\hline
\textit{HoldOut-I.D.} & \textit{Prior} & 81.69 & 82.08 & 81.89 & \multirow{2}{*}{48.49} & 205.01 & 0.558\\
\textit{HoldOut-Comp.} & \textit{Prior} & 41.07 & 43.29 & 42.18 &  & 167.01 & 0.535\\
\hline
\textit{FewShot-I.D.} & \textit{Prior} & 85.56 & 87.42 & 86.49 & \multirow{2}{*}{44.02} & 199.85 & 0.541\\
\textit{FewShot-Comp.} & \textit{Prior} & 49.40 & 47.43 & 48.42 &  & 145.01 & 0.540\\
\hlineB{2.5}
\end{tabularx}
\caption{\label{main-results-Mixture-prior}
The result of baseline \textit{Prior}~\cite{gu-etal-2023-controllable} in dataset \textit{Mixture}.
}
\end{table*}

\begin{table*}
\centering
\begin{tabularx}{\linewidth}{|l*{9}{X}|}
\hlineB{2.5}
\textbf{Protocol} & \textbf{Method} & $\boldsymbol{Acc_s}$ & $\boldsymbol{Acc_g}$ & $\boldsymbol{Acc_c}$ & $\boldsymbol{
Acc_t}$ & $\boldsymbol{Acc_{avg}}$ & $\boldsymbol{Acc_{gap}}$ & $\boldsymbol{PPL \downarrow}$ & $\boldsymbol{Dist3}$\\
\hline\hline
\multirow{1}{*}{\textit{Original}} & \textit{Con.P} & 93.47 & 59.39 & 50.41 & 69.11 & 68.10 & - & 51.76 & 0.704\\
\hline
\multirow{1}{*}{\textit{HoldOut-I.D.}} & \textit{Con.P} & 93.67 & 59.25 & 49.64 & 68.79 & 67.84 & \multirow{2}{*}{0.50} & 52.48 & 0.701\\
\multirow{1}{*}{\textit{HoldOut-Comp.}} & \textit{Con.P} & 93.66 & 59.24 & 48.30 & 68.78 & 67.50 &  & 52.32 & 0.705\\
\hline
\multirow{1}{*}{\textit{ACD-I.D.}} & \textit{Con.P} & 92.50 & 57.39 & 39.04 & 64.68 & 63.40 & \multirow{2}{*}{-0.84} & 53.11 & 0.704\\
\multirow{1}{*}{\textit{ACD-Comp.}} & \textit{Con.P} & 93.85 & 58.24 & 40.18 & 63.44 & 63.93 &  & 49.78 & 0.745\\
\hline
\multirow{1}{*}{\textit{FewShot-I.D.}} & \textit{Con.P} & 81.69 & 72.09 & 24.49 & 60.40 & 59.67 & \multirow{2}{*}{24.03} & 76.80 & 0.744\\
\multirow{1}{*}{\textit{FewShot-Comp.}} & \textit{Con.P} & 58.89 & 47.51 & 22.39 & 52.51 & 45.33 &  & 86.49 & 0.745\\
\hlineB{2.5}
\end{tabularx}
\caption{\label{main-results-Fyelpv3-contrastive_prefix}
The result of baseline \textit{Contrastive Prefix}~\cite{qian-etal-2022-controllable} in dataset \textit{Fyelp}.
}
\end{table*}

\begin{table*}
\centering
\begin{tabularx}{\linewidth}{|l*{7}{X}|}
\hlineB{2.5}
\textbf{Protocol} & \textbf{Method} & $\boldsymbol{Acc_s}$ & $\boldsymbol{Acc_t}$ & $\boldsymbol{Acc_{avg}}$ & $\boldsymbol{Acc_{gap}}$ & $\boldsymbol{PPL \downarrow}$ & $\boldsymbol{Dist3}$\\
\hline\hline
\textit{Original} & \textit{Con.P} & 93.76 & 81.31 & 87.54 & - & 43.55 & 0.716\\
\hline
\textit{HoldOut-I.D.} & \textit{Con.P} & 94.26 & 81.27 & 87.77 & \multirow{2}{*}{-0.50} & 43.84 & 0.713\\
\textit{HoldOut-Comp.} & \textit{Con.P} & 94.67 & 81.74 & 88.21 &  & 44.49 & 0.716\\
\hline
\textit{FewShot-I.D.} & \textit{Con.P} & 92.93 & 77.13 & 85.03 & \multirow{2}{*}{19.45} & 43.92 & 0.713\\
\textit{FewShot-Comp.} & \textit{Con.P} & 82.72 & 54.26 & 68.49 &  & 43.28 & 0.727\\
\hlineB{2.5}
\end{tabularx}
\caption{\label{main-results-Amazon-con.prefix}
The result of baseline \textit{Contrastive Prefix}~\cite{qian-etal-2022-controllable} in dataset \textit{Amazon}.
}
\end{table*}

\begin{table*}
\centering
\begin{tabularx}{\linewidth}{|l*{8}{X}|}
\hlineB{2.5}
\textbf{Protocol} & \textbf{Method} & $\boldsymbol{Acc_s}$ & $\boldsymbol{Acc_p}$ & $\boldsymbol{Acc_t}$ & $\boldsymbol{Acc_{avg}}$ & $\boldsymbol{Acc_{gap}}$ & $\boldsymbol{PPL \downarrow}$ & $\boldsymbol{Dist3}$\\
\hline\hline
\multirow{1}{*}{\textit{Original}} & \textit{Con.P} & 98.21 & 87.11 & 99.21 & 94.84 & - & 139.13 & 0.709\\
\hline
\multirow{1}{*}{\textit{HoldOut-I.D.}} & \textit{Con.P} & 98.03 & 85.91 & 99.26 & 94.40 & \multirow{2}{*}{1.71} & 136.04 & 0.687\\
\multirow{1}{*}{\textit{HoldOut-Comp.}} & \textit{Con.P} & 97.36 & 82.11 & 98.89 & 92.79 & & 132.21 & 0.707\\
\hline
\multirow{1}{*}{\textit{ACD-I.D.}} & \textit{Con.P} & 96.52 & 80.96 & 98.66 & 92.05 & \multirow{2}{*}{3.34} & 139.71 & 0.669\\
\multirow{1}{*}{\textit{ACD-Comp.}} & \textit{Con.P} & 96.27 & 72.73 & 97.93 & 88.98 &  & 131.12 & 0.674\\
\hline
\multirow{1}{*}{\textit{FewShot-I.D.}} & \textit{Con.P} & 96.09 & 78.25 & 97.82 & 90.72 & \multirow{2}{*}{35.53} & 136.95 & 0.527\\
\multirow{1}{*}{\textit{FewShot-Comp.}} & \textit{Con.P} & 60.87 & 52.94 & 61.65 & 58.49 & & 132.02 & 0.624\\
\hlineB{2.5}
\end{tabularx}
\caption{\label{main-results-Yelp-con.prefix}
The result of baseline \textit{Contrastive Prefix}~\cite{qian-etal-2022-controllable} in dataset \textit{YELP}.
}
\end{table*}

\begin{table*}
\centering
\begin{tabularx}{\linewidth}{|l*{7}{X}|}
\hlineB{2.5}
\textbf{Protocol} & \textbf{Method} & $\boldsymbol{Acc_s}$ & $\boldsymbol{Acc_{tc}}$ & $\boldsymbol{Acc_{avg}}$ & $\boldsymbol{Acc_{gap}}$ & $\boldsymbol{PPL \downarrow}$ & $\boldsymbol{Dist3}$\\
\hline\hline
\textit{Original} & \textit{Con.P} & 75.68 & 95.25 & 85.47 & - & 82.73 & 0.676\\
\hline
\textit{HoldOut-I.D.} & \textit{Con.P} & 75.87 & 94.08 & 84.98 & \multirow{2}{*}{14.16} & 89.59 & 0.681\\
\textit{HoldOut-Comp.} & \textit{Con.P} & 66.82 & 79.07 & 72.95 &  & 119.74 & 0.778\\
\hline
\textit{FewShot-I.D.} & \textit{Con.P} & 74.12 & 94.11 & 84.12 & \multirow{2}{*}{31.12} & 86.10 & 0.642\\
\textit{FewShot-Comp.} & \textit{Con.P} & 52.47 & 63.40 & 57.94 &  & 111.43 & 0.723\\
\hlineB{2.5}
\end{tabularx}
\caption{\label{main-results-Mixture-con.prefix}
The result of baseline \textit{Contrastive Prefix}~\cite{qian-etal-2022-controllable} in dataset \textit{Mixture}.
}
\end{table*}

\begin{table*}
\centering
\begin{tabularx}{\linewidth}{|l*{9}{X}|}
\hlineB{2.5}
\textbf{Protocol} & \textbf{Method} & $\boldsymbol{Acc_s}$ & $\boldsymbol{Acc_g}$ & $\boldsymbol{Acc_c}$ & $\boldsymbol{
Acc_t}$ & $\boldsymbol{Acc_{avg}}$ & $\boldsymbol{Acc_{gap}}$ & $\boldsymbol{PPL \downarrow}$ & $\boldsymbol{Dist3}$\\
\hline\hline
\multirow{1}{*}{\textit{Original}} & \textit{PPLM} & 49.86 & 50.00 & 19.91 & 49.91 & 42.42 & - & 355.27 & 0.691\\
\hline
\multirow{1}{*}{\textit{HoldOut-I.D.}} & \textit{PPLM} & 50.43 & 50.03 & 20.34 & 50.31 & 42.78 & \multirow{2}{*}{0.68} & 351.74 & 0.687\\
\multirow{1}{*}{\textit{HoldOut-Comp.}} & \textit{PPLM} & 49.96 & 50.02 & 19.93 & 50.06 & 42.49 &  & 365.57 & 0.688\\
\hline
\multirow{1}{*}{\textit{ACD-I.D.}} & \textit{PPLM} & 49.30 & 52.75 & 20.62 & 54.55 & 44.31 & \multirow{2}{*}{8.31} & 348.59 & 0.688\\
\multirow{1}{*}{\textit{ACD-Comp.}} & \textit{PPLM} & 50.57 & 47.25 & 19.42 & 45.27 & 40.63 &  & 329.13 & 0.688\\
\hline
\multirow{1}{*}{\textit{FewShot-I.D.}} & \textit{PPLM} & 55.11 & 79.57 & 19.06 & 42.14 & 48.97 & \multirow{2}{*}{15.15} & 470.44 & 0.692\\
\multirow{1}{*}{\textit{FewShot-Comp.}} & \textit{PPLM} & 49.42 & 45.79 & 20.09 & 50.90 & 41.55 &  & 332.87 & 0.686\\
\hlineB{2.5}
\end{tabularx}
\caption{\label{main-results-Fyelpv3-pplm}
The result of baseline \textit{PPLM}~\cite{dathathri2019plug} in dataset \textit{Fyelp.}
}
\end{table*}

\begin{table*}
\centering
\begin{tabularx}{\linewidth}{|l*{7}{X}|}
\hlineB{2.5}
\textbf{Protocol} & \textbf{Method} & $\boldsymbol{Acc_s}$ & $\boldsymbol{Acc_t}$ & $\boldsymbol{Acc_{avg}}$ & $\boldsymbol{Acc_{gap}}$ & $\boldsymbol{PPL \downarrow}$ & $\boldsymbol{Dist3}$\\
\hline\hline
\textit{Original} & \textit{PPLM} & 49.60 & 16.62 & 33.11 & - & 340.99 & 0.689\\
\hline
\textit{HoldOut-I.D.} & \textit{PPLM} & 50.31 & 17.24 & 33.78 & \multirow{2}{*}{1.51} & 379.86 & 0.689\\
\textit{HoldOut-Comp.} & \textit{PPLM} & 49.64 & 16.89 & 33.27 &  & 346.97 & 0.691\\
\hline
\textit{FewShot-I.D.} & \textit{PPLM} & 53.04 & 16.75 & 34.90 & \multirow{2}{*}{8.51} & 343.87 & 0.690\\
\textit{FewShot-Comp.} & \textit{PPLM} & 47.01 & 16.85 & 31.93 &  & 355.93 & 0.686 \\
\hlineB{2.5}
\end{tabularx}
\caption{\label{main-results-Amazon-pplm}
The result of baseline \textit{PPLM}~\cite{dathathri2019plug} in dataset \textit{Amazon}.
}
\end{table*}

\begin{table*}
\centering
\begin{tabularx}{\linewidth}{|l*{8}{X}|}
\hlineB{2.5}
\textbf{Protocol} & \textbf{Method} & $\boldsymbol{Acc_s}$ & $\boldsymbol{Acc_p}$ & $\boldsymbol{Acc_t}$ & $\boldsymbol{Acc_{avg}}$ & $\boldsymbol{Acc_{gap}}$ & $\boldsymbol{PPL \downarrow}$ & $\boldsymbol{Dist3}$\\
\hline\hline
\multirow{1}{*}{\textit{Original}} & \textit{PPLM} & 50.43 & 49.86 & 49.75 & 50.01 & - & 297.53 & 0.704\\
\hline
\multirow{1}{*}{\textit{HoldOut-I.D.}} & \textit{PPLM} & 50.46 & 49.43 & 48.79 & 49.56 & \multirow{2}{*}{0.93} & 294.58 & 0.422\\
\multirow{1}{*}{\textit{HoldOut-Comp.}} & \textit{PPLM} & 50.32 & 48.28 & 48.70 & 49.10 & & 294.58 & 0.695\\
\hline
\multirow{1}{*}{\textit{ACD-I.D.}} & \textit{PPLM} & 54.46 & 50.04 & 50.42 & 51.64 & \multirow{2}{*}{5.58} & 289.95 & 0.439\\
\multirow{1}{*}{\textit{ACD-Comp.}} & \textit{PPLM} & 45.54 & 50.10 & 50.65 & 48.76 &  & 285.21 & 0.434\\
\hline
\multirow{1}{*}{\textit{FewShot-I.D.}} & \textit{PPLM} & 49.86 & 49.71 & 51.25 & 50.27 & \multirow{2}{*}{0} & 302.25 & 0.492\\
\multirow{1}{*}{\textit{FewShot-Comp.}} & \textit{PPLM} & 49.86 & 49.71 & 51.25 & 50.27 & & 302.26 & 0.438\\
\hlineB{2.5}
\end{tabularx}
\caption{\label{main-results-Yelp-pplm}
The result of baseline \textit{PPLM}~\cite{dathathri2019plug} in dataset \textit{YELP}.
}
\end{table*}

\begin{table*}
\centering
\begin{tabularx}{\linewidth}{|l*{7}{X}|}
\hlineB{2.5}
\textbf{Protocol} & \textbf{Method} & $\boldsymbol{Acc_s}$ & $\boldsymbol{Acc_{tc}}$ & $\boldsymbol{Acc_{avg}}$ & $\boldsymbol{Acc_{gap}}$ & $\boldsymbol{PPL \downarrow}$ & $\boldsymbol{Dist3}$\\
\hline\hline
\textit{Original} & \textit{PPLM} & 51.71 & 24.50 & 38.11 & - & 296.57 & 0.704\\
\hline
\textit{HoldOut-I.D.} & \textit{PPLM} & 51.18 & 24.93 & 38.06 & \multirow{2}{*}{1.21} & 274.16 & 0.690\\
\textit{HoldOut-Comp.} & \textit{PPLM} & 50.14 & 25.05 & 37.60 &  & 355.92 & 0.702\\
\hline
\textit{FewShot-I.D.} & \textit{PPLM} & 50.94 & 25.35 & 38.15 & \multirow{2}{*}{2.80} & 329.85 & 0.665\\
\textit{FewShot-Comp.} & \textit{PPLM} & 48.93 & 25.22 & 37.08 &  & 332.68 & 0.660\\
\hlineB{2.5}
\end{tabularx}
\caption{\label{main-results-Mixture-pplm}
The result of baseline \textit{PPLM}~\cite{dathathri2019plug} in dataset \textit{Mixture}.
}
\end{table*}

\begin{table*}
\centering
\begin{tabularx}{\linewidth}{|l*{9}{X}|}
\hlineB{2.5}
\textbf{Protocol} & \textbf{Method} & $\boldsymbol{Acc_s}$ & $\boldsymbol{Acc_g}$ & $\boldsymbol{Acc_c}$ & $\boldsymbol{
Acc_t}$ & $\boldsymbol{Acc_{avg}}$ & $\boldsymbol{Acc_{gap}}$ & $\boldsymbol{PPL \downarrow}$ & $\boldsymbol{Dist3}$\\
\hline\hline
\multirow{1}{*}{\textit{Original}} & \textit{llama2} & 66.57 & 52.00 & 32.50 & 56.07 & 51.78 & - & 17.64 & 0.473\\
\hline
\multirow{1}{*}{\textit{HoldOut-I.D.}} & \textit{llama2} & 66.94 & 52.72 & 30.81 & 55.99 & 51.61 & \multirow{2}{*}{15.09} & 17.08 & 0.387\\
\multirow{1}{*}{\textit{HoldOut-Comp.}} & \textit{llama2} & 56.43 & 49.79 & 20.36 & 48.71 & 43.82 &  & 16.56 & 0.449\\
\hline
\multirow{1}{*}{\textit{ACD-I.D.}} & \textit{llama2} & 68.36 & 51.51 & 29.50 & 56.94 & 51.58 & \multirow{2}{*}{15.99} & 16.72 & 0.379\\
\multirow{1}{*}{\textit{ACD-Comp.}} & \textit{llama2} & 55.31 & 49.37 & 20.67 & 47.96 & 43.33 &  & 17.34 & 0.371\\
\hline
\multirow{1}{*}{\textit{FewShot-I.D.}} & \textit{llama2} & 65.37 & 52.17 & 29.77 & 56.11 & 50.86 & \multirow{2}{*}{12.09} & 17.21 & 0.444\\
\multirow{1}{*}{\textit{FewShot-Comp.}} & \textit{llama2} & 57.59 & 49.17 & 21.07 & 50.99 & 44.71 &  & 17.46 & 0.374\\
\hlineB{2.5}
\end{tabularx}
\caption{\label{main-results-Fyelpv3-llama2-anony}
The result of baseline \textit{LLaMA-2}~\cite{llama2} in dataset \textit{Fyelp}.
}
\end{table*}

\begin{table*}
\centering
\begin{tabularx}{\linewidth}{|l*{7}{X}|}
\hlineB{2.5}
\textbf{Protocol} & \textbf{Method} & $\boldsymbol{Acc_s}$ & $\boldsymbol{Acc_t}$ & $\boldsymbol{Acc_{avg}}$ & $\boldsymbol{Acc_{gap}}$ & $\boldsymbol{PPL \downarrow}$ & $\boldsymbol{Dist3}$\\
\hline\hline
\textit{Original} & \textit{llama2} & 68.10 & 53.10 & 60.60 & - & 15.25 & 0.633\\
\hline
\textit{HoldOut-I.D.} & \textit{llama2} & 72.03 & 51.13 & 61.58 & \multirow{2}{*}{47.22} & 15.16 & 0.442\\
\textit{HoldOut-Comp.} & \textit{llama2} & 47.86 & 17.14 & 32.50 &  & 15.50 & 0.622\\
\hline
\textit{FewShot-I.D.} & \textit{llama2} & 75.81 & 51.10 & 63.45 & \multirow{2}{*}{49.24} & 15.14 & 0.474\\
\textit{FewShot-Comp.} & \textit{llama2} & 47.86 & 16.57 & 32.21 &  & 15.23 & 0.474 \\
\hlineB{2.5}
\end{tabularx}
\caption{\label{main-results-Amazon-llama2-anony}
The result of baseline \textit{LLaMA-2}~\cite{llama2} in dataset \textit{Amazon}.
}
\end{table*}

\begin{table*}
\centering
\begin{tabularx}{\linewidth}{|l*{8}{X}|}
\hlineB{2.5}
\textbf{Protocol} & \textbf{Method} & $\boldsymbol{Acc_s}$ & $\boldsymbol{Acc_p}$ & $\boldsymbol{Acc_t}$ & $\boldsymbol{Acc_{avg}}$ & $\boldsymbol{Acc_{gap}}$ & $\boldsymbol{PPL \downarrow}$ & $\boldsymbol{Dist3}$\\
\hline\hline
\multirow{1}{*}{\textit{Original}} & \textit{llama2} & 74.29 & 51.43 & 70.36 & 65.36 & - & 48.79 & 0.575\\
\hline
\multirow{1}{*}{\textit{HoldOut-I.D.}} & \textit{llama2} & 70.92 & 53.06 & 72.81 & 65.60 & \multirow{2}{*}{27.59} & 46.45 & 0.391\\
\multirow{1}{*}{\textit{HoldOut-Comp.}} & \textit{llama2} & 49.64 & 50.00 & 42.86 & 47.50 & & 47.49 & 0.551\\
\hline
\multirow{1}{*}{\textit{ACD-I.D.}} & \textit{llama2} & 68.93 & 54.64 & 72.29 & 65.29 & \multirow{2}{*}{22.81} & 54.56 & 0.410\\
\multirow{1}{*}{\textit{ACD-Comp.}} & \textit{llama2} & 50.86 & 49.71 & 50.64 & 50.40 &  & 49.36 & 0.399\\
\hline
\multirow{1}{*}{\textit{FewShot-I.D.}} & \textit{llama2} & 72.68 & 52.50 & 70.36 & 65.18 & \multirow{2}{*}{19.42} & 45.17 & 0.486\\
\multirow{1}{*}{\textit{FewShot-Comp.}} & \textit{llama2} & 56.61 & 50.06 & 50.89 & 52.52 & & 46.32 & 0.384\\
\hlineB{2.5}
\end{tabularx}
\caption{\label{main-results-Yelp-llama2-anony}
The result of baseline \textit{LLaMA-2}~\cite{llama2} in dataset \textit{YELP}.
}
\end{table*}

\begin{table*}
\centering
\begin{tabularx}{\linewidth}{|l*{7}{X}|}
\hlineB{2.5}
\textbf{Protocol} & \textbf{Method} & $\boldsymbol{Acc_s}$ & $\boldsymbol{Acc_{tc}}$ & $\boldsymbol{Acc_{avg}}$ & $\boldsymbol{Acc_{gap}}$ & $\boldsymbol{PPL \downarrow}$ & $\boldsymbol{Dist3}$\\
\hline\hline
\textit{Original} & \textit{llama2} & 52.14 & 84.64 & 68.39 & - & 27.53 & 0.667\\
\hline
\textit{HoldOut-I.D.} & \textit{llama2} & 58.78 & 84.54 & 71.66 & \multirow{2}{*}{44.92} & 23.49 & 0.500\\
\textit{HoldOut-Comp.} & \textit{llama2} & 51.07 & 27.86 & 39.47 &  & 15.65 & 0.686\\
\hline
\textit{FewShot-I.D.} & \textit{llama2} & 56.52 & 86.70 & 71.61 & \multirow{2}{*}{40.65} & 26.81 & 0.559\\
\textit{FewShot-Comp.} & \textit{llama2} & 56.79 & 28.21 & 42.50 &  & 16.57 & 0.558\\
\hlineB{2.5}
\end{tabularx}
\caption{\label{main-results-Mixture-llama2-anony}
The result of baseline \textit{LLaMA-2}~\cite{llama2} in dataset \textit{Mixture}.
}
\end{table*}

\begin{table*}
\centering
\begin{tabularx}{\linewidth}{|l*{9}{X}|}
\hlineB{2.5}
\textbf{Protocol} & \textbf{Method} & $\boldsymbol{Acc_s}$ & $\boldsymbol{Acc_g}$ & $\boldsymbol{Acc_c}$ & $\boldsymbol{
Acc_t}$ & $\boldsymbol{Acc_{avg}}$ & $\boldsymbol{Acc_{gap}}$ & $\boldsymbol{PPL \downarrow}$ & $Dist3$\\
\hline\hline
\multirow{1}{*}{\textit{Original}} & \textit{gpt3.5} & 66.29 & 52.29 & 28.14 & 57.00 & 50.93 & - & 13.41 & 0.454\\
\hline
\multirow{1}{*}{\textit{HoldOut-I.D.}} & \textit{gpt3.5} & 67.07 & 51.10 & 27.90 & 56.29 & 50.59 & \multirow{2}{*}{7.61} & 13.39 & 0.347\\
\multirow{1}{*}{\textit{HoldOut-Comp.}} & \textit{gpt3.5} & 59.05 & 52.06 & 31.11 & 44.76 & 46.74 &  & 12.50 & 0.652\\
\hline
\multirow{1}{*}{\textit{ACD-I.D.}} & \textit{gpt3.5} & 64.25 & 50.68 & 29.34 & 56.43 & 50.17 & \multirow{2}{*}{5.74} & 13.52 & 0.347\\
\multirow{1}{*}{\textit{ACD-Comp.}} & \textit{gpt3.5} & 60.12 & 49.45 & 27.77 & 51.80 & 47.29 &  & 13.29 & 0.369\\
\hline
\multirow{1}{*}{\textit{FewShot-I.D.}} & \textit{gpt3.5} & 49.14 & 58.00 & 26.00 & 62.29 & 48.86 & \multirow{2}{*}{2.89} & 13.06 & 0.627\\
\multirow{1}{*}{\textit{FewShot-Comp.}} & \textit{gpt3.5} & 68.65 & 48.08 & 25.35 & 47.71 & 47.45 &  & 13.07 & 0.401\\
\hlineB{2.5}
\end{tabularx}
\caption{\label{main-results-Fyelpv3-gpt-3.5-anony}
The result of baseline \textit{ChatGPT} (gpt-3.5-turbo-0613)~\cite{openaiChatGPT} in dataset \textit{Fyelp}.
}
\end{table*}

\begin{table*}
\centering
\begin{tabularx}{\linewidth}{|l*{7}{X}|}
\hlineB{2.5}
\textbf{Protocol} & \textbf{Method} & $\boldsymbol{Acc_s}$ & $\boldsymbol{Acc_t}$ & $\boldsymbol{Acc_{avg}}$ & $\boldsymbol{Acc_{gap}}$ & $\boldsymbol{PPL \downarrow}$ & $\boldsymbol{Dist3}$\\
\hline\hline
\textit{Original} & \textit{gpt3.5} & 77.86 & 33.33 & 55.59 & - & 14.13 & 0.670\\
\hline
\textit{HoldOut-I.D.} & \textit{gpt3.5} & 74.72 & 36.54 & 55.63 & \multirow{2}{*}{15.69} & 14.50 & 0.417\\
\textit{HoldOut-Comp.} & \textit{gpt3.5} & 75.71 & 18.10 & 46.90 &  & 14.94 & 0.667\\
\hline
\textit{FewShot-I.D.} & \textit{gpt3.5} & 79.29 & 36.43 & 57.86 & \multirow{2}{*}{20.26} & 14.50 & 0.472\\
\textit{FewShot-Comp.} & \textit{gpt3.5} & 71.52 & 20.76 & 46.14 &  & 14.24 &  0.474\\
\hlineB{2.5}
\end{tabularx}
\caption{\label{main-results-Amazon-gpt-3.5-anony}
The result of baseline \textit{ChatGPT} (gpt-3.5-turbo-0613)~\cite{openaiChatGPT} in dataset \textit{Amazon}.
}
\end{table*}

\begin{table*}
\centering
\begin{tabularx}{\linewidth}{|l*{8}{X}|}
\hlineB{2.5}
\textbf{Protocol} & \textbf{Method} & $\boldsymbol{Acc_s}$ & $\boldsymbol{Acc_p}$ & $\boldsymbol{Acc_t}$ & $\boldsymbol{Acc_{avg}}$ & $\boldsymbol{Acc_{gap}}$ & $\boldsymbol{PPL \downarrow}$ & $\boldsymbol{Dist3}$\\
\hline\hline
\multirow{1}{*}{\textit{Original}} & \textit{gpt3.5} & 53.57 & 51.43 & 66.79 & 57.26 & - & 25.58 & 0.596\\
\hline
\multirow{1}{*}{\textit{HoldOut-I.D.}} & \textit{gpt3.5} & 60.97 & 50.41 & 65.77 & 59.05 & \multirow{2}{*}{6.86} & 26.43 & 0.367\\
\multirow{1}{*}{\textit{HoldOut-Comp.}} & \textit{gpt3.5} & 67.14 & 50.36 & 47.50 & 55.00 & & 26.41 & 0.614\\
\hline
\multirow{1}{*}{\textit{ACD-I.D.}} & \textit{gpt3.5}& 60.86 & 51.43 & 67.71 & 60.00 & \multirow{2}{*}{4.88} & 25.76 & 0.400\\
\multirow{1}{*}{\textit{ACD-Comp.}} & \textit{gpt3.5} & 71.07 & 50.71 & 49.43 & 57.07 &  & 28.81 & 0.421\\
\hline
\multirow{1}{*}{\textit{FewShot-I.D.}} & \textit{gpt3.5} & 58.75 & 51.07 & 68.21 & 59.34 & \multirow{2}{*}{5.73} & 27.61 & 0.498\\
\multirow{1}{*}{\textit{FewShot-Comp.}} & \textit{gpt3.5} & 65.42 & 50.54 & 51.85 & 55.94 & & 26.98 & 0.384\\
\hlineB{2.5}
\end{tabularx}
\caption{\label{main-results-Yelp-gpt-3.5-anony}
The result of baseline \textit{ChatGPT} (gpt-3.5-turbo-0613)~\cite{openaiChatGPT} in dataset \textit{YELP}.
}
\end{table*}

\begin{table*}
\centering
\begin{tabularx}{\linewidth}{|l*{7}{X}|}
\hlineB{2.5}
\textbf{Protocol} & \textbf{Method} & $\boldsymbol{Acc_s}$ & $\boldsymbol{Acc_{tc}}$ & $\boldsymbol{Acc_{avg}}$ & $\boldsymbol{Acc_{gap}}$ & $\boldsymbol{PPL \downarrow}$ & $\boldsymbol{Dist3}$\\
\hline\hline
\textit{Original} & \textit{gpt3.5} & 69.64 & 62.86 & 66.25 & - & 19.00 & 0.722\\
\hline
\textit{HoldOut-I.D.} & \textit{gpt3.5} & 63.47 & 58.93 & 61.20 & \multirow{2}{*}{21.23} & 18.84 & 0.500\\
\textit{HoldOut-Comp.} & \textit{gpt3.5} & 66.43 & 30.00 & 48.21 &  & 20.10 & 0.707\\
\hline
\textit{FewShot-I.D.} & \textit{gpt3.5} & 60.09 & 60.89 & 60.49 & \multirow{2}{*}{19.85} & 19.31 & 0.583\\
\textit{FewShot-Comp.} & \textit{gpt3.5} & 67.41 & 29.55 & 48.48 &  & 16.54 & 0.562\\
\hlineB{2.5}
\end{tabularx}
\caption{\label{main-results-Mixture-gpt-3.5-anony}
The result of baseline \textit{ChatGPT} (gpt-3.5-turbo-0613)~\cite{openaiChatGPT} in dataset \textit{Mixture}.
}
\end{table*}

\label{results on datasets}

\end{document}